\documentclass[journal]{IEEEtran}
\usepackage{xcolor}
\usepackage{enumitem}
\usepackage{amsmath}
\usepackage{amssymb}
\usepackage{colortbl}
\usepackage{color}
\definecolor{mygray}{gray}{.75}
\usepackage{graphicx}
\usepackage{epstopdf}
\usepackage{algorithm,algpseudocode,algorithmicx}
\usepackage{amsmath,latexsym,amssymb,amsthm,array,amsfonts,algorithm,algpseudocode,multirow,cuted,stfloats}
\usepackage{footmisc}
\usepackage{cite}
\usepackage{units}
\usepackage{tabularx}
\usepackage{multirow}
\usepackage{booktabs}
\usepackage{graphicx}
\usepackage[caption=false,font=footnotesize]{subfig}
\usepackage{bm}
\usepackage{float}
\usepackage{soul}
\newlength\savewidth

\newcommand{\itrn}{\ensuremath{N_\m{iter}}}   
\newcommand{\m}[1]{\ensuremath{\mathrm{#1}}}

\ifCLASSINFOpdf
\else
\fi
\begin{document}
%
\title{Alleviating Search Bias in Bayesian Evolutionary Optimization with Many Heterogeneous Objectives}
%
%
%

\author{Xilu~Wang, Yaochu~Jin, \emph{Fellow}, \emph{IEEE}, Sebastian~Schmitt, and Markus~Olhofer 
\thanks{Xilu Wang is with the Department of Computer Science, University of Surrey, Guildford, GU2 7XH, United Kingdom. Email: xilu.wang@surrey.ac.uk.}

\thanks{Yaochu Jin is with the Faculty of Technology, Bielefeld University, 33619 Bielefeld, Germany. He is also with the Department of Computer Science, University of Surrey, Guildford, GU2 7XH, United Kingdom. Email: yaochu.jin@uni-bielefeld.de. (\textit{Corresponding author})}

\thanks{Sebastian Schmitt and Markus Olhofer are with the Honda Research Institute Europe GmbH, Carl-Legien-Strasse 30, D-63073 Offenbach/Main, Germany. Email: \{sebastian.schmitt;markus.olhofer\}@honda-ri.de.}

\thanks{Manuscript received December 6, 2018; revised xx, 2019.}
}

\maketitle

\begin{abstract}
Multi-objective optimization problems whose objectives have different evaluation costs are commonly seen in the real world. Such problems are now known as multi-objective optimization problems with heterogeneous objectives (HE-MOPs). So far, however, only a few studies have been reported to address HE-MOPs, and most of them focus on bi-objective problems with one fast objective and one slow objective. In this work, we aim to deal with HE-MOPs having more than two black-box and heterogeneous objectives. To this end, we develop a multi-objective Bayesian evolutionary optimization approach to HE-MOPs by exploiting the different data sets on the cheap and expensive objectives in HE-MOPs to alleviate the search bias caused by the heterogeneous evaluation costs for evaluating different objectives. 
To make the best use of two different training data sets, one with solutions evaluated on all objectives and the other with those only evaluated on the fast objectives, two separate Gaussian process models are constructed. In addition, a new acquisition function that mitigates search bias towards the fast objectives is suggested, thereby achieving a balance between convergence and diversity. We demonstrate the effectiveness of the proposed algorithm by testing it on widely used multi-/many-objective benchmark problems whose objectives are assumed to be heterogeneously expensive.

\end{abstract}


%
\IEEEpeerreviewmaketitle


\section{Introduction}
%
%
%
%

\IEEEPARstart{S}{urrogate-assisted} evolutionary algorithms (SAEAs) are powerful tools for optimizing computationally expensive multi-objective problems (MOPs), where several conflicting objective functions must be simultaneously optimized and the evaluations of the objectives are highly time-consuming or costly. While conventional multi-objective evolutionary algorithms (MOEAs) assume that each candidate solution can be accurately evaluated, SAEAs typically construct computationally efficient surrogate models to approximate the expensive real objective functions, and then the surrogates are used together with the expensive real objective functions to guide the evolutionary optimization. Alongside the choice of surrogate model, a model management strategy that determines which candidate solutions are evaluated using the expensive objective functions is pivotal for the success of an SAEA \cite{jin2011surrogate,allmendinger2015multiobjective,jin2019data}. Various classification or regression models are commonly used as surrogates in SAEAs, including support vector machines \cite{kang2016slope}, radial basis function networks \cite{guo2018heterogeneous}, feedforward neural networks \cite{pan2018classification}, and Gaussian processes (GPs), also known as Kriging which is a special case of GPs \cite{chugh2018surrogate,wang2020adaptive}. Among them, the GP is a popular choice for modelling expensive objective functions due to its ability of capturing the model's beliefs over the unknown objective function, providing both estimated objective values and the uncertainty of the estimations. The estimations provided by GPs can be utilized to design an acquisition function to select the next new data point to query (i.e., to evaluate the solution using the real objective functions), guiding the search of the optimum. An SAEA with a GP as the surrogate model and an acquisition function for model management is known as Bayesian optimization \cite{shahriari2015taking}. Bayesian optimization has become popular and well-known in engineering since Jones \emph{et al.} \cite{jones1998efficient} introduced Efficient Global Optimization (EGO), which employs expected improvement (EI) as the acquisition function. The past decade has witnessed a rapid development of Bayesian optimization, including various new acquisition functions and various real-world applications, and Bayesian optimization algorithms have exhibited promising performance in many studies \cite{snoek2012practical}. 

SAEAs typically assume that the computational complexities of all objective functions are similar, which enables the evaluation of different objective functions of a candidate solution to be completed at the same time. Consequently, the selection operator of an MOEA can be conducted and the evolutionary search can proceed to the next generation. This assumption, however, can be violated in practice, e.g., different computationally expensive objective functions in an expensive MOP are of varying computational complexities: the evaluation of aerodynamic and structural mechanics performance of an airplane wing design or a car shape involves computationally intensive computational fluid dynamics (CFD) and finite element analysis simulations, where several hours of evaluation time are typical for a single fitness evaluation. Additionally, some types of evaluations can be an order of magnitude slower than others, for example, crashworthiness assessment is much more resource- and time-consuming than CFD simulations. Such MOPs exhibit so-called heterogeneous objectives \cite{allmendinger2021heterogeneous} and, in particular, we consider heterogeneously expensive MOPs (HE-MOPs) in this paper, where non-uniform evaluation times in expensive MOPs give rise to the heterogeneity.

Most recently, new MOEAs have been proposed to effectively address HE-MOPs. Most of the algorithms, however, are limited to considering a class of bi-objective HE-MOPs having one computationally cheap (fast) objective function $f^c$ and one computationally expensive objective function $f^e$ (also called delayed or slow objective). Existing methods for handling HE-MOPs can be roughly categorized into two groups, i.e., non-surrogate based and surrogate based methods, which are briefly reviewed below.

\textbf{Non-surrogate based methods}: Allmendinger \emph{et al.} \cite{allmendinger2013hang} first introduced HE-MOPs and proposed a ranking-based MOEA to allow solutions with missing objective function values caused by $f^e$ to guide the search. Three strategies are developed to generate a pseudovalue for filling a missing objective value: 1) drawing a random pseudovalue from the interval of the minimum and maximum objective values obtained so far; 2) adding noise to the value of $f^e$ associated with a randomly selected solution; 3) inheriting the delayed objective value of the nearest neighbor. Subsequently, new selection strategies are proposed based on the ranking subject to missing objective values. In a follow-up work by Allmendinger \emph{et al.} \cite{allmendinger2015multiobjective}, MOPs with non-uniform latencies are defined more formally based on the framework of MOEAs, and three general schemes are proposed to handle heterogeneous objectives, including \emph{Waiting}, \emph{Fast-first} and \emph{Interleaving schemes}. While \emph{Waiting} directly applies an MOEA to HE-MOPs by waiting for the completion of fitness evaluations of the delayed objective, \emph{Fast-first} employs a single-objective evolutionary algorithm to consume the additional fitness evaluations available for $f^c$ during the waiting of expensive evaluations, maximizing the use of available cheap evaluations. Unlike \emph{Fast-first}, more elaborated strategies are introduced in \emph{Interleaving schemes} (i.e. \emph{brood} and \emph{speculative interleaving}) to utilize the limited evaluations by integrating the search results of each objective. Although the non-surrogate based methods shed light on possible directions for handling HE-MOPs, a major remaining issue is that the obtained solutions may be still far from Pareto optimal due to the limited evaluation budget available in expensive MOPs. In addition, how well they can scale to more complex problem settings with more objectives has not been explored.

\textbf{Surrogate based methods}: More recently SAEAs have been extended to HE-MOPs, which is motivated by the fact that SAEAs have emerged as powerful methods for the optimization of MOPs with expensive evaluations. Chugh \emph{et al.} \cite{chugh2018hkrvea} proposed a heterogeneous Kriging-assisted evolutionary algorithm, called HK-RVEA, to make use of the heterogeneity within the SAEA framework. Similar to \emph{Interleaving schemes}, HK-RVEA adopted a single-objective evolutionary algorithm (SOEA) and genetic operators to generate solutions for cheap evaluations available when the initial population and the new samples are submitted for evaluations on both objectives, respectively. To make use of the additional evaluations on $f^c$, Wang \emph{et al.} \cite{XiluWang2020TSAEA} developed a parameter-based transfer learning strategy based on a GP-assisted evolutionary algorithm (T-SAEA). In T-SAEA, common decision variables related to both $f^c$ and $f^e$ are determined first using a filter-based feature selection, then the corresponding parameters in the GP models can be shared, thereby improving the quality of the GP surrogate for $f^e$. However, the performance of T-SAEA degrades when the delay length becomes large. Moreover, the performance of T-SAEA depends on the selected pivot features by the filter-based feature selection. In a follow-up work, Wang \emph{et al.} \cite{wang2021transfer} proposed an instance-based transfer learning method (Tr-SAEA) to address the heterogeneous data in bi-objective problems. Domain adaptation techniques are adopted to generate synthetic samples for $f^e$, and a GP-based co-training method is introduced to augment the training data for surrogate models of $f^e$ using the unlabeled synthetic data. Unfortunately, Tr-SAEA only learns the mapping in the objective space and requires an additional optimization method to obtain the corresponding solutions in the search space. It is well-known that there exists a functional relationship between $f^e$ and $f^c$ (usually a trade-off relation) for solutions on the Pareto front. Hence, in \cite{WangTC-SAEA2021} a co-surrogate is adopted to model the relationship between the objective functions. Subsequently, transferable instances are identified from the search of $f^c$ to speed up the optimization of $f^e$. Interestingly, trust region methods, instead of evolutionary algorithms, with the use of surrogates have also been successfully applied to HE-MOPs. For example, multi-objective heterogeneous trust region algorithms \cite{thomann2019representation,thomann2019trust} are proposed to apply a modified trust region algorithm with quadratic surrogate models to deal with heterogeneity. While HK-RVEA, T-SAEA and Tr-SAEA are developed to tackle bi-objective problems with heterogeneous objectives, multi-objective heterogeneous trust region algorithms are easily scalable to any number of objectives; however, the multi-objective heterogeneous trust region algorithms hinges on strong assumptions: the expensive objective functions are black-box and twice continuously differentiable, and the cheap ones are given analytically and derivatives are easily available \cite{thomann2019trust,thomann2019representation}.
To address the scalability issue in handling HE-MOPs with limited evaluation budgets, this work extends SAEAs to HE-MOPs with more general and more practical problem settings, i.e., HE-MOPs with more than three objectives (called HE-many-objective problems, HE-MaOPs) and different combinations of computationally cheap and expensive objectives. We propose an MOEA to reduce the search bias resulting from the heterogeneous computational complexities of the objectives and limit the computational time for training GP models with an increased amount of data on the cheap objectives. To this end, an ensemble surrogate consisting of two GPs is proposed, one trained only using the offline data sampled before optimization starts and the other constantly updated using newly sampled data during the optimization. In addition, a new acquisition function that penalizes search bias towards the cheap objectives is designed.  

The key contributions of the proposed search bias penalized BO, termed as SBP-BO, can be summarized as follows:
\begin{enumerate}
  \item To make use of the different amount of available data evaluated on cheap and expensive objectives, an ensemble consisting of two GPs is constructed for each cheap objective, while one GP is used to approximate each expensive objective. Before optimization starts, while the initial population is evaluated on all objective functions, the cheap objectives can be explored using a single-objective optimization algorithm, resulting in abundant extra solutions on the cheap objectives. Hence, for each cheap objective, one GP is trained with the solutions evaluated on all objectives, while the other is trained with the data only evaluated on cheap objectives during the initialization. Correspondingly, the former is re-trained with a selected subset of the offline data and new data sampled during the optimization, while the latter remains unchanged. For each expensive objective, the GP is always trained and updated with solutions evaluated on all objectives. This way, the prediction performance of the surrogate can be enhanced, while the computational cost is limited when the amount of data of the fast objectives grows. 

  \item To alleviate the search bias towards the fast objectives and achieve a good balance between the exploitation and exploration, a new acquisition function is proposed. The proposed acquisition function can not only evaluate the quality of a solution in terms of convergence and diversity, but also promotes the exploration on the expensive objectives. 


\end{enumerate}
The rest of paper is organized as follows. Section \ref{Preliminaries} provides a problem description, followed by an introduction to multi-objective Bayesian optimization including GPs and acquisition functions. Then, the proposed SBP-BO is introduced in Section III. Section IV provides details about the experimental settings and Section V presents the experimental results to demonstrate the effectiveness of SBP-BO. Finally, we draw conclusions and discuss future research.

\section{Background}\label{Preliminaries}
\subsection{Problem Description}
 Before we tackle the challenges posed by the presence of computationally heterogeneous objectives, we present the problem description at first. We consider expensive multi- and many-objective optimization problems with heterogeneous objectives (HE-MOPs and HE-MaOPs) in the following form:
\begin{equation}
\begin{array}{ll}
\min _{\mathbf{x}} & \mathbf{f}(\mathbf{x})=\left(f_{1}(\mathbf{x}), f_{2}(\mathbf{x}), \ldots, f_{m}(\mathbf{x})\right) \\
\text { s.t. } & \mathbf{x} \in X
\end{array}
\end{equation}
where $\mathbf{x}=\left(x_{1}, x_{2}, \cdots, x_{d}\right)$ is the decision vector with $d$ decision variables, $X$ denotes the decision space, the objective vector $\mathbf{f}$ consists of $m$ $(m>2)$ objectives and for MaOPs the number of objectives $m$ is larger than 3. The evaluation time of each objective is denoted by $\mathbf{t}=(t_1,t_2,\cdots,t_m)$, where we assume the objectives are ordered in terms of their computational complexity, ranked from the fastest with $i=1$ to the slowest $i=m$, i.e.\ $t_1\leq t_2\leq \cdots \leq t_m$.
The objective functions are black-boxes that can be evaluated by either time-consuming numerical simulations, or costly physical experiments. Building surrogate models based on the data collected via numerical simulations or experiments has been shown to be an efficient approach to such black-box expensive optimization problems \cite{jin2019data}. 
In this work, we assume that the evaluation of each objective function can be done in parallel, and the computation time for building surrogates and applying the genetic operators of the evolutionary algorithm is negligible compared to that for evaluating the true objective functions. 
Therefore, we characterize the heterogeneity of the objectives by the number of affordable evaluations of an objective $f_i$ relative to the slowest objective, $f_m$, which can be calculated given the evaluation time as the ratio $r_i=\lfloor \frac{t_m}{t_i} \rfloor$ \cite{allmendinger2021heterogeneous,allmendinger2015multiobjective}. Here, $\lfloor . \rfloor$ denotes the floor operation.

For convenience, we introduce a notation to divide the $m$ objectives into two groups based on the value of $r_i$. The $i$-th objective is called \textit{cheap}, denoted as $f^{c}$, if $r_i>r_\m{thres}$; the objective is called \textit{expensive}, denoted as $f^{e}$, if $r_i \leq r_\m{thres}$, where $r_\m{thres}$ is a threshold separating the cheap objectives from expensive objectives. For real-world problems that have a natural separation between  cheap and expensive objectives \cite{gelbart2014bayesian,hernandez2016predictive}, the threshold can be defined straightforwardly. In case there is no intuitive separation between cheap and expensive objectives, the threshold can be defined according to the user's preference. It should be pointed out that the partitioning of the objectives has no direct influence on the effectiveness of the proposed algorithm, which does not make any assumptions on the ratios. 
Together with the idea of constructing surrogates for all objectives, this makes the proposed algorithm generic and applicable to a wide range of problems. 

\subsection{Multi-objective Bayesian Optimization}
  Multi-objective Bayesian optimization, an extension of Bayesian optimization to MOPs, has been successfully applied to simultaneously optimizing expensive black-box multi-objective problems \cite{suzuki2020multi,belakaria2019max,allmendinger2017surrogate,chugh2018surrogate}. Multi-objective Bayesian optimization aims to perform a limited number of objective evaluations to identify a set of non-dominated solutions using an MOEA. Therefore, multi-objective Bayesian optimization typically first trains a Gaussian process using data collected from the previous evaluations to approximate each objective function of an MOP. A brief introduction to GPs is presented in Section I in the Supplementary material. An acquisition function is then adopted to determine where in the decision space to sample in the next by balancing exploration and exploitation. Based on the way in which Bayesian optimization and evolutionary algorithms work together, multi-objective Bayesian optimization can be further divided into two groups, evolutionary Bayesian optimization (EBO) and Bayesian evolutionary optimization (BEO) \cite{qin2019bayesian}. In BEO the evolutionary algorithm is the basic framework where the acquisition function is adopted as a criterion for model management, while in EBO Bayesian optimization is the basic framework in which the acquisition function is optimized using an evolutionary algorithm. In the following, we will briefly review the existing work on multi-objective Bayesian optimization.
  
A straightforward way to address expensive MOPs using Bayesian optimization is to decompose an MOP into multiple single-objective problems, so that existing acquisition functions for single-objective optimization can be directly applied to MOPs. The combination of scalarization and evolutionary algorithms is a prominent approach in this line of research. In the work of Knowles \cite{knowles2006parego}, a vector of objective values is converted into a scalar value using the Tchebycheff scalarization function, so that the standard EI can be adopted as the acquisition function when optimizing MOPs. Similarly, some decomposition-based MOEAs (e.g., MOEA/D \cite{zhang2007moea} and RVEA \cite{cheng2016reference}) that decompose an MOP into a set of single-objective subproblems, have been extended to address expensive MOPs with GPs for function approximation and the EI as the acquisition function \cite{zhang2009expensive,chugh2018surrogate}. Wang \emph{et al.} \cite{wang2020adaptive} employed RVEA as the optimizer and proposed an adaptive acquisition function based on lower confidence bound to dynamically tune the weights of the uncertainty and the predicted mean fitness value according to the search dynamics. 

Quality indicators for a non-dominated solution set (performance metrics) were originally developed to assess and compare the quality of solution sets (rather than a single solution) obtained by different algorithms \cite{zitzler2003performance}. Interestingly, some quality indicators have been employed as a scalar measure for assessing the contribution of single solutions, reducing an MOP to a single-objective optimization problem. Various multi-objective Bayesian optimization methods with extended acquisition functions based on performance indicators have been developed, among which the hypervolume (HV) is the most commonly used. An early and popular performance indicator based algorithm is $\mathcal{S}$-Metric-Selection-based Efficient Global Optimization (SMS-EGO) \cite{ponweiser2008multiobjective}, which is based on the $\mathcal{S}$ metric or HV metric \cite{beume2007sms}. The combination of the EI and HV, which is known as Expected hypervolume improvement (EHVI), is more commonly seen in the context of expensive MOPs. Arguably, EHVI was first introduced in \cite{emmerich2006single} to provide a scalar measure for improvement for pre-screening solutions in SAEAs, and then became popular as a treatment of expensive optimization with multiple objectives \cite{couckuyt2014fast,yang2019multi}.
 
Given the popularity of information theoretic approaches to single-objective Bayesian optimization, it is not surprising that there have been ample extensions of information-based acquisition functions for tackling expensive MOPs based on the information theory. For example, predictive entropy search \cite{hernandez2014predictive} has been extended to MOPs by maximally reducing the entropy of the posterior distribution over the Pareto set, called PESMO; however, it is computationally expensive to approximate and optimize PESMO. Belakaria \emph{et al.} \cite{belakaria2019max} developed a max-value entropy search for multi-objective Bayesian optimization to reduce the computational burden incurred by the optimization of information-theoretic acquisition functions. A subsequent work is the extension of the output space entropy based acquisition function in the context of MOPs, known as MESMO \cite{belakaria2019max}. Empirical results show that MESMO is more efficient than PESMO. However, MESMO fails to capture the trade-off relations between objectives for MOPs where no points in the Pareto front are near the maximum of each axis \cite{suzuki2020multi}. To address this issue, Suzuki \emph{at al.} \cite{suzuki2020multi} proposed a \emph{Pareto-frontier entropy search} (PFES) that considers the entire Pareto front.

 In this work, the proposed acquisition function is based on the adaptive acquisition function $\boldsymbol{AF}_{A}$ \cite{wang2020adaptive} to evaluate the quality of a solution, owing to its computational efficiency and promising performance on MOPs. Considering an $m$-objective minimization problem $\left \{ f_{i}(\mathbf{x})\right \}_{i=1}^m$, the predictions of a candidate solution $\mathbf{x}$ is derived from the GP of each objective. The vectors of the predicted mean and variance of $\mathbf{x}$ are denoted as $\boldsymbol{\sigma}(\mathbf{x})=\left \{  \sigma_{i}(\mathbf{x})\right \}_{i=1}^m$ and $\boldsymbol{\mu}(\mathbf{x})=\left \{ \mu_{i}(\mathbf{x})\right \}_{i=1}^m$, respectively. The adaptive acquisition function for the candidate solution $\mathbf{x}$ is defined as
\begin{equation}
\label{eq:AFF}
\boldsymbol{AF}_{A}(\mathbf{x})=(1-\alpha )\cdot (\boldsymbol{\mu}(\mathbf{x})./\boldsymbol{\mu}_{max})+ \alpha\cdot(\boldsymbol{\sigma}(\mathbf{x})./\boldsymbol{\sigma}_{max})
\end{equation}
where 
\begin{equation}
\alpha =-0.5\cdot {\rm cos}\Big(\frac{FE}{FE_{max}}\cdot \pi\Big)+0.5
\end{equation}
is an adaptation parameter defined by a cosine function. `$./$' denotes element-by-element division for two vectors of the same size. $FE$ and $FE_{max}$ denote the current and maximum number of real objective function evaluations; $\boldsymbol{\mu}_{max}$ and $\boldsymbol{\sigma}_{max}$ represent the maximum values of the mean and variance values provided by GPs on the current population. Hence, both the predicted objective value and the uncertainty are normalized to $[0,1]$. Note that $\boldsymbol{AF}_{A}( {\mathbf{x}})$ is a vector with a length of $m$ and minimised by an MOEA. Even though the formulation of $\boldsymbol{AF}_{A}( {\mathbf{x}})$ looks similar to lower confidence bound \cite{cox1992statistical}, the different sign in front of the term proportional to the variance makes a big difference. 
While the lower confidence bound  tries to balance exploration and exploitation by maximising the uncertainty and at the same time minimising the mean value, the motivation behind $\boldsymbol{AF}_{A}( {\mathbf{x}})$ is different. 
The $\boldsymbol{AF}_{A}( {\mathbf{x}})$ attempts to achieve relatively fast convergence in the early stage by assigning a large weight to the mean value. 
As the optimization progresses, the contribution of the uncertainty is increased in the acquisition function, which makes regions of the search space with high variance less attractive for the search. 
Therefore, an even more exploitative search is achieved by selecting samples with minimized uncertainty.  
The explorational aspect of the search is realized by the multi-objective nature of the selection operator in the MOEA, and a set of evenly distributed and well converged solutions will be realized. 

\section{Proposed Algorithm}
\label{section:Algorithm}
In this section, the proposed search bias penalized Bayesian optimization (SBP-BO) algorithm for solving HE-MOPs and HE-MaOPs is introduced. Consequently, SBP-BO has two main components. An ensemble surrogate to approximate the cheap objective functions, and a search bias penalized acquisition function to select new samples considering the impact of the heterogeneous evaluation time. In the following, we will introduce the notations and the framework step by step, and then present the details of the two key components.
\subsection{Basic Ideas and the Overall Framework}

\begin{figure}[ht]
\centering
\includegraphics[width=\columnwidth]{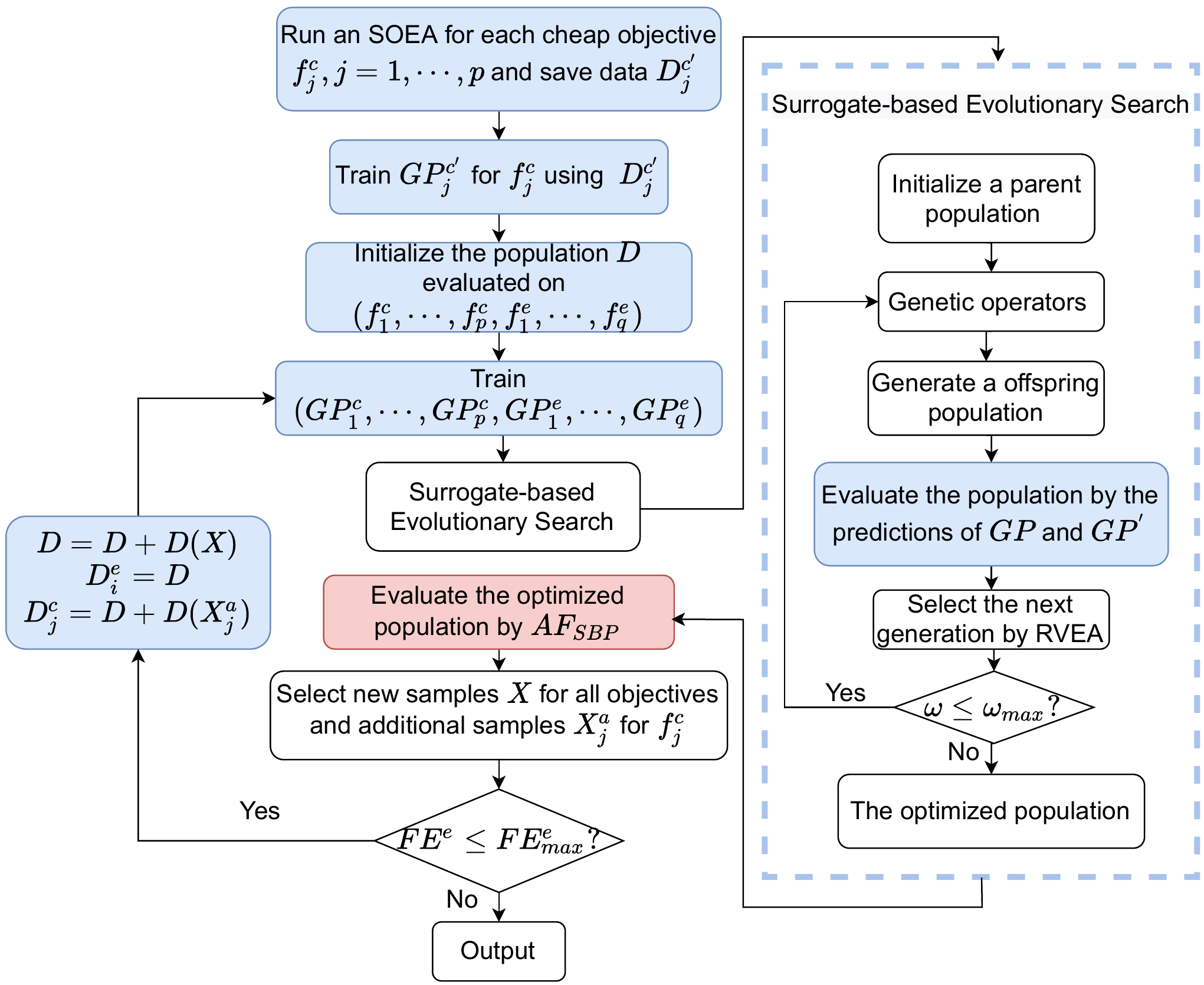}
\caption{The overall framework of SBP-BO.}
\label{Fig.1}
\end{figure}
\begin{algorithm}[htbp]\footnotesize{
\renewcommand{\algorithmicrequire}{\textbf{Input:}}
\renewcommand{\algorithmicensure}{\textbf{Output:}}
\caption{The framework of SBP-BO} \algblock{Begin}{End}
\label{Algorithm 1}
\begin{algorithmic}[1]
\Require $FE_{max}^{e}$: the maximum number of the expensive objective function evaluations; $\boldsymbol{r}$: the ratio of the evaluation times between the expensive and cheap objectives; $u$: the number of new samples for updating the GP models; $w_{max}$: the maximum number of generations before updating GP models;
\Ensure Optimal solutions in $D$;
\State Initialization: Use the Latin Hybercube Sampling method to sample an initial population $\boldsymbol{P}$; $\boldsymbol{P}$ is evaluated on all objective functions, obtaining the objective values $\boldsymbol{Y_P}$; set $D={(\boldsymbol{P}, \boldsymbol{Y_P})}$ and train Gaussian processes $GP=[GP_{1}^{c},\cdots,GP_{p}^{c},GP_{1}^{e},\cdots,GP_{q}^{e}]$ for each objective using $D$ that evaluated on all objectives; run an SOEA to optimize $f_{j}^{c},j=1,\cdots,p$ and save data in $D_{j}^{c'}$, then $GP_{j}^{c'}$ for each cheap objective $f_{j}^{c}$ is trained on $D_{j}^{c'}$; set $FE= |D|$, $w=1$ and $\itrn=1$.\;
 \While{$FE^{e}\leqslant FE_{max}^{e}$}
    \State //Using the surrogate in the RVEA//
    \State Create the initial population;\;
     \While{$w\leqslant w_{max}$}
       \State Generate offspring using the simulated binary crossover and polynomial mutation;\;
       \State Use the ensemble to predict cheap objective values and the $GP_{1}^{e},\cdots, GP_{q}^{e}$ to predict the expensive objective values on the combined population;\;
       \State Use the reference vector guided selection to select the next generation;\;
       \State Perform the reference-vector-adaptation;\;
       \State $w=w+1$;\;
      \EndWhile
  \State Use the proposed acquisition function to evaluate the optimized solutions found by RVEA;\;
  \State Use the reference vector guided selection to determine $u$ new solutions $\boldsymbol{X}$ and $u\cdot r_{j}-u$ additional new solutions $\boldsymbol{X}_{j}^{a}, j=1,\cdots,p$ to be evaluated on all objectives and on each $f_{j}^{c}, j=1,\cdots,p$, respectively. Consequently, the newly solutions are saved in the corresponding datasets $D^{new}=(\boldsymbol{X}, \boldsymbol{Y})$ and $D_{j}^{a}=(\boldsymbol{X}_{j}^{a}, \boldsymbol{Y}_{j}^{c})$, respectively;\;
   \State Add $D^{new}$ to $D$ and select training data $D^t$ from data set $D$ evaluated on all objectives;\;
   \State $GP=[GP_{1}^{c},\cdots,GP_{p}^{c},GP_{1}^{e},\cdots,GP_{q}^{e}]$ is updated: $GP_{i}^{e}, i=1,\cdots,q$ is updated with $D^t$, while $GP_{j}^{c}, j=1,\cdots,p$ is updated with $D^t$ and $D_{j}^{a}$.
  \State Update $FE^{e}=FE^{e}+u$, $\itrn=\itrn+1$;
\EndWhile
 \State  Return the optimized solutions;
\end{algorithmic}}
\end{algorithm}
The overall framework of SBP-BO is presented in Fig. \ref{Fig.1}. The main steps of the proposed algorithm are as follows, the pseudo code containing more details is given in Algorithm \ref{Algorithm 1}:
\begin{itemize}
\item Step 1: Initialization. 
An initial population $\mathbf{P}$ with $n$ individuals is evaluated on $m$ real objective functions using Latin Hypercube sampling (LHS) that produces the offline data set $D=(\mathbf{P},\mathbf{Y_P})$. During the evaluation of the expensive objectives, each cheap objective $f_{j}^{c}, j=1,\cdots,p$ can be evaluated $n\cdot r_{j}$ times. This is done by using a single-objective evolutionary algorithm to optimize each objective $f_{j}^{c}$, and the corresponding data is saved as $D_{j}^{c'}$.

\item Step 2: Initial construction of GP models. SBP-BO constructs Gaussian processes $GP=[GP_{1}^{c},\cdots,GP_{p}^{c},GP_{1}^{e},\cdots,GP_{q}^{e}]$ for each objective using the data $D$ evaluated on all objectives. 
For each cheap objective $f_{j}^{c}, j=1,\cdots,p$, an extra Gaussian process $GP_{j}^{c'}$ is trained on the additional data set $D_{j}^{c'}$, allowing us to construct an ensemble with two members $GP_{j}^{c}$ and $GP_{j}^{c'}$. 

\item Step 3: Selection of new samples. Similar to the standard GP-assisted SAEA, SBP-BO adopts a baseline MOEA to optimize the HE-MOP/HE-MaOP for a certain number of generations, in which the ensemble surrogate predicts the value of each cheap objective $f_{j}^{c}, j=1,\cdots,p$ for the candidate solutions, while $GP_{i}^{e}$ predicts the value of the expensive objectives $f_{i}^{e}, i=1,\cdots,q$. In our case, the reference vector guided evolutionary algorithm (RVEA) \cite{cheng2016reference} is adopted as the baseline MOEA, where a reference vector guided selection is introduced to select the next generation according to the angle-penalized distance. Subsequently, all individuals in the optimized population are evaluated by the proposed acquisition function. Again, the reference vector guided selection is employed to identify a set of promising solutions, in which $u$ new query points $\mathbf{X}$ and $u\cdot r_{j}-u$ additional query points $\mathbf{X}_{j}^{a}$ are randomly selected to be evaluated on all objectives and on cheap objectives $f_{j}^{c}$, respectively. Consequently, the newly evaluated solutions $\mathbf{X}$ are added to dataset $D$.

\item Step 4: Update of GP models. SBP-BO follows a strategy used in \cite{guo2018heterogeneous,knowles2006parego} to manage the training data: A predefined maximum number $L$ of training data is set to $11d-1+25$ \cite{knowles2006parego}, where $d$ is the number of decision variables. When the number of data samples in $D$ is less than $L$, e.g., in the beginning of the optimization, all solutions in $D$ are used to train the GP models, i.e.\ $GP=[GP_{1}^{c},\cdots,GP_{p}^{c},GP_{1}^{e},\cdots,GP_{q}^{e}]$. If $D$ contains more samples than $L$, a subset $D^t$ will be selected from the training data archive to limit the computation time, where the quality of the Gaussian processes and the optimization performance are considered. Since the solutions in $D$ have been evaluated on all objectives, existing training data management methods for GP-assisted MOEAs, such as K-RVEA and MOEA/D-EGO, also can be used. While the GPs for the slow objectives $GP_{i}^{e}$, $i=1,\cdots,q$ are updated with the selected $L$ training data samples $D^t$, each $GP_{j}^{c},j=1,\cdots,p$ is trained using both $D^t$ and the extra new samples $\boldsymbol{X}_{j}^{a}$. 
Note that $GP_{j}^{c'}, j=1,\cdots,p$ remain unchanged during the optimization, which are trained once with the offline data generated in Step 1 only. 
\item Repeat Step 3 and Step 4 until the allowed computation budget is exhausted.
\end{itemize}

\subsection{Ensemble Surrogate for Cheap Objectives}
As described above, an ensemble surrogate including two GP models, $GP_{j}^{c}$ and $GP_{j}^{c'}$, are constructed for each cheap objective $f_{j}^{c}, j=1,\cdots,p$. The predicted mean value of $f_{j}^{c}$ on a new sampled solution $\mathbf{x}$ provided by the ensemble is a weighted combination of the predictions of $GP_{j}^{c}$ and $GP_{j}^{c'}$. Motivated by product of experts \cite{cao2014generalized}, confident predictions should have more influence on the combined prediction than the less confident ones. Hence, the weight is calculated based on the level of uncertainty of each GP’s prediction, and the ensemble prediction is 
\begin{equation}
\begin{aligned}
\mu_{j}^{c}(\mathbf{x})&=\alpha_j\,\mu_{GP_{j}^{c}}(\mathbf{x})+\beta_j\,\mu_{GP_{j}^{c'}}(\mathbf{x})\\
\alpha_j &= \frac{\sigma_{GP_{j}^{c'}}(\mathbf{x})}{\sigma_{GP_{j}^{c'}}(\mathbf{x})+\sigma_{GP_{j}^{c}}(\mathbf{x})} \\
\beta_j &= \frac{\sigma_{GP_{j}^{c}}(\mathbf{x})}{\sigma_{GP_{j}^{c'}}(\mathbf{x})+\sigma_{GP_{j}^{c}}(\mathbf{x})}\
\end{aligned}
\end{equation}
where $\mu_{GP_{j}^{c}}$ and $\sigma_{GP_{j}^{c}}$ are the predictions of $GP_{j}^{c}$, and $\mu_{GP_{j}^{c'}}$ and $\sigma_{GP_{j}^{c'}}$ are the predictions of $GP_{j}^{c'}$. As mentioned earlier, training two separate GP models on $D$ and $D^{c'}$ makes the update of the GP models more efficient and computationally more effective. Specifically, while $GP=[GP_{1}^{c},\cdots,GP_{p}^{c},GP_{1}^{e},\cdots,GP_{q}^{e}]$ are updated with newly sampled data selected according to the acquisition function, we do not re-train $GP^{c'}=[GP_{1}^{c'},\cdots,GP_{p}^{c'}]$ during the optimization.
Moreover, the method for selecting training samples used to retrain $GP=[GP_{1}^{c},\cdots,GP_{p}^{c},GP_{1}^{e},\cdots,GP_{q}^{e}]$,  as presented in Step 4, limits the maximum training time while allowing for building an effective model using relevant samples. 

 This training scheme makes it possible to adopt an existing strategy for selecting training data to update $GP=[GP_{1}^{c},\cdots,GP_{p}^{c},GP_{1}^{e},\cdots,GP_{q}^{e}]$ in the context of HE-MOPs. Specifically, the number of training data is typically capped to limit the computational complexity of constructing the surrogates, which is a common practice in GP-based evolutionary algorithms \cite{chugh2018surrogate,knowles2006parego,guo2018heterogeneous}. For standard MOPs, it is desirable to select a subset so that the quality of the surrogates can be improved as much as possible, and the resulting surrogate-assisted search can maintain a balance between convergence and diversity. Note that each solution is evaluated on all objectives in standard MOPs, so that the balance between convergence and diversity can be estimated by selection criteria in MOEAs, such as nondominated sorting and the crowding distance. However, this is not the case for HE-MOPs since many solutions are partially evaluated. Consequently, it is difficult to select a subset that can balance convergence and diversity. To tackle this challenge, we train two GPs ($GP_{j}^{c}$ and $GP_{j}^{c'}$) using solutions evaluated on all objectives and solutions evaluated on fast objectives, respectively. Hence, the existing strategies for selecting training data \cite{knowles2006parego,guo2018heterogeneous} can be directly applied to HE-MOPs as we update $GP_{j}^{c}$ only.

\subsection{Search Bias Penalized Acquisition Function}


In order to alleviate the search bias towards the cheap objectives resulting from the heterogeneous evaluation times, we propose to include a penalty term in the acquisition function for alleviating the search bias, which prioritizes the expensive objectives in minimizing the acquisition function. 
This penalty term is multiplied by the adaptive acquisition function reported in Eq.\ \eqref{eq:AFF}, resulting in a search bias penalized acquisition function ($\boldsymbol{AF}_{SBP}$). Having obtained the optimized population by RVEA using the GP ensemble, the mean and variance of the objective values of all individuals are predicted by the GP models at first. Given a candidate solution $\mathbf{x}$ in the optimized population, the proposed acquisition function can be computed analytically as follows:

\begin{equation}
\boldsymbol{AF}_{SBP}(\mathbf{x},\itrn)=\boldsymbol{AF}_{A}(\mathbf{x})\,\circ\, \mathbf{SBP}(\mathbf{x}, \itrn)
\end{equation}
where $\itrn$ is the current iteration number of the Bayesian optimization loop,  $\mathbf{SBP}(\mathbf{x}, \itrn)$ is the penalty term to be described below in detail, $\boldsymbol{AF}_{A}(\mathbf{x})$ denotes the acquisition function of Eq.~\eqref{eq:AFF}, and $\circ$ denotes component-wise multiplication. As a result, each individual in the population can be evaluated according to $\boldsymbol{AF}_{SBP}$, obtaining a vector with length $m$. Hence, minimising $\boldsymbol{AF}_{SBP}$ is still a multi-objective optimization problem, and therefore the reference vector guided selection in RVEA is adopted in this work.

Let $\boldsymbol{\mu}(\mathbf{x})=\left\{\mu_i\right\}_{i=1}^{m}$ denote the predicted mean value of the objective vector with length $m$ on the candidate solution $\mathbf{x}$ in the optimized population, and the maximum and minimum value of the predicted mean of the optimized population will be identified and denoted as $\boldsymbol{\mu}^{max}=\left\{\mu_i^{max} \right\}_{i=1}^{m}$ and $\boldsymbol{\mu}^{min}=\left\{\mu_i^{min} \right\}_{i=1}^{m}$, respectively. In order to calculate the penalty terms for $\mathbf{x}$, first we normalize the objective vector into the same range, i.e, [0,1] 
\begin{equation}
\bar{\boldsymbol{\mu}}(\mathbf{x})=(\boldsymbol{\mu}(\mathbf{x})-\boldsymbol{\mu}^{min})./(\boldsymbol{\mu}^{max}-\boldsymbol{\mu}^{min}),
\end{equation}
where again $./$ indicates component-wise division. Hence, $\bar{\boldsymbol{\mu}}(\mathbf{x})=\left\{\bar{\mu}_i\right\}_{i=1}^{m}$ is a vector with length $m$, and the corresponding penalty term  $\mathbf{SBP}=\left\{SBP_i\right\}_{i=1}^{m}$ for the data sample $\mathbf{x}$ is also defined as a vector including the penalty term for each objective. 

For the $i$-th objective, the penalty term $SBP_i$ is calculated as 
\begin{equation}
SBP_i(\bar{\mu}_i, \itrn)=
1-\pi\left(\bar{\mu}_i, \itrn\right)
\label{Eq.SBP}
\end{equation}
where $\pi\left(\bar{\mu}_i, \itrn\right)$ is calculated using an exponential distribution function,
\begin{align}
\pi\left(\bar{\mu}_{i}, \itrn\right)=\lambda\left(\itrn\right) e^{-\lambda\left(\itrn\right) \bar{\mu}_{i}}
\end{align}
with
\begin{equation}
\lambda(\itrn)=\frac{1}{w_i\itrn+1},
\end{equation}
and
\begin{equation}
w_i=\frac{r_i}{\sum_{i}^{m}r_i},
\end{equation}
where $w_i$ encodes the relative number of affordable evaluations of objective $i$.
This penalty is intuitively motivated by the exponential distribution in modeling situations in which certain events occur at a constant probability per unit length \cite{abdolshah2019cost}. Due to the fact that fast objectives can be explored more often than slow objectives in HE-MOPs, we construct different exponential distribution functions $\pi\left(\bar{\mu}_{i}, \itrn\right)$ for each objective function based on the evaluation times of different objectives to alleviate search biases. For example, the exploration on a fast objective is expected to occur at a lower probability than that on slow objectives. This is achieved by generating different $\lambda(\itrn)$ values with respect to the evaluation times and the number of iterations. Hence,
a large number of affordable function evaluations, i.e.\ a larger value of $r_i$, of an objective function $f_i$ will result in a higher value of $w_i$, and accordingly a smaller value of $\lambda$. Since $0\leq \bar{\mu}_i\leq 1$ and $0\leq \lambda \leq 1$, this leads to more uniformly distributed and smaller values for $\pi$ and therefore results in a larger penalty value on the corresponding fast objectives. Therefore, the acquisition function will prefer new samples that not only balance the local exploitation and global exploration, but also reduce the search bias by prioritizing the exploration for selecting slow functions. Similarly, as optimization progresses and $\itrn$ increases, the penalty term will approach a value of 1 for all objectives, gradually reducing the disadvantage over the fast objectives in the acquisition function.

\begin{figure}[!htbp]
\centerline{
\subfloat[$\itrn$=1]{\includegraphics[width=1.8in]{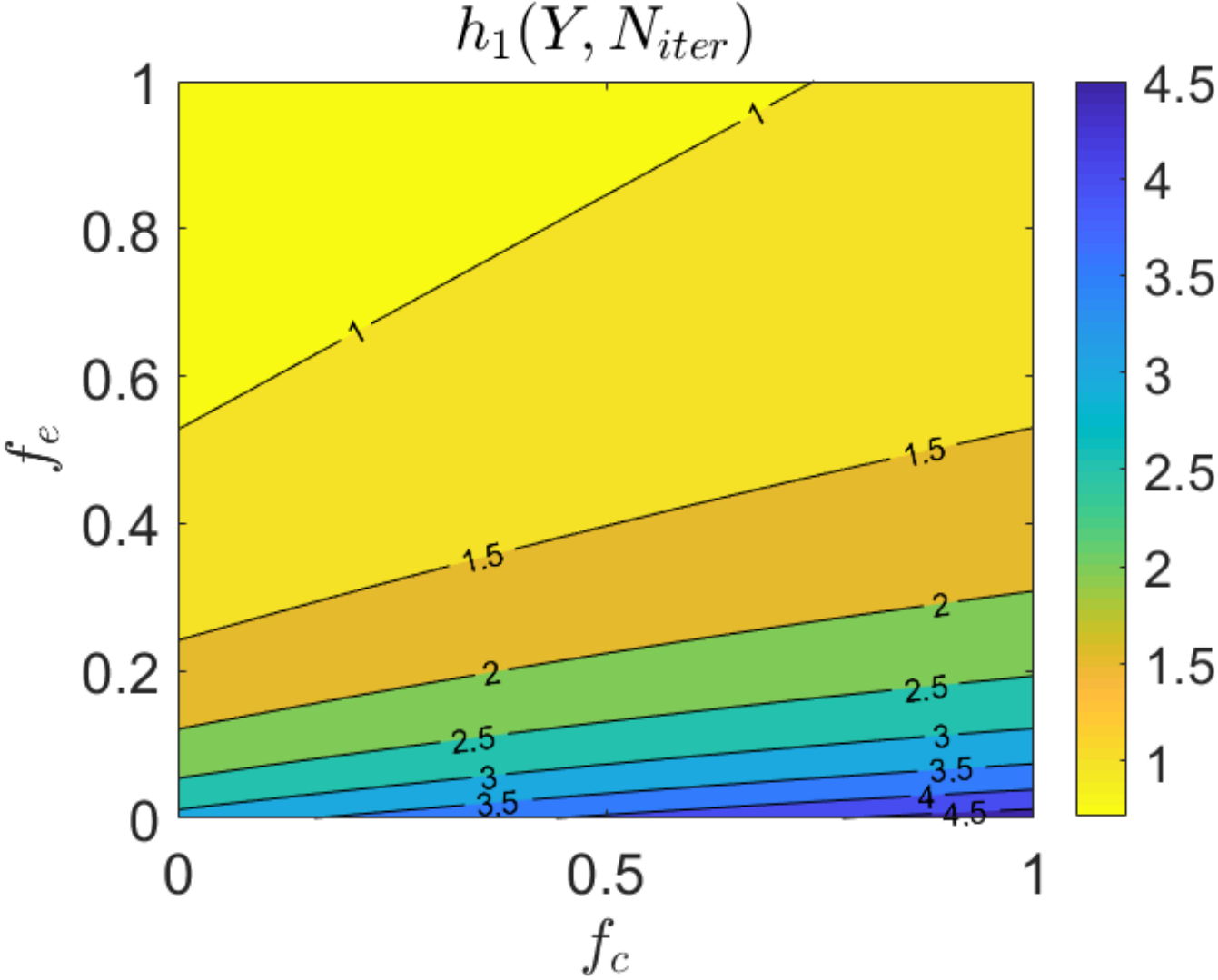}%
\hfil
}
\subfloat[$\itrn$=1]{\includegraphics[width=1.8in]{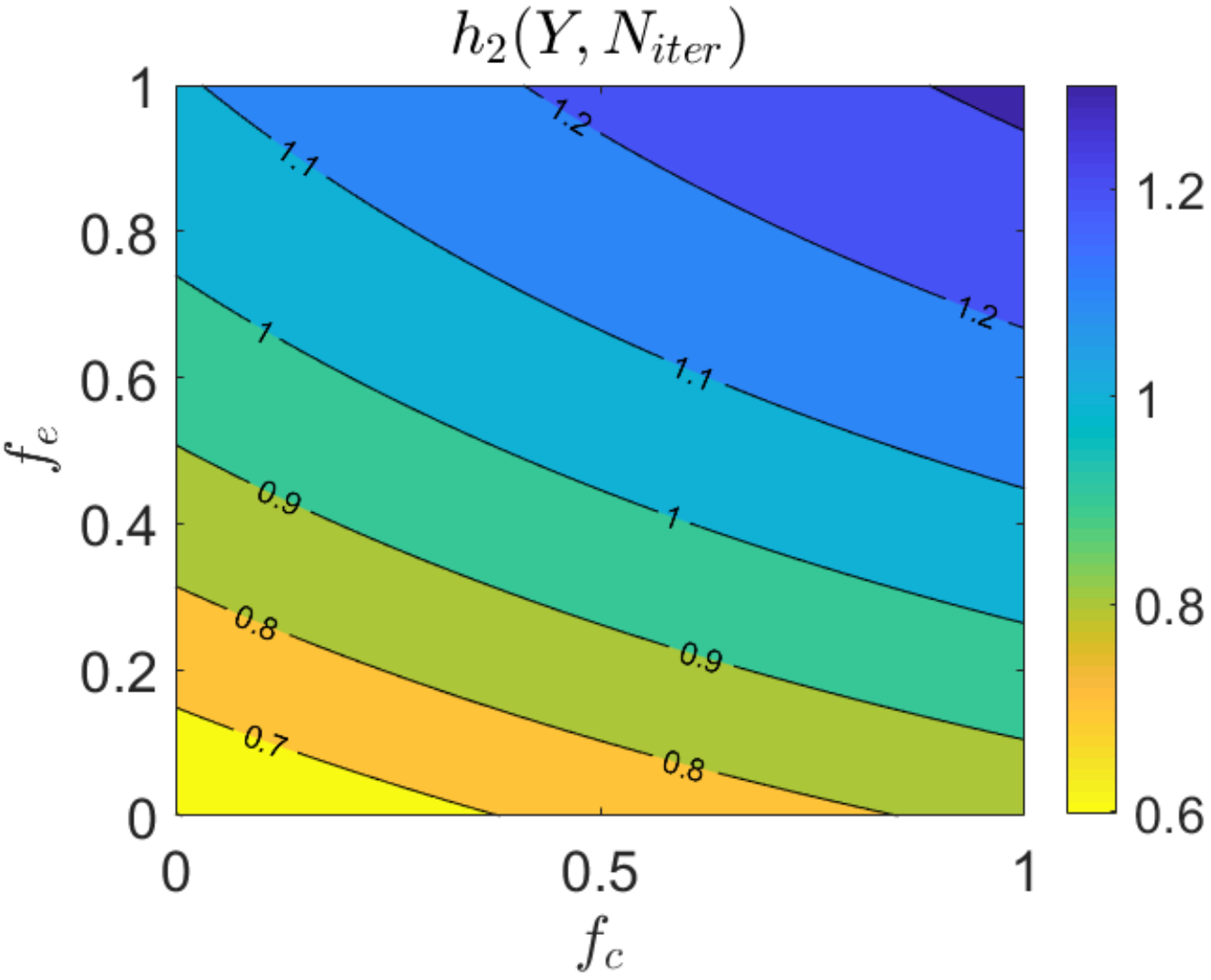}%
\hfil
}
}
\centerline{
\subfloat[$\itrn$=10]{\includegraphics[width=1.8in]{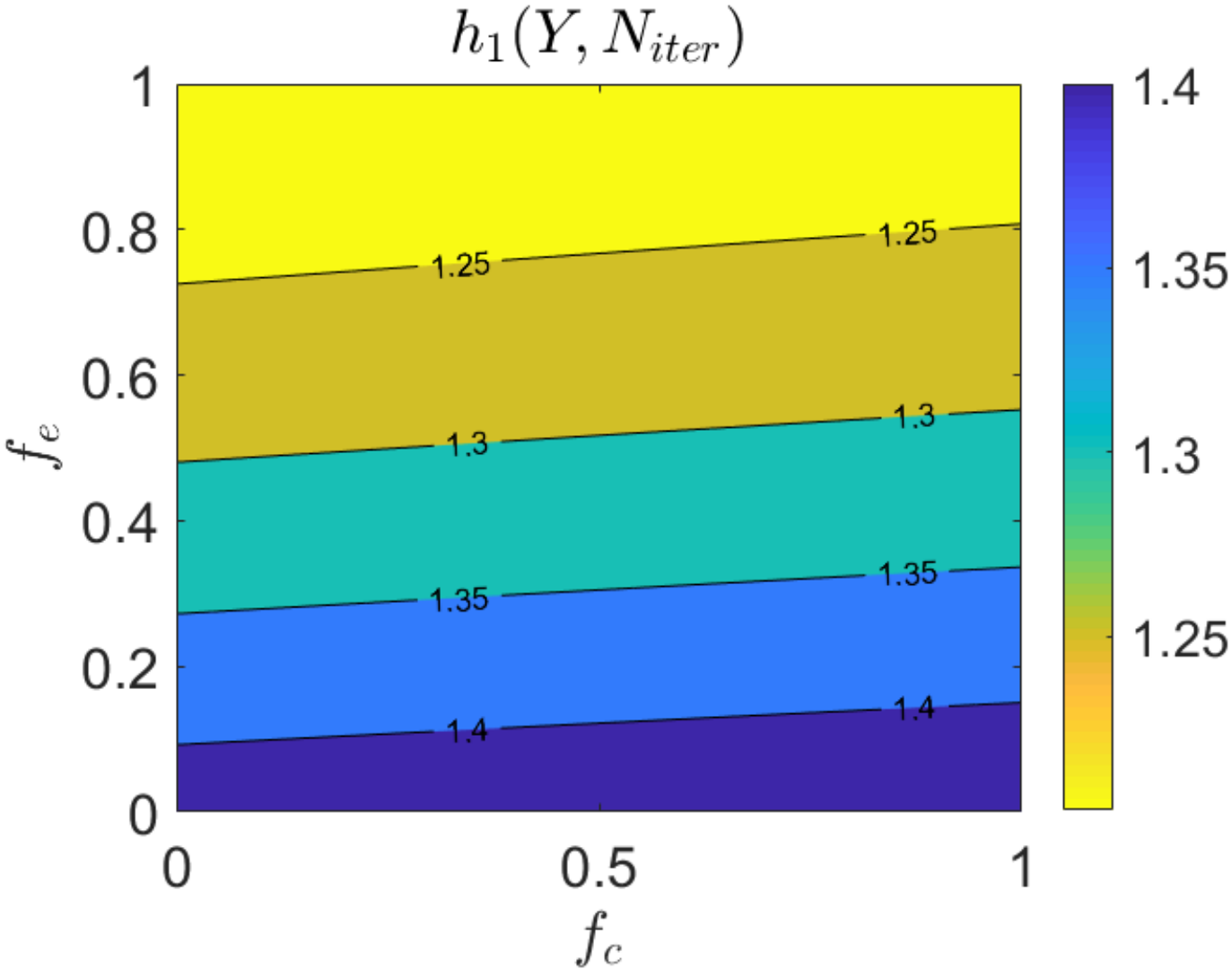}%
\hfil
}
\subfloat[$\itrn$=10]{\includegraphics[width=1.8in]{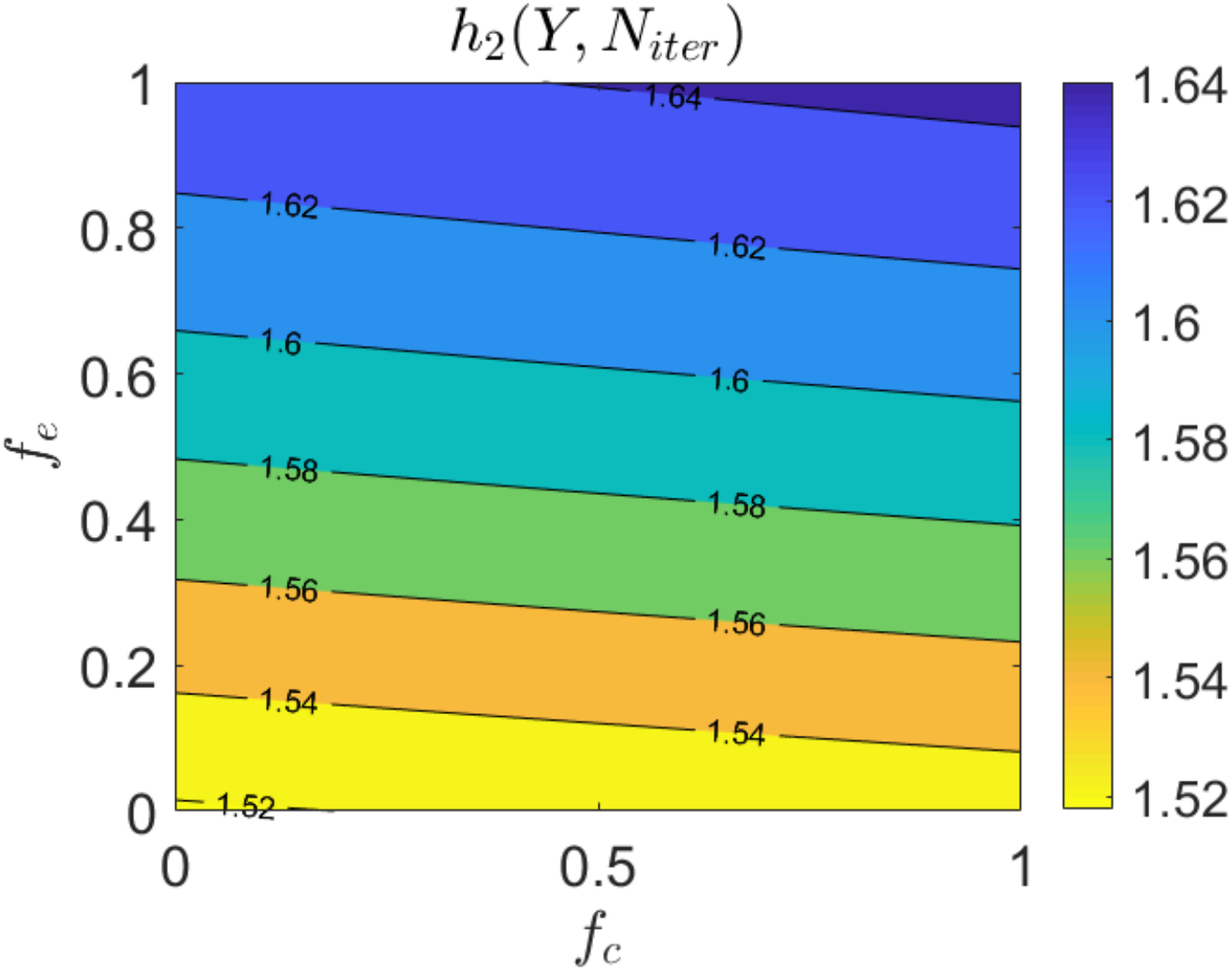}%
\hfil
}
}
\centerline{
\subfloat[$\itrn$=100]{\includegraphics[width=1.8in]{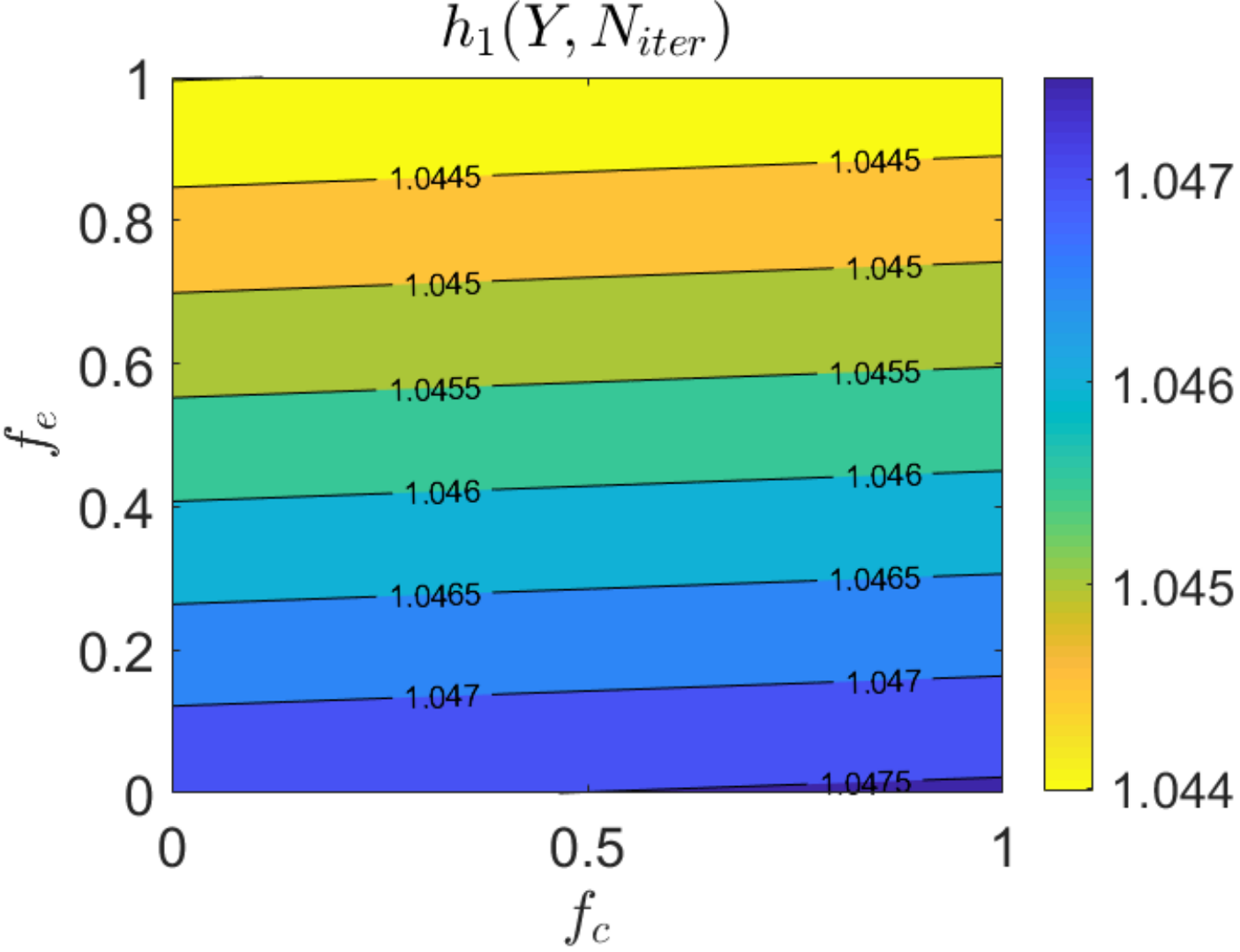}%
\hfil
}
\subfloat[$\itrn$=100]{\includegraphics[width=1.8in]{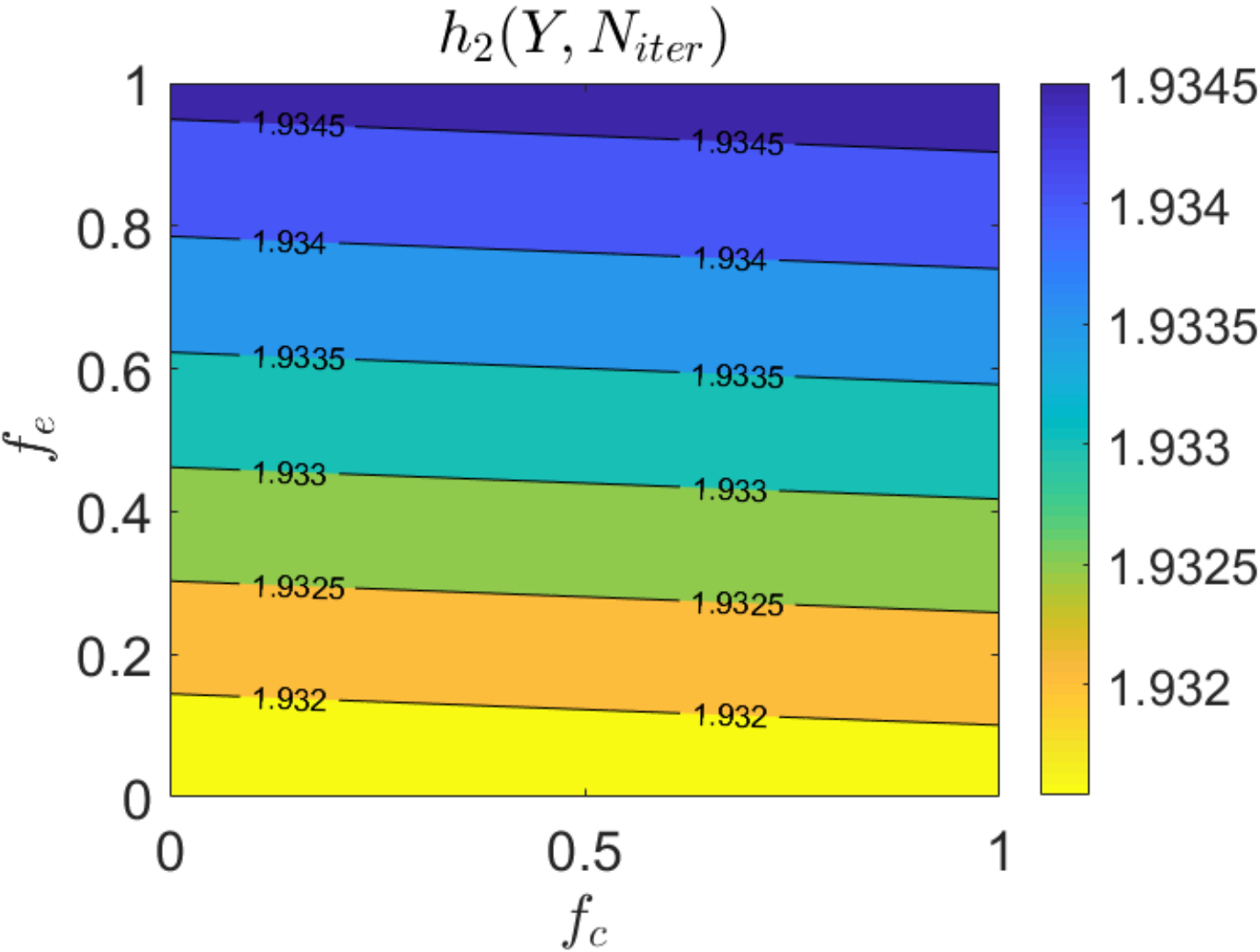}%
\hfil
}
}
\caption{Contour plots of $h_1(\boldsymbol{Y}, \itrn)=\frac{SBP_c(\boldsymbol{Y}^c, \itrn)}{SBP_e(\boldsymbol{Y}^e, \itrn)}$ ((a), (c) and (e)) and  $h_2(\boldsymbol{Y}, \itrn)=SBP_c(\boldsymbol{Y}^c, \itrn)+SBP_e(\boldsymbol{Y}^e, \itrn)$ ((b), (d) and (f))  with $\mathbf{r}=(5, 1)$.}
\label{Fig.SBP}
\end{figure}

To take a closer look at the proposed SBP, we consider an example HE-MOP with $\mathbf{r}=(5, 1)$ having a cheap and an expensive objective functions (denoted as $f^c$ and $f^e$, respectively). Contour plots of $h_1(\boldsymbol{Y}, \itrn)=\frac{SBP_c(\boldsymbol{Y}^c, \itrn)}{SBP_e(\boldsymbol{Y}^e, \itrn)}$ and $h_2(\boldsymbol{Y}, \itrn)=SBP_c(\boldsymbol{Y}^c, \itrn)+SBP_e(\boldsymbol{Y}^e, \itrn)$ with respect to $f^c$, $f^e$ and $\itrn$ are given in Fig. \ref{Fig.SBP}. For $\itrn=1$ shown in Fig. \ref{Fig.SBP}(a), the penalty on $f^e$ is always smaller than that on $f^c$. This indicates that including such a penalty term into an acquisition function will guide the selection of new samples towards exploring the expensive objective function. Hence, the proposed algorithm will encourage the search towards $f^e$ and mitigate the intrinsic search bias due to heterogeneous objectives. For $\itrn=1$ shown in Fig. \ref{Fig.SBP}(b), the search bias penalty varies a lot in different regions of the objective space: there is a significant difference with respect to the penalty between the regions with smaller objective values and the regions with lager objective values. As the optimization proceeds, the difference between the cheap and expensive objectives will gradually shrink, as illustrated in Fig. \ref{Fig.SBP}(c) and \ref{Fig.SBP}(d), indicating a decreasing influence on the search bias. Finally, the penalty is almost equal across the whole objective space, as shown in Figs. \ref{Fig.SBP}(e) and \ref{Fig.SBP}(f). This allows SBP-BO to find out a set of satisfying solutions that cover the whole Pareto front.

\section{Experimental Studies}
\subsection{Experimental Settings}
1) \emph{Test Problems}: Although there are no test problems available that have inherently heterogeneous objectives, any existing multi-/many-objective benchmark problem can be adopted as HE-MOPs/MaOPs, assuming the evaluation times of the objectives are substantially different. Therefore, we have selected three widely used test suites of scalable multi-objective test problems, i.e., the DTLZ \cite{deb2002scalable} and WFG \cite{huband2006review} test suites, and extend them to simulate HE-MOPs and HE-MaOPs. For all the test instances used in the experimental studies, the number of decision variables is set to 10.

2) \emph{Performance Indicators}: The modified inverted generational distance (IGD) \cite{zitzler2003performance}, the IGD$^+$ indicator \cite{ishibuchi2015modified}, is adopted as the performance indicator due to its computational efficiency and precise evaluation of the quality of the obtained non-dominated solutions in terms of convergence and diversity.  Let $Z=\left\{\boldsymbol{z}_{1}, \boldsymbol{z}_{2}, \ldots, \boldsymbol{z}_{|Z|}\right\}$ be a given reference solution set, where $|Z|$ is the number of reference solutions, and $A=\left\{\boldsymbol{a}_{1}, \boldsymbol{a}_{2}, \ldots, \boldsymbol{a}_{|A|}\right\}$ be an obtained approximation to the Pareto front, $IGD^{+}$ is calculated as follows:
\begin{equation}
IGD^{+}(A,Z)=\frac{1}{|Z|} \sum_{j=1}^{|Z|} \min _{\boldsymbol{a}_{i} \in A} d^{+}\left(\boldsymbol{a}_{i}, \boldsymbol{z}_{j}\right).
\end{equation}
The distance $d^{+}$ is the distance between a reference solution $\boldsymbol{z}=(z_1,z_2,\cdots,z_m)$ and an objective vector $\boldsymbol{a}=(a_1,\cdots,a_m)$. Here, $m$ is the number of objectives. $d^{+}$ is defined as:
\begin{equation}
d^{+}(\boldsymbol{a}, \boldsymbol{z})=\sqrt{\sum_{k=1}^{m}\left(\max \left\{z_{k}-a_{k}, 0\right\}\right)^{2}}
\end{equation}
The smaller the IGD$^{+}$ value is, the better the quality of the non-dominated solution set.

Each algorithm under comparison is performed on each benchmark problem for 20 independent runs. The Wilcoxon rank sum test at a significance level of 0.05 is adopted to compare the results obtained by SBP-BO and other algorithms under comparison. To reduce the probability of making a type I error, the Holm-Bonferroni correction is adopted. The corresponding statistical results are presented in Tables \ref{Tab.1}-\ref{Tab.4} and Tables SI-SVI in the Supplementary material, where symbols "(+)", "(--)", and "($\approx)$" indicate that the compared algorithm performs significantly better than, significantly worse than, or as well as the proposed algorithm, respectively. Note that we use notation `S' to indicate tables and figures in the Supplementary materials in order to avoid confusion. 

3) \emph{Algorithms Under Comparison}: To the best of authors knowledge, SBP-BO is the first algorithm designed for addressing HE-MOPs and HE-MaOPs. For comparison, one of state-of-the-art heterogeneity-handling methods, HK-RVEA \cite{chugh2018hkrvea}, is slightly adapted to the proposed problem setting. Specifically, the SOEA in HK-RVEA optimizes each cheap objective using the different number of additional evaluations in the initialization. The genetic operators are used to generate additional samples for each cheap objectives while waiting for the expensive evaluation on new samples.As existing surrogate-assisted heterogeneity handling methods, e.g., T-SAEA \cite{XiluWang2020TSAEA}, Tr-SAEA \cite{wang2021transfer} and TC-SAEA \cite{WangTC-SAEA2021}, cannot address HE-MOPs or HE-MaOPs with more than two objectives, we compared them with the proposed algorithm on heterogeneous bi-objective optimization problems reported in \cite{WangTC-SAEA2021}. Note that we did not include non-surrogate assisted algorithms for two reasons: 1) it has been shown that surrogate-assisted methods outperform non-surrogate assisted methods; 2) most existing non-surrogate assisted methods work only for bi-objective problems with one fast and one slow objective. Since SBP-BO is based on GP assisted RVEA, a representative GP-assisted MOEA, K-RVEA \cite{chugh2018surrogate}, is also adopted as a surrogate assisted \emph{Waiting} method to examine the performance of the proposed algorithm.

To further investigate the efficacy of the proposed surrogate ensemble and the search bias penalized acquisition function, we perform the following ablation studies:
\begin{itemize}
\item \textbf{Ablations on the use of additional data of the fast objectives}: In order to confirm that the GP ensemble for the fast objectives contributes to performance improvement, we consider several variants of SBP-BO for utilizing the extra data: i) SBP-BO-R: to limit the computational time for re-training GPs, a fixed number of evaluated samples are randomly selected from all available data to train a GP for each objective; ii) SBP-BO-C: to effectively re-train a GP for each objective, SBP-BO-C first clusters all available data in the decision space, and then selects one sample from each cluster, constructing a small yet diverse subset. iii) SBP-NoGP$^c$: to investigate whether the use of surrogate models on the relatively cheap objectives is useful or not, SBP-NoGP$^c$ assumes that the evaluation of the relatively cheap objectives $f_{j}^{c}, j=1,\cdots, p$ can be calculated instantly. Hence, there is no surrogate constructed on the cheap objectives, and the cheap objective values in line 7 in Algorithm 1 is provided by the true cheap evaluation.
\item \textbf{Ablations on the proposed SBP}: To test the effect of the proposed SBP on the optimization of HE-MOPs/HE-MaOPs, we remove the SBP from SBP-BO. Therefore, the acquisition function is degrades into the adaptive acquisition function (AAF), introducing the corresponding variant, BO-AAF.
\end{itemize}
4) \emph{Parameter settings}: We use RVEA as the multi-objective optimizer and a real-coded genetic algorithm that uses the simulated binary crossover and polynomial mutation is employed as the single-objective optimizer. In addition, we use the DACE toolbox \cite{lophaven2002dace} to construct the GP models.

All experiments are performed in MATLAB R2019a on an Intel Core i7-8750H with 2.21 GHz CPU. The parameter settings used in the experiments are summarized as follows:
\begin{itemize}
\item The initial population size for all the compared algorithms is set to $11d-1$ \cite{knowles2006parego} where $d$ is the number of decision variables. The initial population is generated by LHS.
\item The maximum number of generations before updating the GPs ($w_{max}$) is set to 20.
\item The number of new solutions selected for evaluations on all objectives at each BO iteration is set to $u=3$.
\item The maximum number of function evaluations for the slow objectives ($FE_{max}^{e}$) is set to 200 for heterogeneous bi-objective problems, and 300 for HE-MOPs/MaOPs.
\end{itemize}

\subsection{Experimental Results}
1) \emph{Comparison with state-of-the-art methods}: Each algorithm is performed on the test problems with $m=3, 5, 10$ objectives, respectively. For simplicity, only one slow objective is included while all other objectives are considered as fast objectives with the same evaluation time, resulting in $\mathbf{r}=(r^{c}, r^{c},\cdots,r^{c},1)$. Tables \ref{Tab.1}-\ref{Tab.2} present the experimental results in terms of the IGD$^{+}$ values obtained on each test instance with $r^c=5$ and $r^c=10$, respectively. Hence, the threshold $r_\m{thres}$ is set to 1 for simplicity, and we will test different $\mathbf{r}$ and $r_\m{thres}$ in the next subsection.  

Firstly, the results presented in Table \ref{Tab.1} show that the proposed SBP-BO significantly outperforms K-RVEA and HK-RVEA on 27 and 29 out of 48 test instances, respectively, indicating the effectiveness of the proposed strategies for handling HE-MOPs/HE-MaOPs. Secondly, it is interesting to see that both K-RVEA and HK-RVEA show significantly better performance than SBP-BO on DTLZ7. A possible explanation for this might be that in K-RVEA and HK-RVEA a fixed reference set is used to evaluate whether diversity or convergence should be prioritized in the selection of new samples. In this way, the exploration can be guaranteed, so that K-RVEA and HK-RVEA are able to account for the disconnected Pareto front of DTLZ7. According to the results on WFG1-WFG9 summarized in Table \ref{Tab.1}, SBP-BO achieves the best performance in terms of IGD$^{+}$ metric on 20 test instances, followed by K-RVEA with 5 best results. Note that K-RVEA significantly outperforms the proposed algorithm on WFG1 with $m=3$ and WFG9 with $m=10$. Recall that both WFG1 and WFG9 are difficult for optimization algorithms to achieve a good diversity. WFG1 is designed by using the most complex transformation function to add complexity to a underlying problem, making it hard an optimization algorithm to converge to the true Pareto front. Similarly, WFG9 features a troublesome transformation function and is also a multi-model and nonseparable problem. Lastly, by comparing K-RVEA using the \emph{Waiting} method with HK-RVEA and SBP-BO we can confirm the effectiveness of using a GP ensemble and the proposed acquisition function.  While HK-RVEA generally shows similar performance with RVEA, SBP-BO exhibits better performance than RVEA on most test problems. 

The results presented in Table \ref{Tab.2} are consistent with those in Table \ref{Tab.1}, further supporting the benefit of using a GP ensemble to make use of the additional data on the fast objectives and the search bias penalized acquisition function. It is noteworthy that with $r^c$ increasing from 5 to 10, the proposed algorithm maintains its advantage for solving HE-MOPs and HE-MaOPs by properly making use of the information obtained from the optimization of the fast objectives. Note, however, that the proposed algorithm may fail to maintain a good diversity of the obtained solutions on some test problems, such as DTLZ7 and WFG9. Hence, although the proposed SBP acquisition function promotes the search towards the slow objectives, but it may lead to a poor diversity of the solutions on some problems.

To gain a  deeper insight into the quality of the final solution sets obtained by K-RVEA, HK-RVEA and SBP-BO, Figs. \ref{SolutionM3} and Figs. S1-S2 in the Supplementary materials show the nondominated solution set with the median IGD$^{+}$ value among 20 runs obtained by K-RVEA, HK-RVEA and SBP-BO on DTLZ2, DTLZ4, WFG2 and WFG6 with $m=3, 5, 10$, respectively. Fig. \ref{SolutionM3} demonstrate that the proposed SBP-BO shows promising performance in terms of both convergence and diversity on the selected three-objective test problems, compared with K-RVEA and HK-RVEA. For example, it is clear that SBP-BO finds a better approximation of the true Pareto front on DTLZ2 when compared with the other algorithms, indicating its good balance between diversity and convergence. It is worth noting that while HK-RVEA covers a small part of the true Pareto front, SBP-BO is able to achieve a set of well distributed solutions. This observation further supports the advantage of the proposed SBP for reducing the search bias towards to the cheap objectives. Regarding HE-MaOPs, similar observations can be made, as illustrated in Figs. S1-S2, where the solutions on the estimated Pareto front are shown in the parallel coordinate plots. These observations can be explained from the perspective of the use of additional data and the reduction of search bias by the SBP acquisition function.

To explore the performance of SBP-BO as the evolution proceeds, the IGD$^+$ values obtained by each algorithm over the number of real fitness evaluations (FEs) on test problems with $r^c=\left \{5,10\right \}$ and the corresponding statistically significant differences are summarised in Tables SI-SII, respectively, in the Supplementary materials. As can be seen from the tables, the proposed SBP-BO shows significantly better performance than K-RVEA on DTLZ2, WFG3, WFG5 and WFG6 with $FE^e=150$, indicating the fast convergence of SBP-BO. Although HK-RVEA shows better performance than K-RVEA, SBP-BO can significantly outperform HK-RVEA on DTLZ2, WFG4, WFG6 and WFG7 with $FE^e=200$. Subsequently, Figs. S3-S4 plot the boxplots of the IGD$^+$ values obtained by each algorithm on DTLZ2, DTLZ4, WFG2 and WFG6 with different number of FEs over 20 runs, confirming the fast convergence achieved by SBO-BO. Moreover, we demonstrate the search process of each algorithm in terms of IGD$^+$ values in Figs. S5-S6, where the error bars indicate the variance of IGD$^+$ values over 20 runs. According to Figs. S5-S6, similar conclusion can be made.

\begin{table}[]
\caption{Mean (Standard Deviation) IGD$^{+}$ values obtained by K-RVEA, HK-RVEA, and SBP-BO with $FE_{max}^{e}=300$ and $r = (r^c,\dots,r^c,1)$ where $r^c=5$}
\label{Tab.1}
\renewcommand{\arraystretch}{1}
\resizebox{0.47\textwidth}{!}{
\begin{tabular}{llccl}
\toprule

Problem                    & m  & K-RVEA                                       & HK-RVEA                                      & SBP-BO                                 \\   \midrule[0.3pt]
                           & 3  & 7.88e+1 (1.16e+1) --                         & 9.07e+1 (1.42e+1) --                         & \cellcolor[HTML]{C0C0C0}5.35e+1 (2.10e+1) \\
                           & 5  & 3.96e+1 (1.21e+1) --                        & 5.41e+1 (8.50e+0) --                         & \cellcolor[HTML]{C0C0C0}2.45e+1 (8.88e+0) \\
                         
\multirow{-3}{*}{DTLZ1}    & 10 & 3.31e-1 (1.42e-1) --                         & 1.97e-1 (8.40e-2) $\approx$                        & \cellcolor[HTML]{C0C0C0}1.58e-1 (2.31e-2) \\ \midrule[0.3pt]
                           & 3  & 7.52e-2 (1.02e-2) --                         & 5.98e-2 (1.05e-2) --                        & \cellcolor[HTML]{C0C0C0}4.15e-2 (3.27e-3) \\
                           & 5  & 1.79e-1 (1.88e-2) --                         & 1.38e-1 (9.34e-3) --                         & \cellcolor[HTML]{C0C0C0}1.06e-1 (7.07e-3) \\
\multirow{-3}{*}{DTLZ2}    & 10 & 2.38e-1 (1.32e-2) --                         & 2.73e-1 (4.02e-2) --                         & \cellcolor[HTML]{C0C0C0}2.04e-1 (9.70e-3) \\ \midrule[0.3pt]
                           & 3  & 2.13e+2 (3.73e+1) $\approx$                         & 2.35e+2 (3.55e+1) --                         & \cellcolor[HTML]{C0C0C0}1.72e+2 (4.57e+1) \\
                           & 5  & 1.25e+2 (3.49e+1) --                         & 1.64e+2 (3.70e+1) --                         & \cellcolor[HTML]{C0C0C0}9.17e+1 (2.53e+1) \\
\multirow{-3}{*}{DTLZ3}    & 10 & 7.96e-1 (3.52e-1) $\approx$                         & 8.05e-1 (3.58e-1) $\approx$                         & \cellcolor[HTML]{C0C0C0}6.12e-1 (1.48e-1) \\ \midrule[0.3pt]
                           & 3  & 2.59e-1 (7.57e-2) $\approx$                      & 3.41e-1 (1.30e-1) --                       & \cellcolor[HTML]{C0C0C0}2.31e-1 (1.29e-1) \\
                           & 5  & \cellcolor[HTML]{C0C0C0}2.73e-1 (5.59e-2) $\approx$ & 3.66e-1 (6.29e-2) --                        & \cellcolor[HTML]{FFFFFF}2.89e-1 (9.04e-2) \\
\multirow{-3}{*}{DTLZ4}    & 10 & 2.58e-1 (1.77e-2) $\approx$                        & 2.68e-1 (3.08e-2) $\approx$                       & \cellcolor[HTML]{C0C0C0}2.57e-1 (2.60e-2) \\\midrule[0.3pt]
                           & 3  & 6.76e-2 (1.15e-2) --                         & 6.52e-2 (9.38e-3) --                         & \cellcolor[HTML]{C0C0C0}3.08e-2 (2.87e-3) \\
                           & 5  & 2.90e-2 (6.77e-3) --                         & \cellcolor[HTML]{C0C0C0}1.90e-2 (3.57e-3) $\approx$ & 1.89e-2 (3.55e-3)                         \\
\multirow{-3}{*}{DTLZ5}    & 10 & \cellcolor[HTML]{C0C0C0}6.22e-3 (7.70e-4) $\approx$  & 7.31e-3 (1.43e-3) $\approx$                         & 7.22e-3 (9.88e-4)                         \\ \midrule[0.3pt]
                           & 3  & \cellcolor[HTML]{C0C0C0}3.03e+0 (6.09e-1) $\approx$ & 3.15e+0 (4.09e-1) $\approx$                        & 3.05e+0 (4.83e-1)                         \\
                           & 5  & \cellcolor[HTML]{C0C0C0}1.77e+0 (3.28e-1) $\approx$ & 1.90e+0 (3.09e-1) $\approx$                         & 1.83e+0 (4.89e-1)                         \\
\multirow{-3}{*}{DTLZ6}    & 10 & 3.85e-2 (7.26e-3) --                         & 2.69e-2 (8.73e-3) --                         & \cellcolor[HTML]{C0C0C0}2.47e-2 (7.67e-3) \\ \midrule[0.3pt]
                           & 3  & 1.09e-1 (2.63e-2) + & \cellcolor[HTML]{C0C0C0}6.64e-2 (1.04e-2) +                         & 1.65e-1 (4.89e-2)                         \\
                           & 5  & 4.79e-1 (2.94e-1) +                         & \cellcolor[HTML]{C0C0C0}3.36e-1 (7.50e-2) + & 9.98e-1 (3.48e-1)                         \\
\multirow{-3}{*}{DTLZ7}    & 10 & \cellcolor[HTML]{C0C0C0}9.08e-1 (3.88e-2) + & 9.36e-1 (2.65e-2) +                       & 9.54e-1 (1.70e-1)                         \\ \midrule[0.3pt]
                           & 3  & \cellcolor[HTML]{C0C0C0}1.64e+0 (4.10e-2) + & 1.74e+0 (1.15e-1) $\approx$                         & 1.76e+0 (1.34e-1)                         \\
                           & 5  & \cellcolor[HTML]{C0C0C0}2.19e+0 (7.70e-2) $\approx$ & 2.22e+0 (8.27e-2) $\approx$                         & 2.20e+0 (6.82e-2)                         \\
\multirow{-3}{*}{WFG1}     & 10 & 2.81e+0 (1.32e-1) $\approx$                         & 2.81e+0 (1.26e-1) $\approx$                         & \cellcolor[HTML]{C0C0C0}2.79e+0 (1.46e-1) \\ \midrule[0.3pt]
                           & 3  & 2.91e-1 (2.74e-2) --                         & 2.20e-1 (2.22e-2) --                         & \cellcolor[HTML]{C0C0C0}1.80e-1 (2.81e-2) \\
                           & 5  & 3.93e-1 (6.16e-2) --                        & 2.82e-1 (3.07e-2) --                         & \cellcolor[HTML]{C0C0C0}2.10e-1 (2.89e-2) \\
\multirow{-3}{*}{WFG2}     & 10 & 3.96e-1 (1.38e-1) --                         & 3.98e-1 (1.45e-1) --                         & \cellcolor[HTML]{C0C0C0}3.11e-1 (8.55e-2) \\ \midrule[0.3pt]
                           & 3  & 4.12e-1 (5.12e-2) --                         & 4.53e-1 (5.97e-2) --                         & \cellcolor[HTML]{C0C0C0}2.10e-1 (3.02e-2) \\
                           & 5  & 4.29e-1 (8.76e-2) --                         & 3.43e-1 (3.57e-2) $\approx$                         & \cellcolor[HTML]{C0C0C0}3.48e-1 (8.35e-2) \\
\multirow{-3}{*}{WFG3}     & 10 & 5.59e-1 (6.59e-2) --                         & \cellcolor[HTML]{C0C0C0}5.35e-1 (7.26e-2) $\approx$ & 5.36e-1 (7.64e-2)                         \\ \midrule[0.3pt]
                           & 3  & 3.92e-1 (2.82e-2) --                         & 3.84e-1 (2.41e-2) --                         & \cellcolor[HTML]{C0C0C0}3.22e-1 (2.28e-2) \\
                           & 5  & 7.73e-1 (4.30e-2) --                         & 8.55e-1 (6.66e-2) --                         & \cellcolor[HTML]{C0C0C0}7.06e-1 (3.89e-2) \\
\multirow{-3}{*}{WFG4}     & 10 & 3.25e+0 (8.96e-1) $\approx$                         & 3.45e+0 (8.52e-1) --                         & \cellcolor[HTML]{C0C0C0}3.18e+0 (7.82e-1) \\ \midrule[0.3pt]
                           & 3  & 3.78e-1 (6.53e-2) $\approx$                         & \cellcolor[HTML]{C0C0C0}2.59e-1 (1.69e-2) $\approx$ & 2.99e-1 (8.32e-2)                         \\
                           & 5  & 7.96e-1 (6.80e-2) --                         & 7.12e-1 (3.17e-2) --                         & \cellcolor[HTML]{C0C0C0}6.59e-1 (4.20e-2) \\
\multirow{-3}{*}{WFG5}     & 10 & 2.12e+0 (4.55e-1) --                         & 1.97e+0 (4.58e-1) $\approx$                         & \cellcolor[HTML]{C0C0C0}1.78e+0 (3.69e-1) \\\midrule[0.3pt]
                           & 3  & 6.77e-1 (5.71e-2) --                         & 5.04e-1 (7.46e-2) --                         & \cellcolor[HTML]{C0C0C0}4.16e-1 (8.37e-2) \\
                           & 5  & 1.18e+0 (1.41e-1) --                         & 9.28e-1 (8.25e-2) --                         & \cellcolor[HTML]{C0C0C0}7.61e-1 (8.25e-2) \\
\multirow{-3}{*}{WFG6}     & 10 & 1.29e+0 (3.37e-2) --                         & 1.28e+0 (5.73e-2) --                         & \cellcolor[HTML]{C0C0C0}1.03e+0 (2.96e-2) \\ \midrule[0.3pt]
                           & 3  & 4.98e-1 (4.06e-2) --                         & 5.44e-1 (3.21e-2) --                         & \cellcolor[HTML]{C0C0C0}4.35e-1 (4.11e-2) \\
                           & 5  & 8.74e-1 (6.09e-2) --                         & 1.03e+0 (8.34e-2) --                         & \cellcolor[HTML]{C0C0C0}7.28e-1 (4.22e-2) \\
\multirow{-3}{*}{WFG7}     & 10 & 3.49e+0 (4.68e-1) $\approx$                         & 3.63e+0 (5.24e-1) --                         & \cellcolor[HTML]{C0C0C0}3.33e+0 (4.32e-1) \\ \midrule[0.3pt]
                           & 3  & 6.57e-1 (5.50e-2) --                         & 5.64e-1 (2.90e-2) --                         & \cellcolor[HTML]{C0C0C0}5.03e-1 (2.86e-2) \\
                           & 5  & 1.49e+0 (4.32e-2) --                         & 1.40e+0 (6.33e-2) --                         & \cellcolor[HTML]{C0C0C0}1.25e+0 (3.85e-2) \\
\multirow{-3}{*}{WFG8}     & 10 & \cellcolor[HTML]{C0C0C0}1.47e+0 (2.47e-1) $\approx$ & 3.57e+0 (1.10e+0) --                         & 1.58e+0 (6.99e-1)                         \\ \midrule[0.3pt]
                           & 3  & 5.76e-1 (7.03e-2) $\approx$                        & 5.72e-1 (1.15e-1) $\approx$                       & \cellcolor[HTML]{C0C0C0}5.51e-1 (1.36e-1) \\
                           & 5  & \cellcolor[HTML]{C0C0C0}1.13e+0 (2.54e-1) $\approx$ & 1.31e+0 (1.95e-1) --                         & 1.14e+0 (1.94e-1)                         \\
\multirow{-3}{*}{WFG9}     & 10 & \cellcolor[HTML]{C0C0C0}3.71e+0 (8.56e-1) + & 5.16e+0 (6.40e-1) $\approx$                        & 4.69e+0 (8.73e-1)                         \\ \midrule[0.3pt]
+/--/$\approx$ &    & 5/27/16                                    & 3/29/16                                     &                                           \\ \bottomrule
\end{tabular}}
\end{table}

\begin{table}[]
\caption{Mean (Standard Deviation) IGD$^{+}$ values obtained by K-RVEA, HK-RVEA, and SBP-BO with $FE_{max}^{e}=300$ and $r = (r^c,\dots,r^c,1)$ where $r^c=10$}
\label{Tab.2}
\renewcommand{\arraystretch}{1}
\resizebox{0.45\textwidth}{!}{
\begin{tabular}{llccl}
\toprule

Problem                    & m  & K-RVEA                                       & HK-RVEA                                      & SBP-BO                                 \\   \midrule[0.3pt]
                           & 3  & 7.88e+1   (1.16e+1) --                                              & 8.43e+1 (1.19e+1) --                                                & \cellcolor[HTML]{C0C0C0}5.44e+1   (1.63e+1) \\
                           & 5  & 3.96e+1   (1.21e+1) $\approx$                                               & 4.58e+1 (7.06e+0) --                                                & \cellcolor[HTML]{C0C0C0}2.68e+1 (1.01e+1)  \\
\multirow{-3}{*}{DTLZ1}    & 10 & 3.31e-1   (1.42e-1) --                                                & 2.26e-1 (7.29e-2) --          & \cellcolor[HTML]{C0C0C0}1.79e-1 (7.83e-2)   \\ \midrule[0.3pt]
                           & 3  & 7.52e-2   (1.02e-2) --                                               & 5.17e-2 (2.82e-3) --                                                & \cellcolor[HTML]{C0C0C0}3.82e-2 (2.19e-3)   \\
                           & 5  & 1.79e-1   (1.88e-2) --                                                & 1.35e-1 (1.06e-2) --                                                & \cellcolor[HTML]{C0C0C0}9.84e-2 (3.10e-3)                          \\
\multirow{-3}{*}{DTLZ2}    & 10 & 2.38e-1   (1.32e-2) --                                                & 2.70e-1 (3.23e-2) --                & \cellcolor[HTML]{C0C0C0}1.97e-1 (5.04e-3) \\ \midrule[0.3pt]
                           & 3  & 2.13e+2   (3.73e+1) $\approx$                                              & 2.38e+2 (2.49e+1) --                                                & \cellcolor[HTML]{C0C0C0}1.75e+2 (4.62e+1)                          \\
                           & 5  & 1.25e+2   (3.49e+1) --                                                & 1.50e+2 (2.45e+1) --                                                & \cellcolor[HTML]{C0C0C0}7.82e+1 (2.35e+1)   \\
\multirow{-3}{*}{DTLZ3}    & 10 & 7.96e-1   (3.52e-1) $\approx$                                   & 9.84e-1 (4.63e-1) $\approx$                        & \cellcolor[HTML]{C0C0C0}7.13e-1 (2.80e-1)                          \\ \midrule[0.3pt]
                           & 3  & 2.59e-1   (7.57e-2) $\approx$                                             & 3.28e-1 (1.29e-1) --                                              & \cellcolor[HTML]{C0C0C0}2.11e-1 (1.09e-1)                          \\
                           & 5  & \cellcolor[HTML]{C0C0C0}2.73e-1   (5.59e-2) $\approx$ & 3.27e-1 (7.94e-2) $\approx$                                               & 2.91e-1   (6.90e-2)                                                \\
\multirow{-3}{*}{DTLZ4}    & 10 & 2.58e-1   (1.77e-2) $\approx$                                              & 2.90e-1 (3.95e-2) --                                               & \cellcolor[HTML]{C0C0C0}2.43e-1 (1.47e-2)   \\ \midrule[0.3pt]
                           & 3  & 6.76e-2   (1.15e-2) --                                                & 7.24e-2 (1.26e-2) --                                                & \cellcolor[HTML]{C0C0C0}3.38e-2 (9.19e-3)                          \\
                           & 5  & 2.90e-2   (6.77e-3) --                                                & 2.22e-2 (4.81e-3) $\approx$                                               & \cellcolor[HTML]{C0C0C0}1.92e-2 (5.01e-3)                          \\
\multirow{-3}{*}{DTLZ5}    & 10 & \cellcolor[HTML]{C0C0C0} 6.22e-3   (7.70e-4) $\approx$ & 7.21e-3 (1.25e-3)    $\approx$                                             & 7.46e-3 (1.72e-3)                                                  \\ \midrule[0.3pt]
                           & 3  & \cellcolor[HTML]{C0C0C0}3.03e+0   (6.09e-1) $\approx$ & 3.15e+0 (4.28e-1) $\approx$                                                & 3.19e+0 (3.57e-1)                                                  \\
                           & 5  & 1.77e+0   (3.28e-1) $\approx$                                              & \cellcolor[HTML]{C0C0C0}1.74e+0 (2.30e-1) $\approx$  & 2.11e+0 (3.84e-1)                                                  \\
\multirow{-3}{*}{DTLZ6}    & 10 & 3.85e-2   (7.26e-3) --                                                & 2.63e-2 (8.52e-3) $\approx$                                              & \cellcolor[HTML]{C0C0C0}2.60e-2 (1.06e-2)                          \\ \midrule[0.3pt]
                           & 3  & 1.09e-1   (2.63e-2) $\approx$                                                 & \cellcolor[HTML]{C0C0C0}6.56e-2 (5.42e-3) + & 1.51e-1 (6.51e-2)                                                  \\
                           & 5  & 4.79e-1   (2.94e-1) +                                                & \cellcolor[HTML]{C0C0C0}3.22e-1 (4.49e-2) + & 8.96e-1 (3.07e-1)                                                  \\
\multirow{-3}{*}{DTLZ7}    & 10 & \cellcolor[HTML]{C0C0C0}9.08e-1   (3.88e-2) + & 9.29e-1 (4.40e-2) +                                                & 1.26e+0 (1.74e-1)                                                  \\ \midrule[0.3pt]
                           & 3  & \cellcolor[HTML]{C0C0C0}1.64e+0   (4.10e-2) + & 1.86e+0 (1.33e-1) $\approx$                                              & 1.81e+0 (1.79e-1)                                                  \\
                           & 5  & \cellcolor[HTML]{C0C0C0}2.19e+0   (7.70e-2) $\approx$ & 2.23e+0 (1.00e-1) $\approx$                                               & 2.22e+0 (1.20e-1)                                                  \\
\multirow{-3}{*}{WFG1}     & 10 & 2.81e+0   (1.32e-1) $\approx$                                                & 2.85e+0 (1.14e-1) $\approx$                                              & \cellcolor[HTML]{C0C0C0}2.80e+0   (9.46e-2) \\ \midrule[0.3pt]
                           & 3  & 2.91e-1   (2.74e-2) --                                                & 2.40e-1 (1.96e-2) --                                                & \cellcolor[HTML]{C0C0C0}1.59e-1 (1.46e-2)                          \\
                           & 5  & 3.93e-1   (6.16e-2) --                                                & 2.77e-1 (2.45e-2) --                                                & \cellcolor[HTML]{C0C0C0}2.01e-1 (2.42e-2)                          \\
\multirow{-3}{*}{WFG2}     & 10 & 3.96e-1   (1.38e-1) --                                                & 4.38e-1 (1.18e-1) --                                                & \cellcolor[HTML]{C0C0C0}2.29e-1 (2.21e-2)   \\ \midrule[0.3pt]
                           & 3  & 4.12e-1   (5.12e-2) --                                                & 4.48e-1 (3.98e-2) --                                                & \cellcolor[HTML]{C0C0C0}2.11e-1   (1.88e-2) \\
                           & 5  & 4.29e-1   (8.76e-2) --                                                & 3.61e-1 (6.32e-2) $\approx$                                               & \cellcolor[HTML]{C0C0C0}3.28e-1 (5.21e-2)  \\
\multirow{-3}{*}{WFG3}     & 10 & 5.59e-1   (6.59e-2) $\approx$                                               & \cellcolor[HTML]{C0C0C0}5.48e-1 (5.76e-2) $\approx$ & 5.58e-1 (7.30e-2)                                                  \\ \midrule[0.3pt]
                           & 3  & 3.92e-1   (2.82e-2) --                                                & 3.55e-1 (2.13e-2) --                                                & \cellcolor[HTML]{C0C0C0}2.90e-1 (1.64e-2)  \\
                           & 5  & 7.72e-1   (4.30e-2) $\approx$                                               & 8.12e-1 (5.71e-2) --                                                & \cellcolor[HTML]{C0C0C0}7.15e-1 (1.11e-1)                          \\
\multirow{-3}{*}{WFG4}     & 10 & 3.25e+0   (8.96e-1) $\approx$                                               & 3.58e+0 (1.10e+0) $\approx$                                              & \cellcolor[HTML]{C0C0C0}3.09e+0 (1.03e+0)   \\ \midrule[0.3pt]
                           & 3  & 3.78e-1   (6.53e-2) --                                                & \cellcolor[HTML]{C0C0C0}2.69e-1 (2.73e-2) $\approx$ & 2.72e-1 (5.38e-2)                                                  \\
                           & 5  & 7.96e-1   (6.80e-2) $\approx$                                               & \cellcolor[HTML]{C0C0C0}7.14e-1 (4.94e-2) $\approx$ & 7.24e-1 (8.37e-2)                                                  \\
\multirow{-3}{*}{WFG5}     & 10 & 2.12e+0   (4.55e-1) $\approx$                                                & 1.95e+0 (8.20e-1) $\approx$                                           & \cellcolor[HTML]{C0C0C0}1.70e+0 (3.72e-1)   \\ \midrule[0.3pt]
                           & 3  & 6.77e-1   (5.71e-2) --                                                & 5.21e-1 (1.03e-1) --                                               & \cellcolor[HTML]{C0C0C0}3.29e-1 (2.74e-2)  \\
                           & 5  & 1.18e+0   (1.41e-1) --                                                & 8.86e-1 (1.03e-1) --                                                & \cellcolor[HTML]{C0C0C0}6.53e-1 (5.58e-2)   \\
\multirow{-3}{*}{WFG6}     & 10 & 1.29e+0   (3.37e-2) --                                                & 1.30e+0 (3.55e-2) --                                                & \cellcolor[HTML]{C0C0C0}1.03e+0   (2.36e-2) \\ \midrule[0.3pt]
                           & 3  & 4.98e-1   (4.06e-2) --                                                & 5.31e-1 (2.92e-2) --                                                & \cellcolor[HTML]{C0C0C0}4.12e-1 (2.88e-2)                          \\
                           & 5  & 8.74e-1   (6.09e-2) --                                                & 9.51e-1 (7.23e-2) --                                                & \cellcolor[HTML]{C0C0C0}7.30e-1 (6.09e-2)                          \\
\multirow{-3}{*}{WFG7}     & 10 & \cellcolor[HTML]{C0C0C0} 3.49e+0   (4.68e-1) $\approx$ & 3.79e+0 (7.49e-1) $\approx$                                             & 3.52e+0 (6.18e-1)                                                  \\ \midrule[0.3pt]
                           & 3  & 6.57e-1   (5.50e-2) --                                                & 5.71e-1 (4.07e-2) --                                                & \cellcolor[HTML]{C0C0C0}4.94e-1   (2.85e-2) \\
                           & 5  & 1.49e+0   (4.32e-2) --                                                & 1.41e+0 (5.06e-2) --                                                & 1.28e+0 (4.85e-2)                                                  \\
\multirow{-3}{*}{WFG8}     & 10 & \cellcolor[HTML]{C0C0C0}1.47e+0   (2.47e-1) $\approx$ & 4.45e+0 (7.78e-1) --                                                & \cellcolor[HTML]{C0C0C0}1.88e+0 (7.38e-1)   \\ \midrule[0.3pt]
                           & 3  & 5.76e-1   (7.03e-2) $\approx$                                         & 5.81e-1 (1.00e-1) $\approx$                                        & \cellcolor[HTML]{C0C0C0}4.89e-1 (1.38e-1)   \\
                           & 5  & \cellcolor[HTML]{C0C0C0}1.13e+0   (2.54e-1) $\approx$ & 1.24e+0 (1.76e-1) $\approx$                                               & 1.19e+0 (2.00e-1)                                                  \\
\multirow{-3}{*}{WFG9}     & 10 & \cellcolor[HTML]{C0C0C0}3.71e+0   (8.56e-1) $\approx$ & 4.84e+0 (4.41e-1) $\approx$                                             & 4.66e+0 (8.77e-1)                                                  \\ \midrule[0.3pt]
+/--/$\approx$ &    & {\color[HTML]{000000} 3/23/22}                                       & 3/25/20                                                           & \multicolumn{1}{l}{}                                                \\ \bottomrule
\end{tabular}}
\end{table}

\begin{figure*}[!htbp]
\centerline{
\subfloat[True PF]{\includegraphics[width=1.8in]{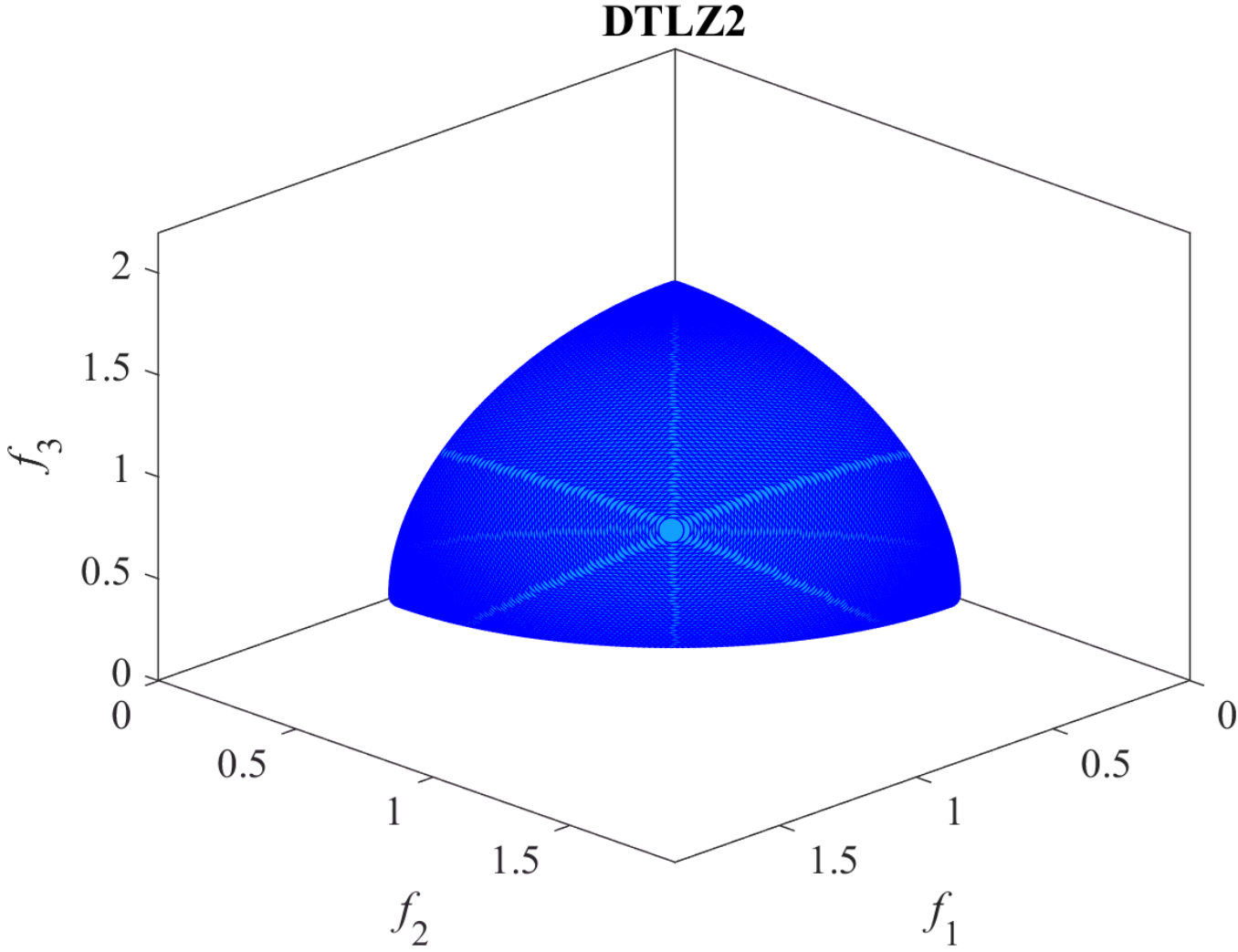}%
\hfil
}
\subfloat[K-RVEA]{\includegraphics[width=1.8in]{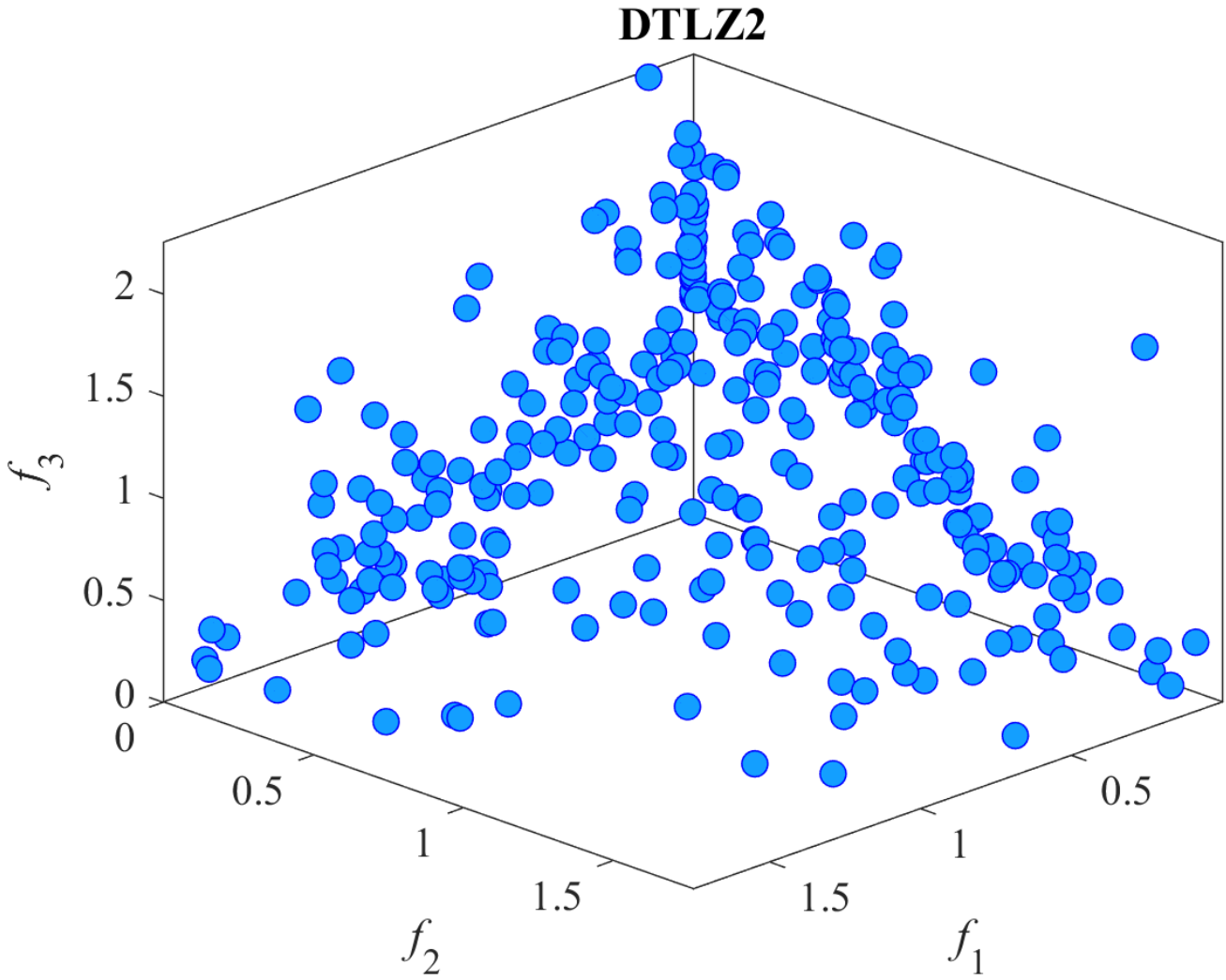}%
\hfil
}
\subfloat[HK-RVEA]{\includegraphics[width=1.8in]{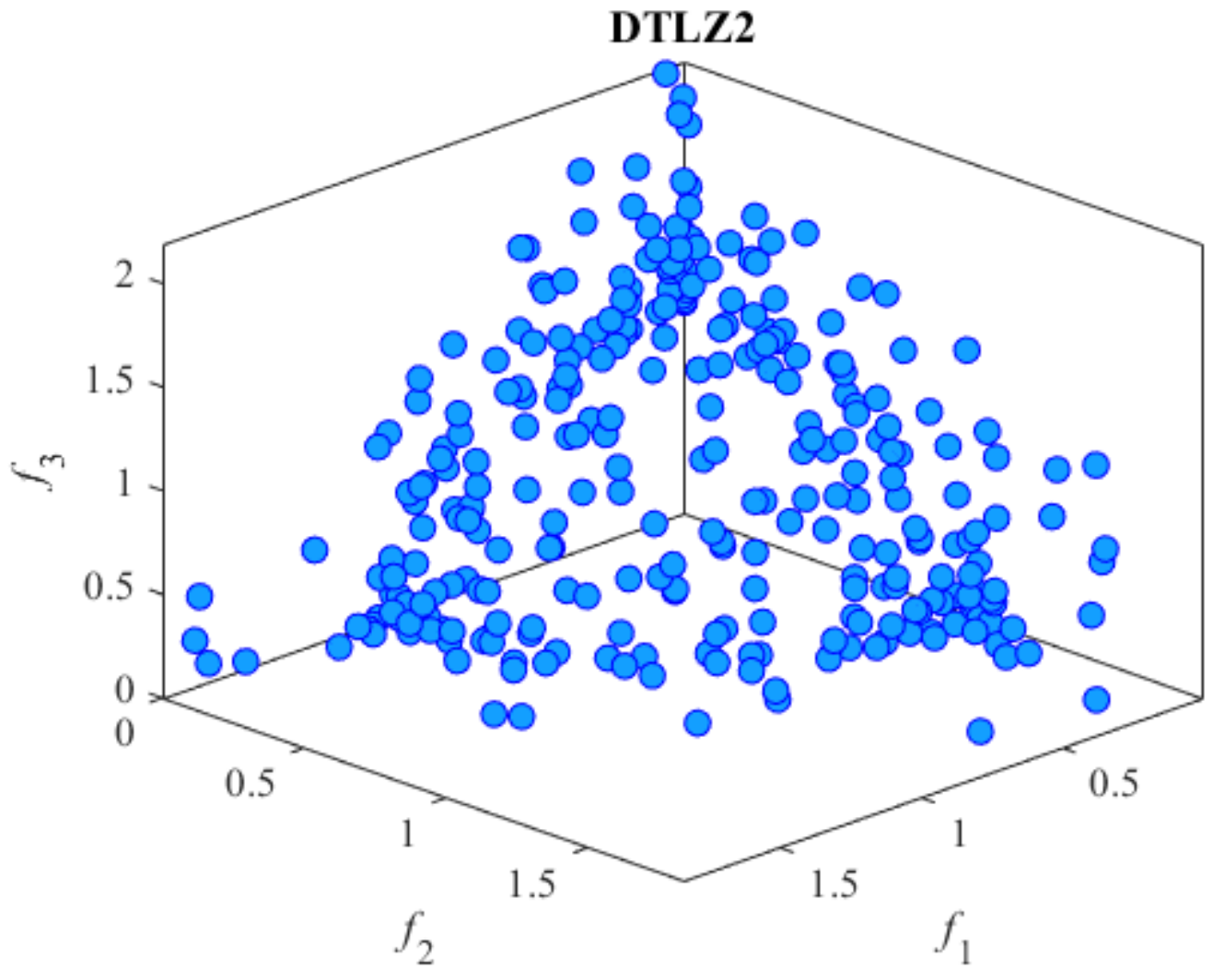}%
\hfil
}
\subfloat[SBP-BO]{\includegraphics[width=1.8in]{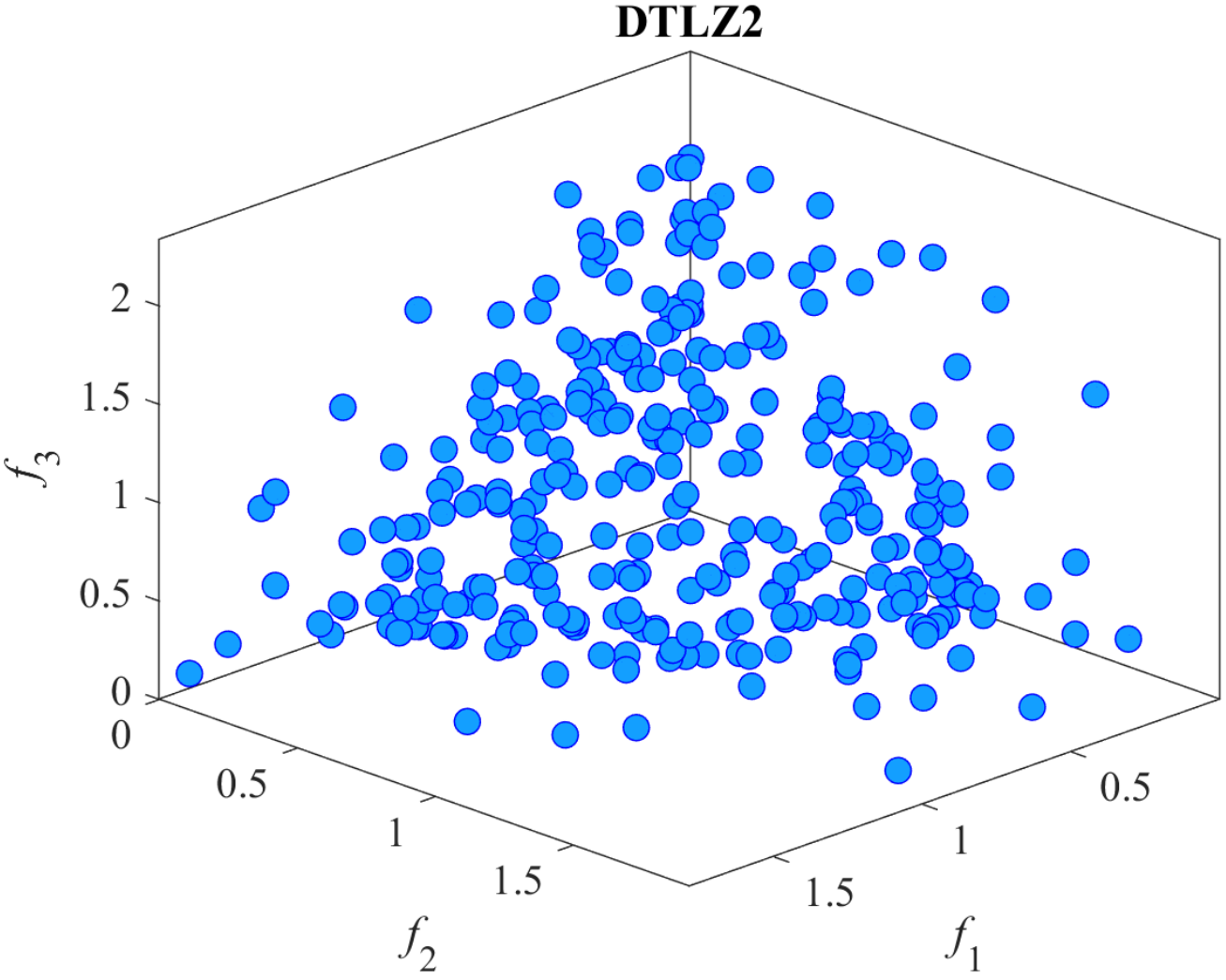}%
\hfil
}
}
\centerline{
\subfloat[True PF]{\includegraphics[width=1.8in]{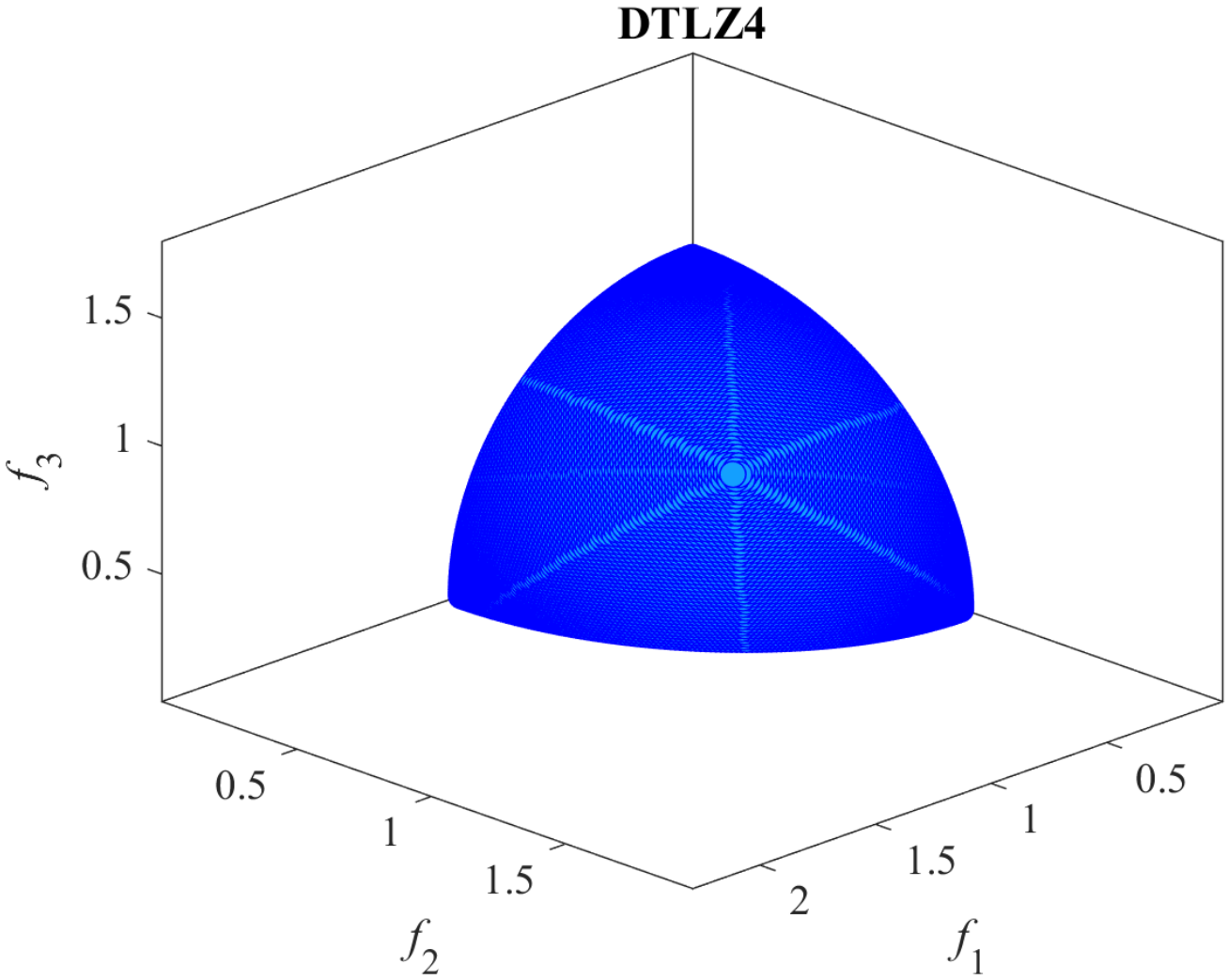}%
\hfil
}
\subfloat[K-RVEA]{\includegraphics[width=1.8in]{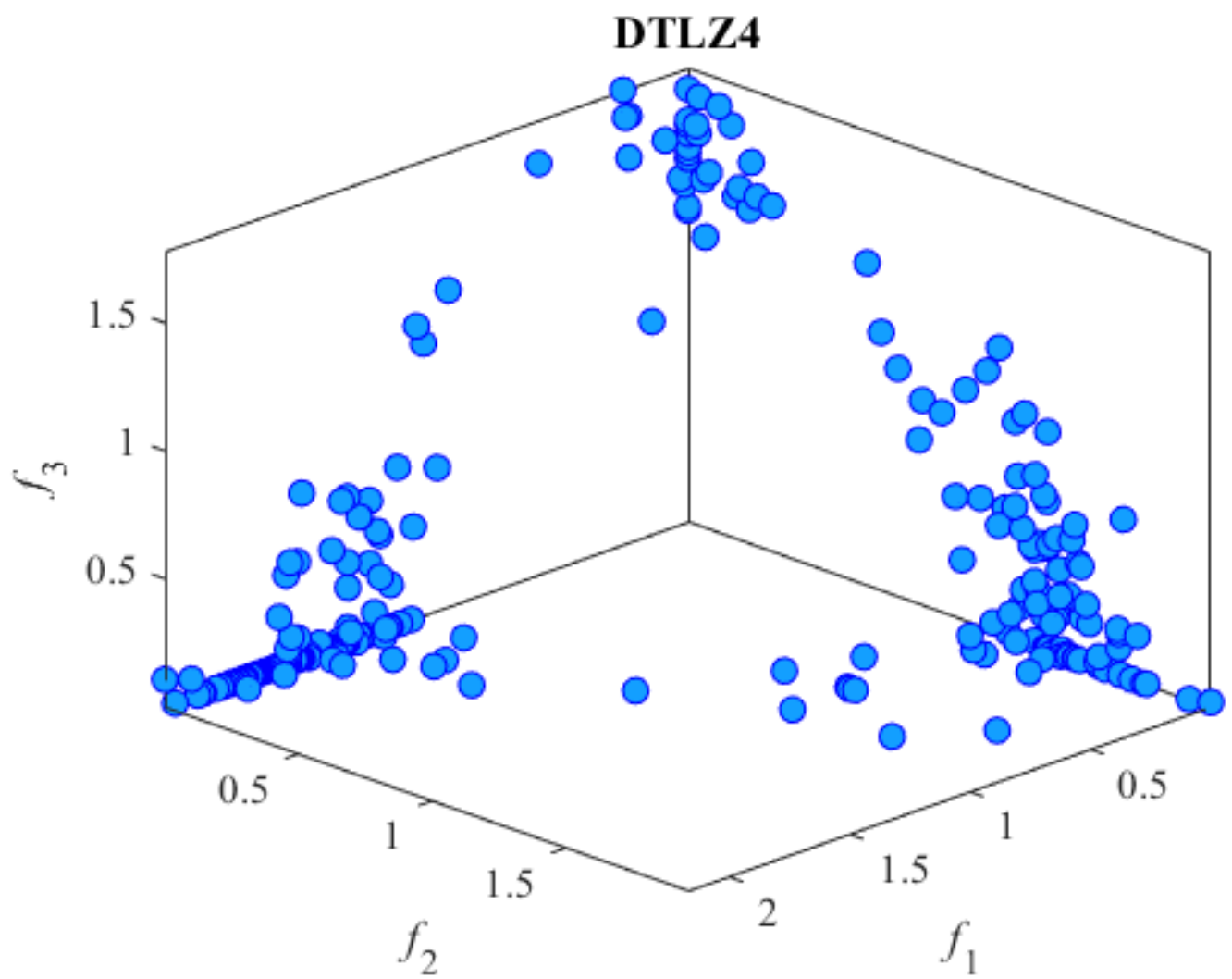}%
\hfil
}
\subfloat[HK-RVEA]{\includegraphics[width=1.8in]{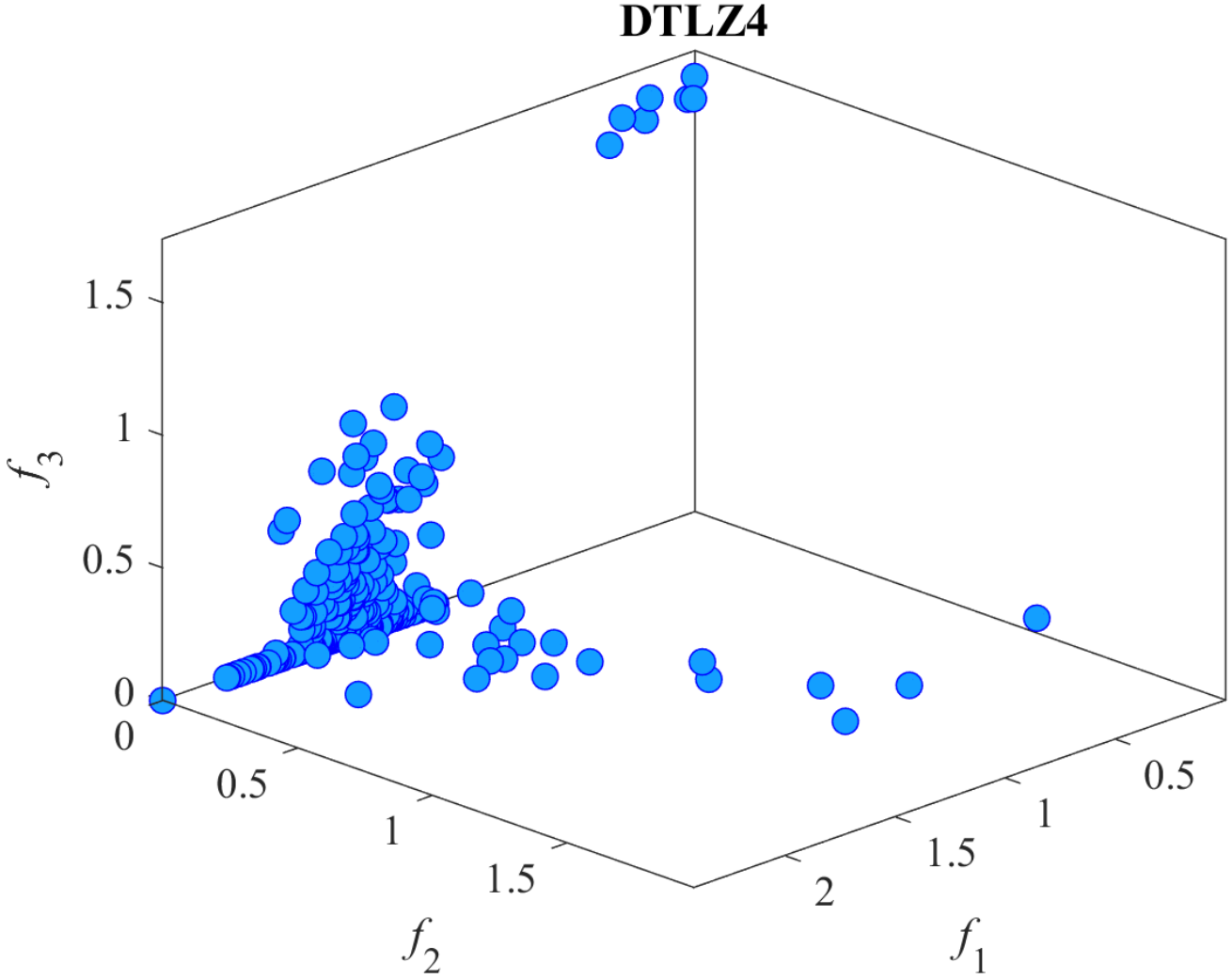}%
\hfil
}
\subfloat[SBP-BO]{\includegraphics[width=1.8in]{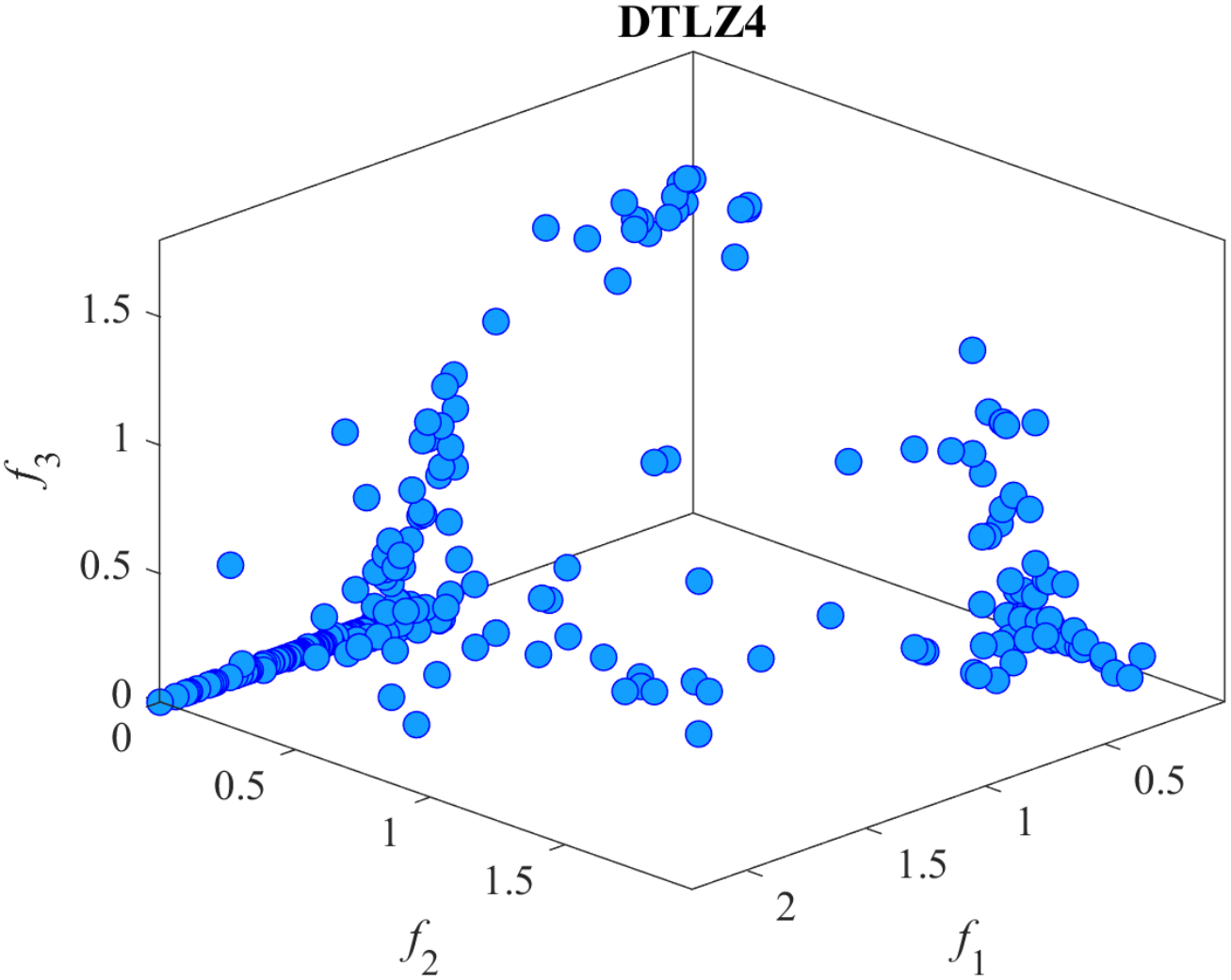}%
\hfil
}
}
\centerline{
\subfloat[True PF]{\includegraphics[width=1.8in]{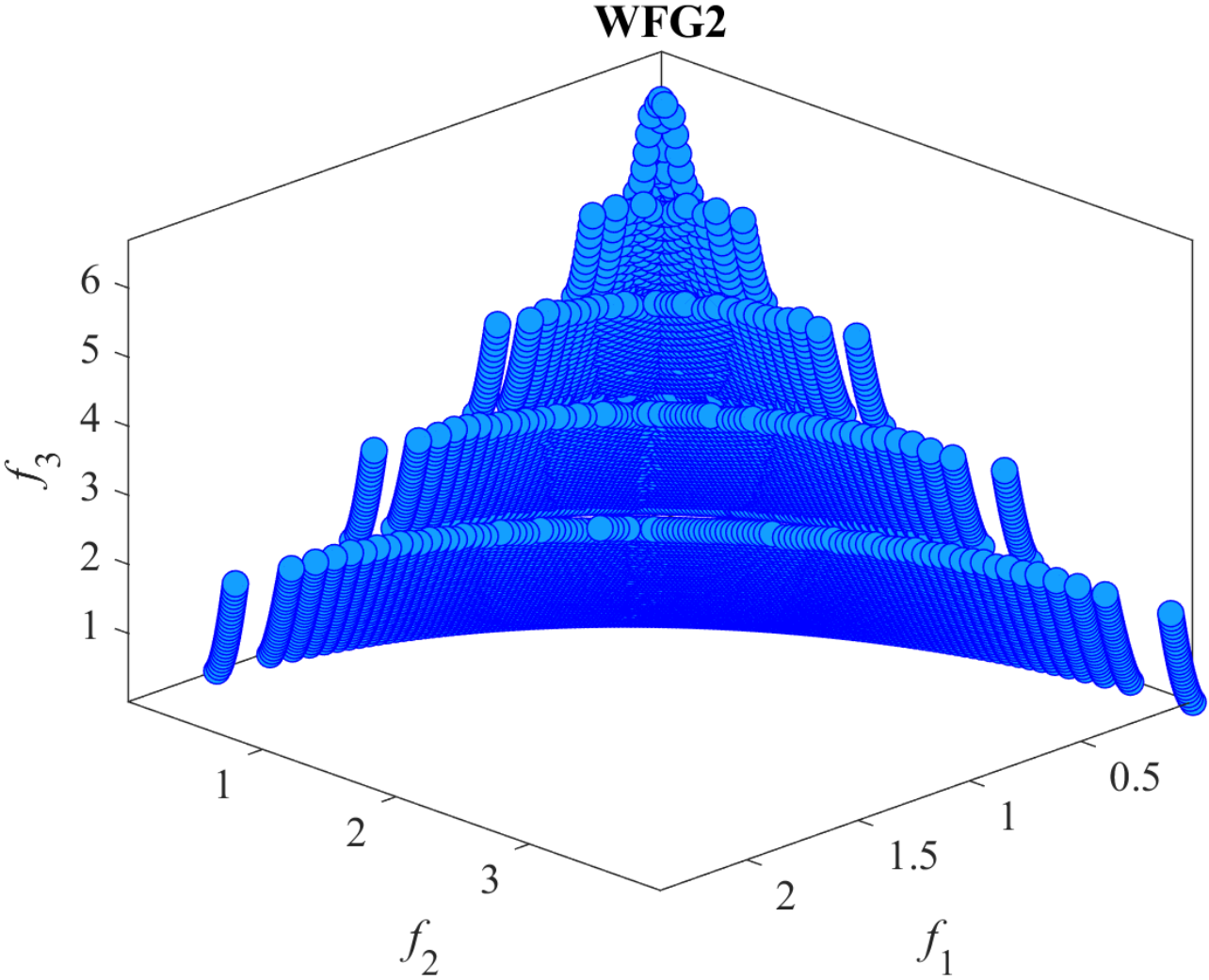}%
\hfil
}
\subfloat[K-RVEA]{\includegraphics[width=1.8in]{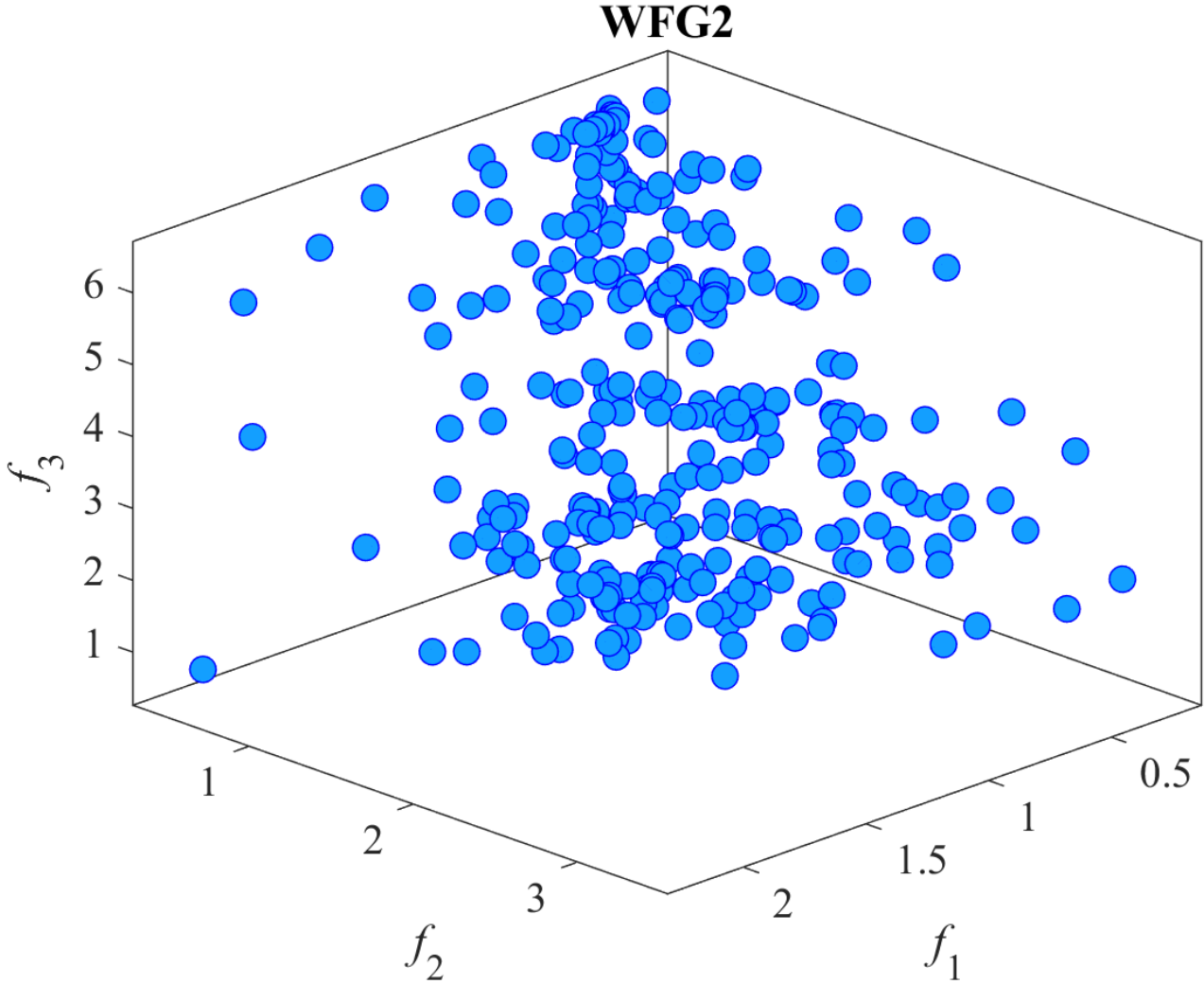}%
\hfil
}
\subfloat[HK-RVEA]{\includegraphics[width=1.8in]{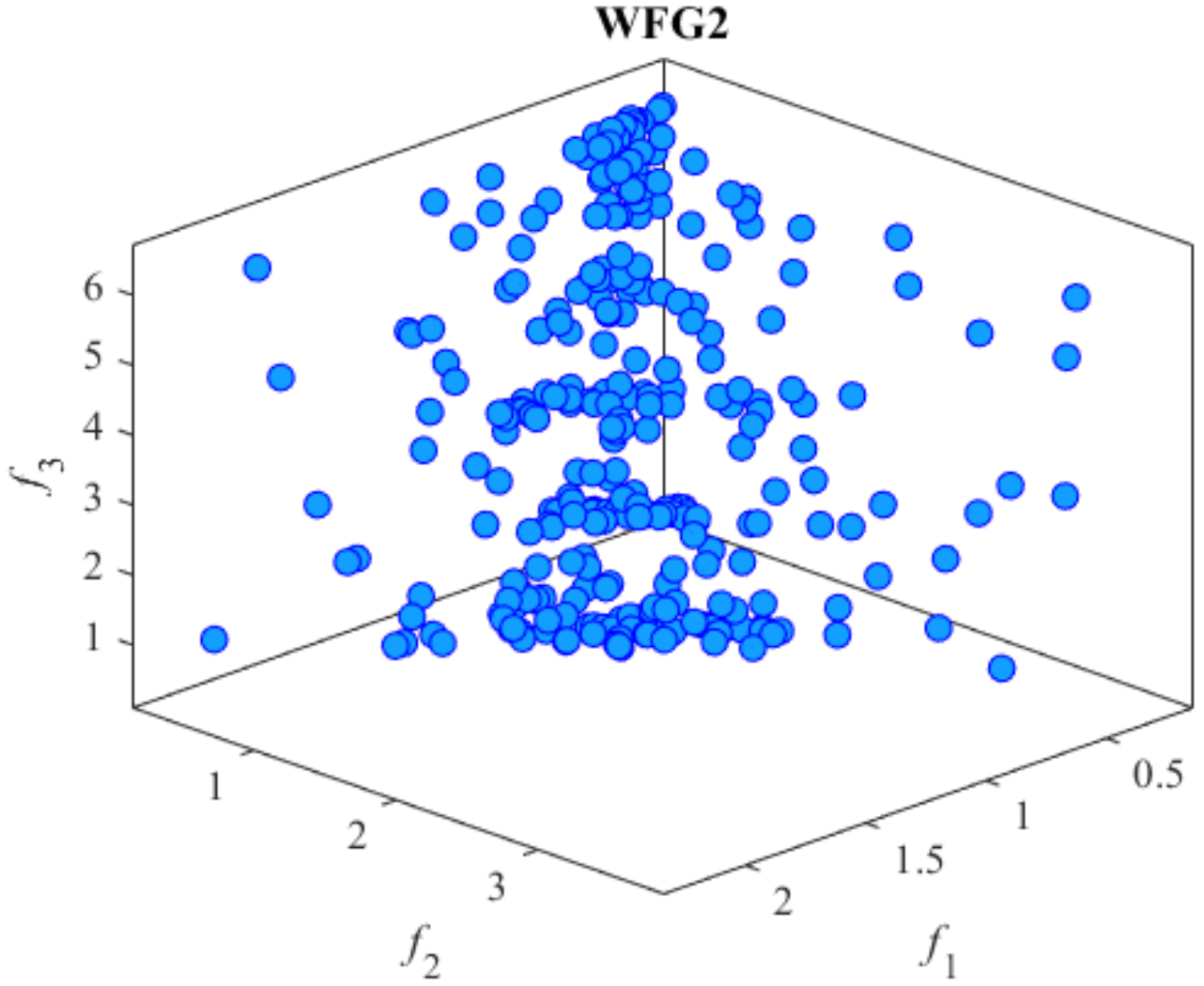}%
\hfil
}
\subfloat[SBP-BO]{\includegraphics[width=1.8in]{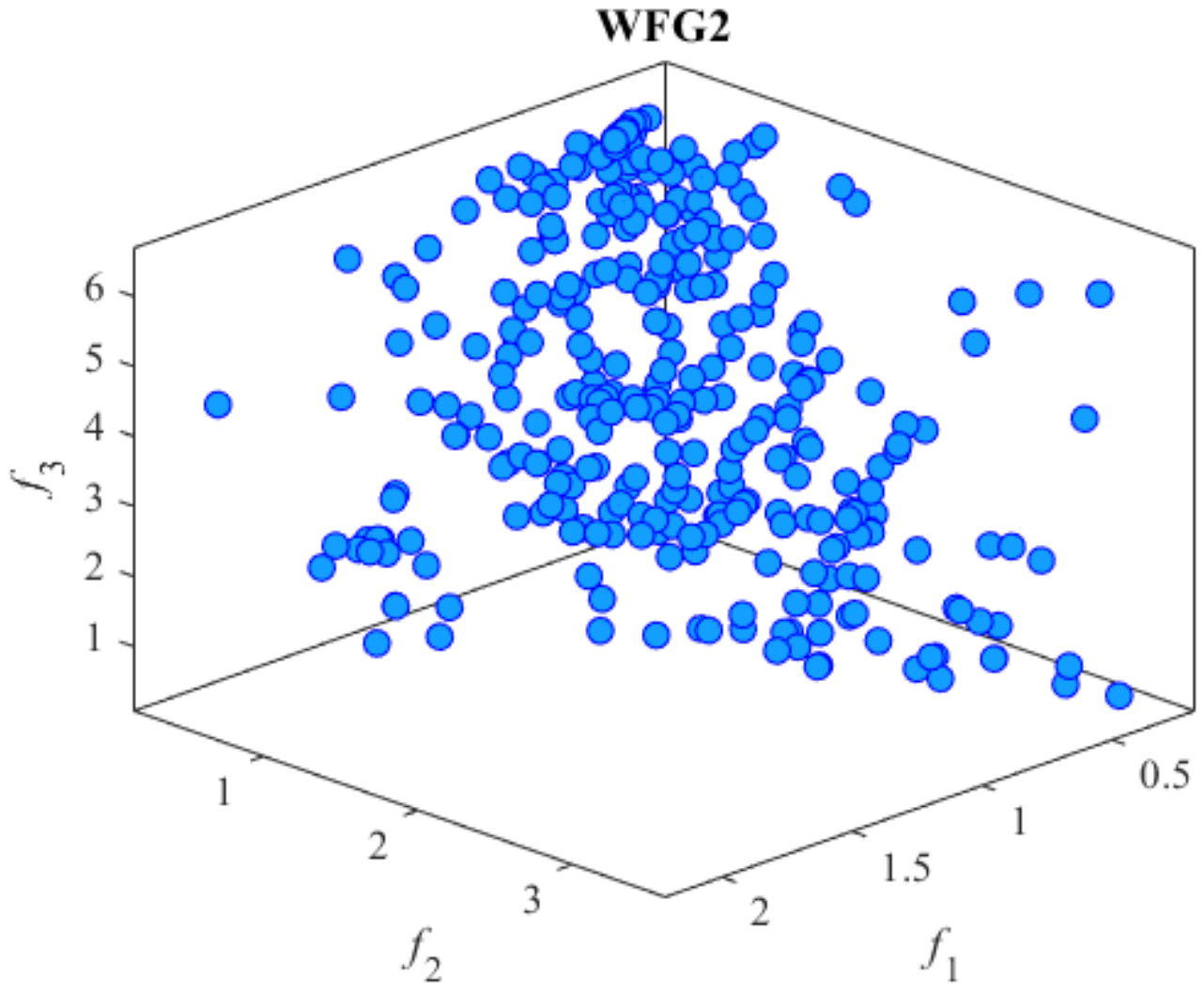}%
\hfil
}
}
\centerline{
\subfloat[True PF]{\includegraphics[width=1.8in]{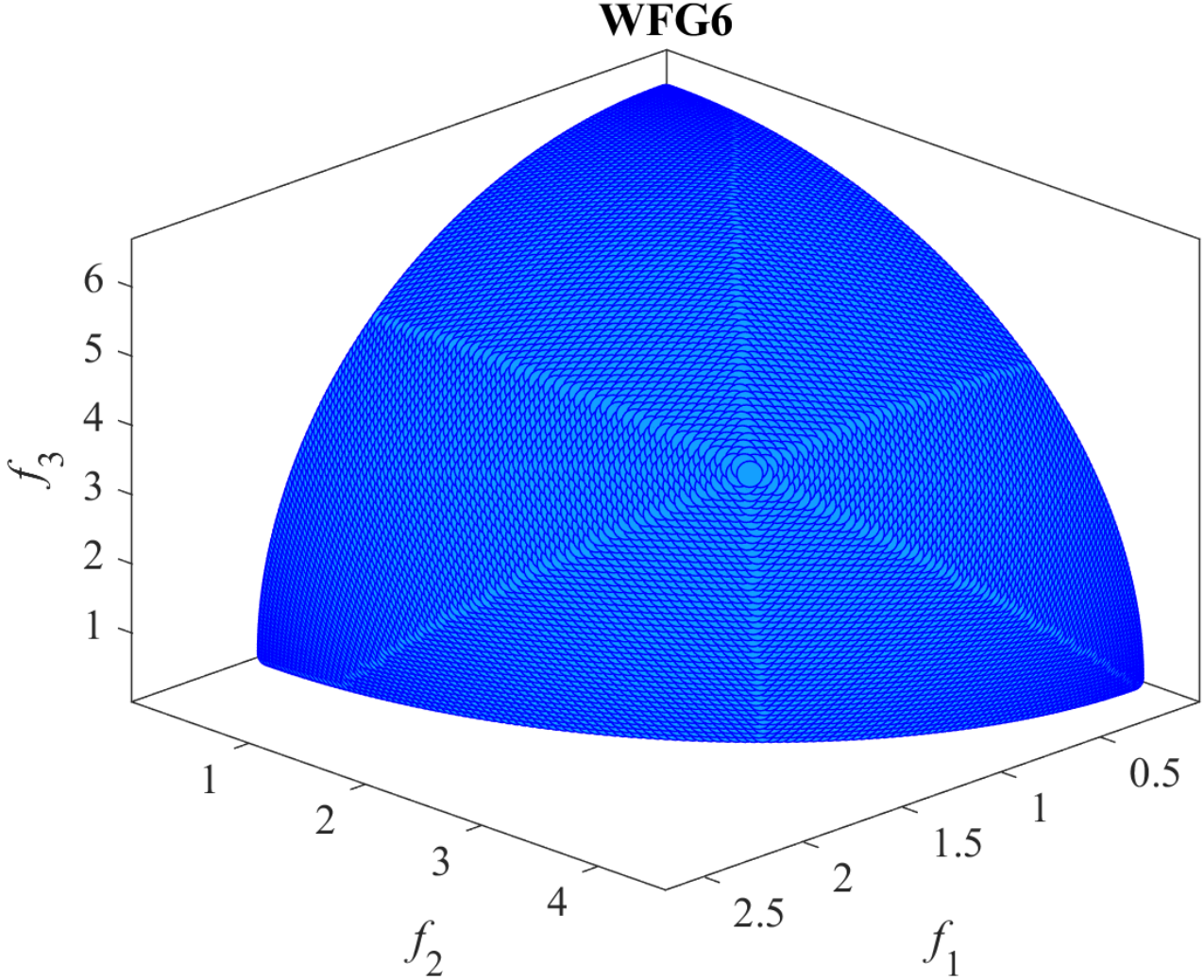}%
\hfil
}
\subfloat[K-RVEA]{\includegraphics[width=1.8in]{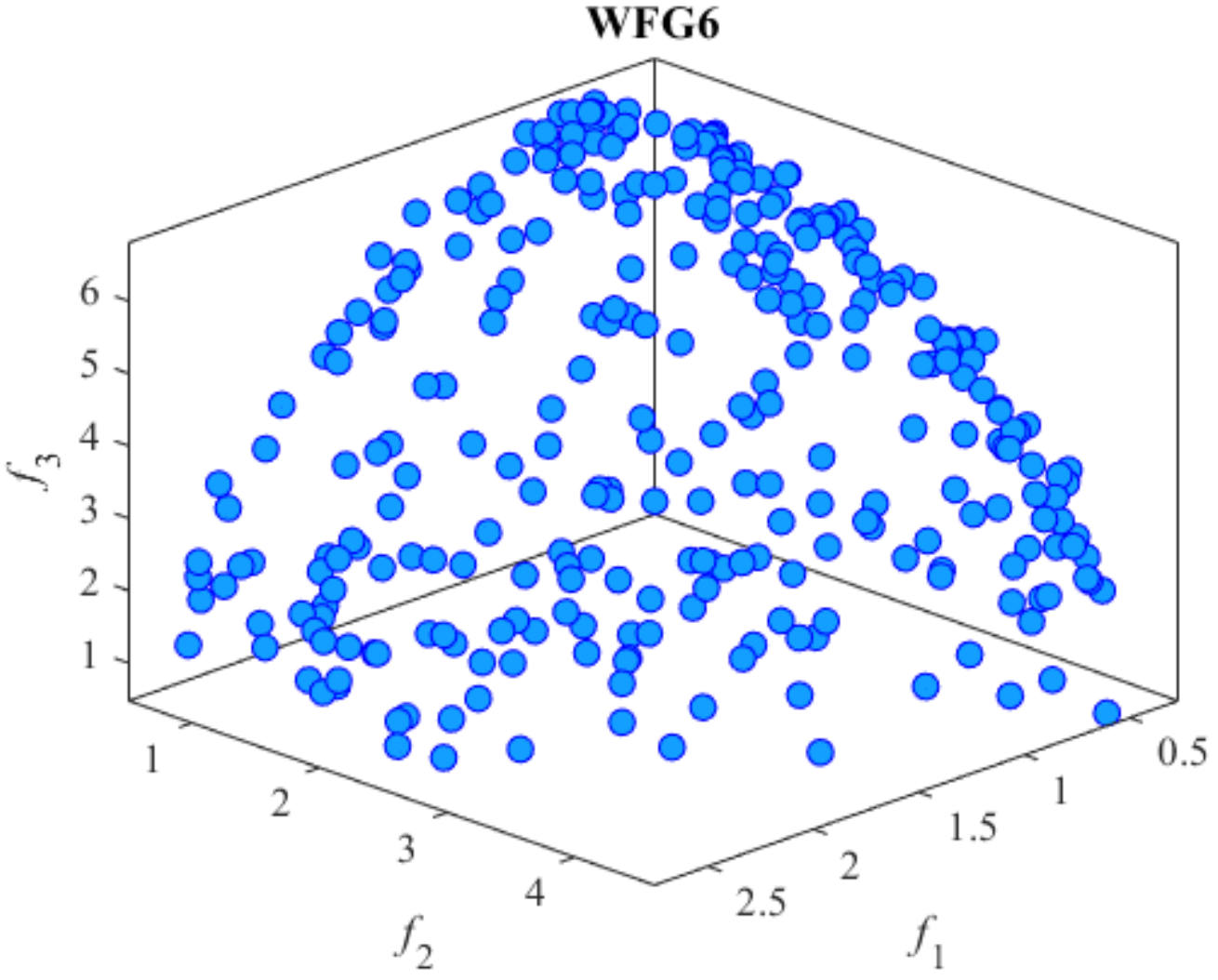}%
\hfil
}
\subfloat[HK-RVEA]{\includegraphics[width=1.8in]{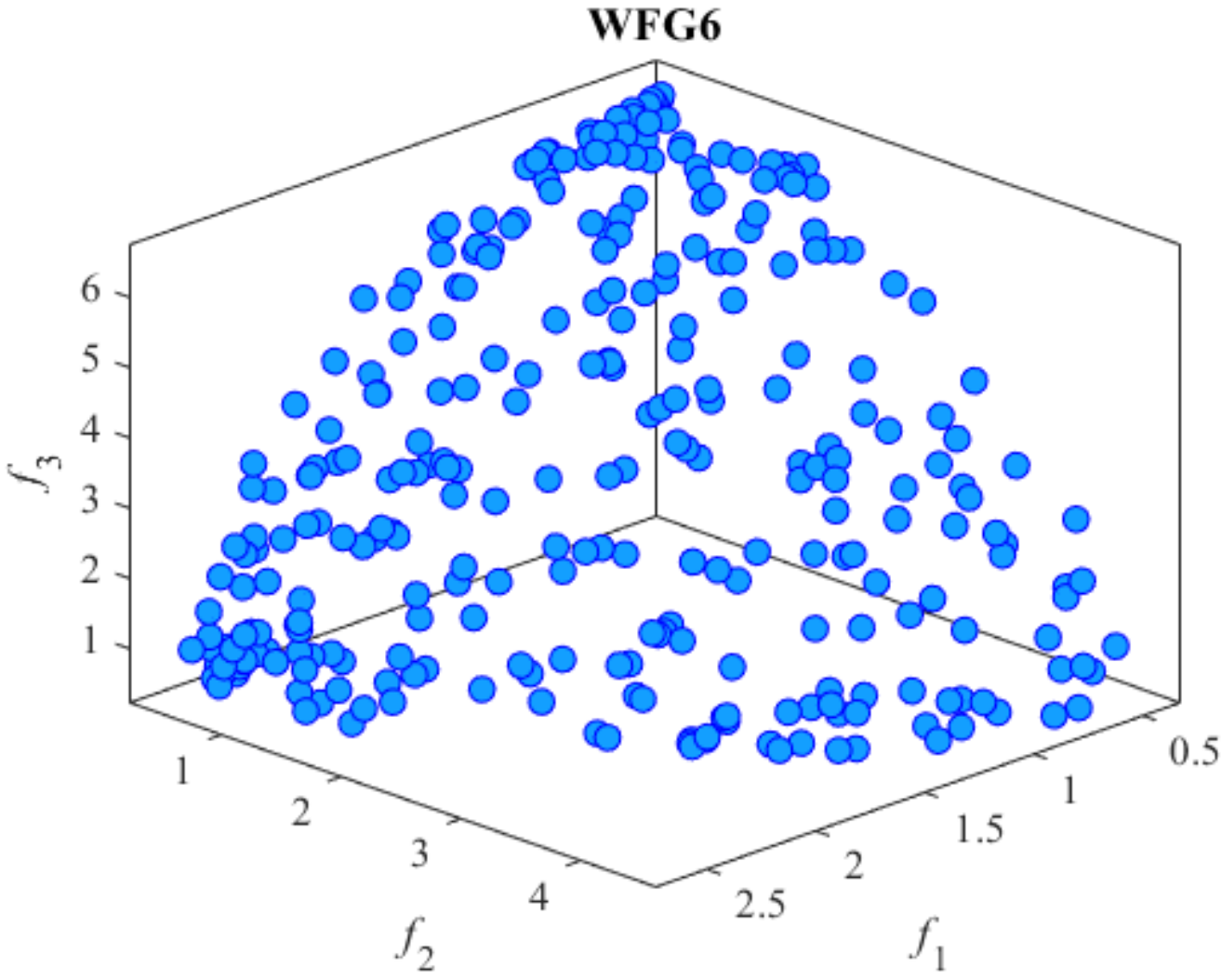}%
\hfil
}
\subfloat[SBP-BO]{\includegraphics[width=1.8in]{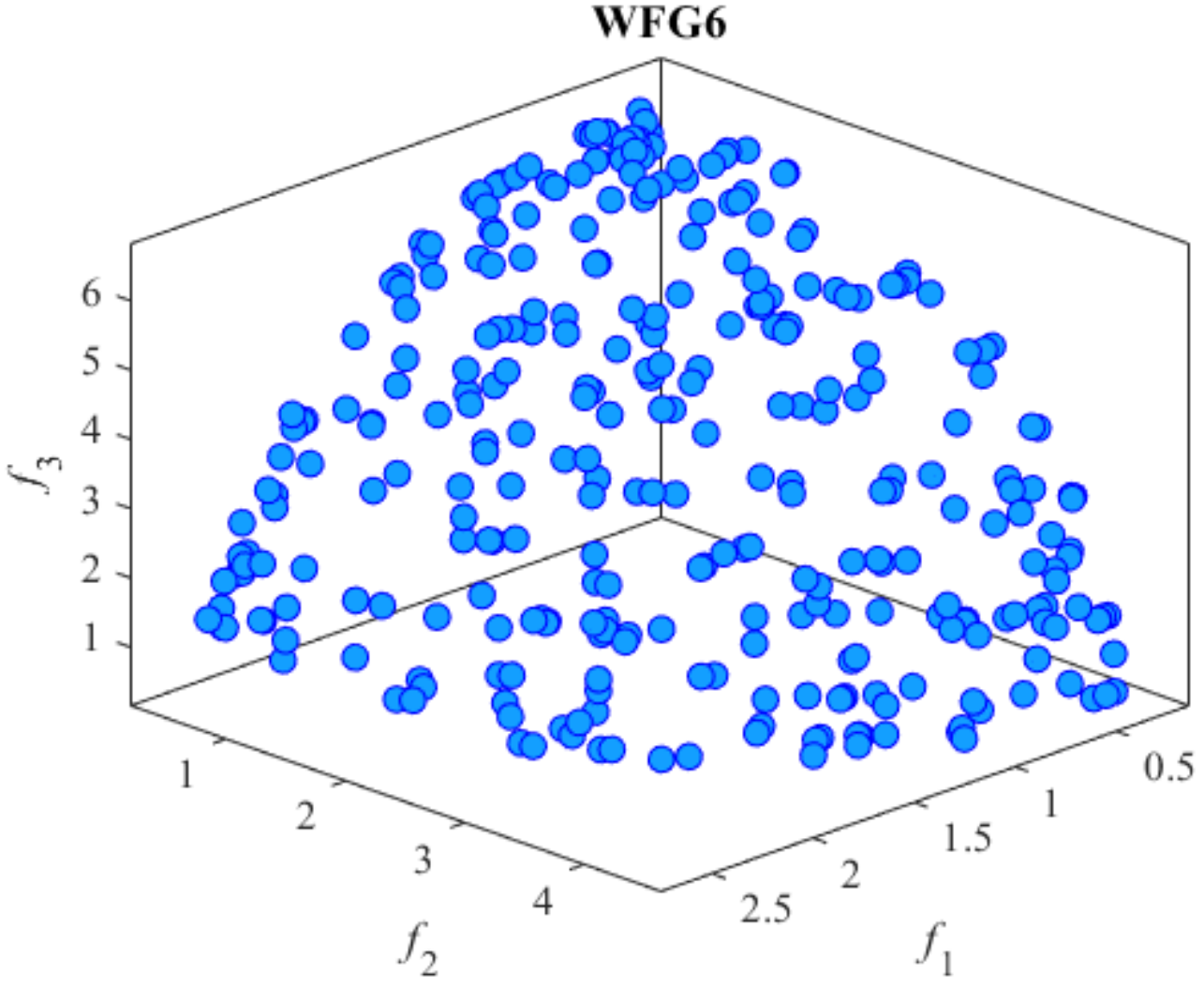}%
\hfil
}
}
\caption{The final solution set with the median IGD$^+$ values found by K-RVEA, HK-RVEA and SBP-BO on DTLZ2, DTLZ4, WFG2 and WFG6 problems with $m=3$ and $\mathbf{r}=(5,5,1)$.}
\label{SolutionM3}
\end{figure*}

2) \emph{Influence of different $\mathbf{r}$ and $r_\m{thres}$ on the optimization performance}: Each objective in an HE-MOP/HE-MaOP requires a distinct period of time to be evaluated, it is therefore expected to test the proposed algorithm on problems with different $\mathbf{r}=(r_{1}^{c}, r_{2}^{c},\cdots,r_{p}^{c},r_{1}^{e},\cdots,r_{q}^{e})$, where $p$ and $q$ are the number of cheap and expensive objectives. In this subsection, the heterogeneity handling ability of SBP-BO is tested on three-objective and ten-objective problems with different numbers of expensive objectives ($q$), different ratios of function evaluation times between objectives ($\mathbf{r}$), and different threshold values ($r_\m{thres}$). Firstly, SBP-BO is examined on three-objective test functions with two sets of $\mathbf{r}$ values, i.e., one set having $\mathbf{r}=(5,5,1)$, $\mathbf{r}=(7,3,1)$, and $\mathbf{r}=(9,1,1)$, and the other having $\mathbf{r}=(15,5,1)$, $\mathbf{r}=(10,10,1)$, and $\mathbf{r}=(19,1,1)$. The statistical results in terms of IGD$^+$ are presented in Tables SIII-SIV. Secondly, ten-objective problems with $\mathbf{r}=(10,9,8,7,6,5,4,3,2,1)$ and $r_\m{thres}=1,3,5$ are used to test the impact of $r_\m{thres}$ on the optimization performance, and the statistical results are presented in Table \ref{Tab.Differtrthre}. Lastly, the impact of different $\mathbf{r}$ is further investigated on ten-objective problems with $r_\m{thres}=3$, where $\mathbf{r}$ is set to $\mathbf{r1}=(10,8,8,7,5,4,3,2,2,1)$, $\mathbf{r2}=(10,9,8,6,3,2,2,2,1,1)$, and $\mathbf{r3}=(9,7,3,3,3,2,2,2,1,1)$, respectively. The results are summarized in Table \ref{Tab.Differtr}. 

Although it is unclear 
what influence the exact form of $\mathbf{r}$ and the exact value of $r_\m{thres}$ will have on the optimization process, we can make the following observations. First, the instance with $\mathbf{r}=(9,1,1)$ will cause a strong search bias towards the objective whose $r_{1}^{c}=9$ compared with $\mathbf{r}=(5,5,1)$ or $\mathbf{r}=(7,3,1)$, making the problem more difficult to solve. Consequently, the instance with $\mathbf{r}=(19,1,1)$ will render the strongest search bias among all $\mathbf{r}$ situations for three-objective problems studied in this work. Secondly, regarding the ten-objective test instances, a larger $r_\m{thres}$ will result in a stronger search bias towards the cheap objectives. Consequently, problems with $r_\m{thres}=5$ become more challenging to solve than those with $r_\m{thres}=1$. Moreover, we have a similar expectation that an optimization algorithm can achieve better performance on ten-objective problems with a smaller $q$ (i.e., $\mathbf{r1}$) than those with a larger one (i.e., $\mathbf{r3}$). 

The IGD$^{+}$ values in Tables SIII-SIV achieved by HK-RVEA and the proposed algorithm accord with our earlier observations that compared with HK-RVEA, SBP-BO shows similar or significantly better performance on all test problems except DTLZ7, with different $\mathbf{r}$. This observation provides evidence for confirming SBP-BO's ability for solving various HE-MOPs. Besides, it is interesting to note that changing $\mathbf{r}$ from $(5,5,1)$ to $(9,1,1)$ generally causes a reduced performance for both HK-RVEA and SBP-BO on most test problems, which is expected. According to the results in Table \ref{Tab.Differtr}, SBP-BO significantly outperforms HK-RVEA on five test instances, while similar performance can be achieved by both algorithms on the remaining ones. We can also observe that the performance of SBP-BO and HK-RVEA degrades as $\mathbf{r_\m{thres}}$ increases. Similar conclusions can be drawn from Table \ref{Tab.Differtrthre}, confirming the effectiveness of SBP-BO in handling HE-MOPs/HE-MaOPs.

\begin{figure}[!htbp]
\centerline{
\subfloat[HK-RVEA]{\includegraphics[width=1.8in]{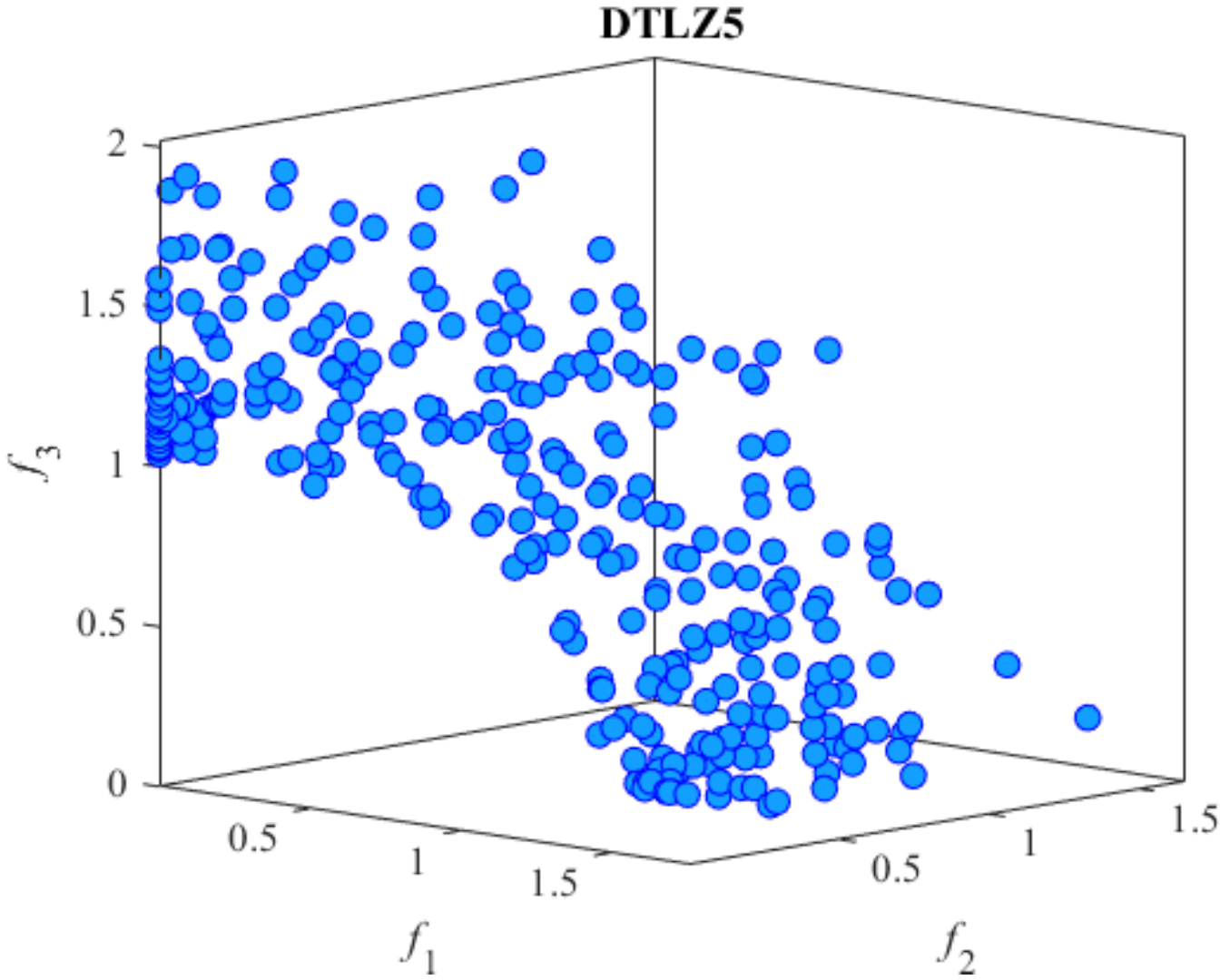}%
\hfil
}
\subfloat[SBP-BO]{\includegraphics[width=1.8in]{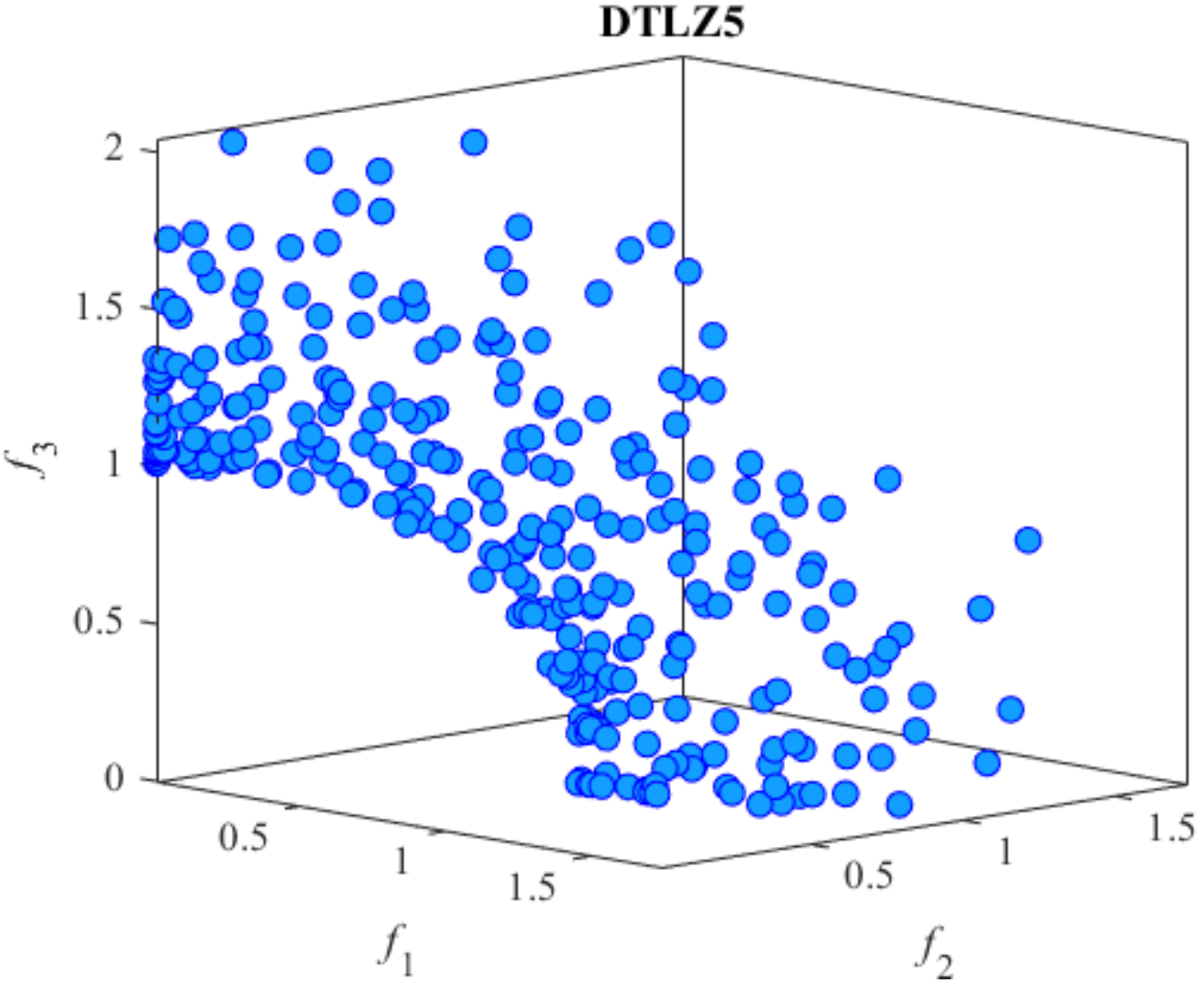}%
\hfil
}}
\centerline{
\subfloat[HK-RVEA]{\includegraphics[width=1.8in]{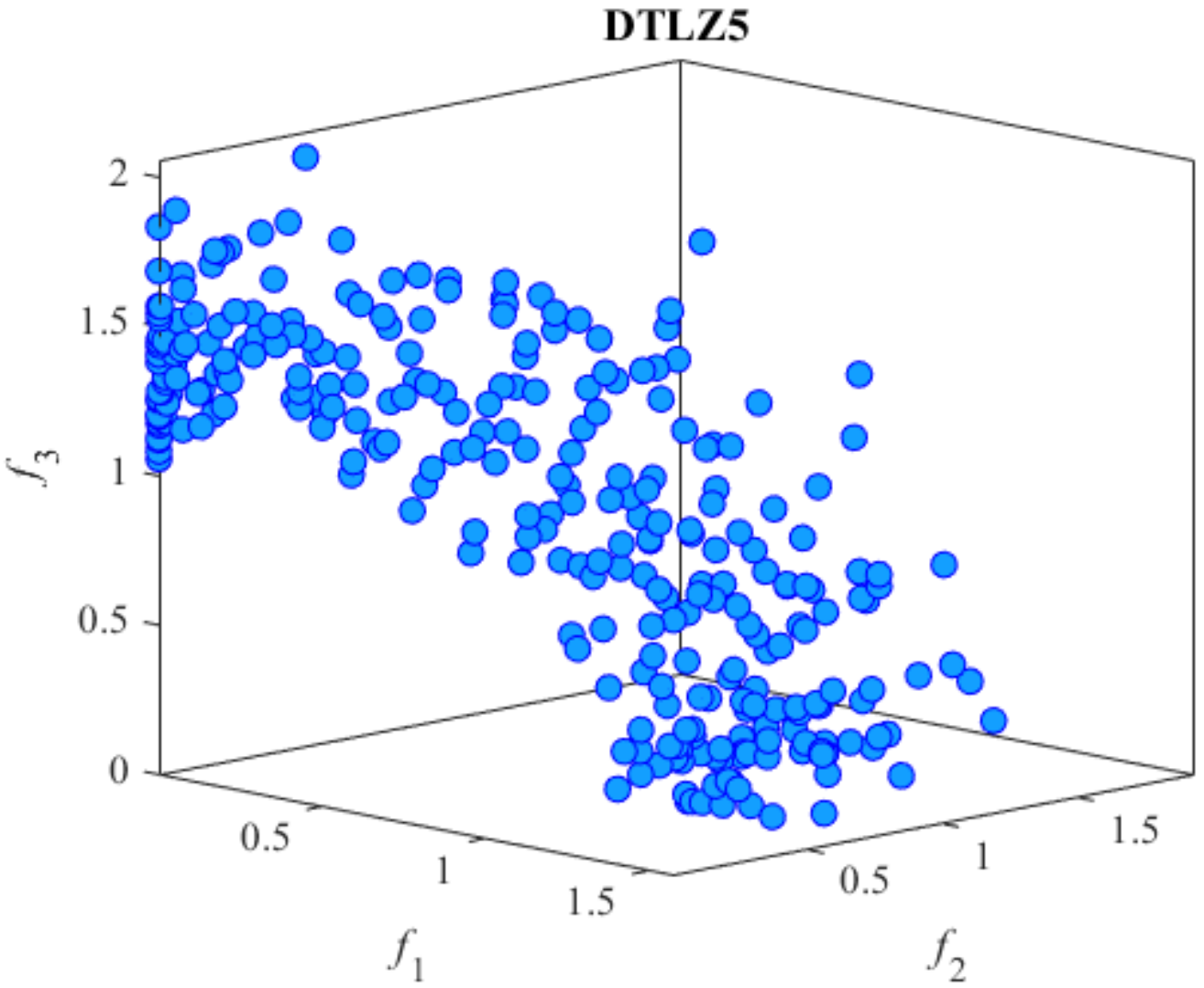}%
\hfil
}
\subfloat[SBP-BO]{\includegraphics[width=1.8in]{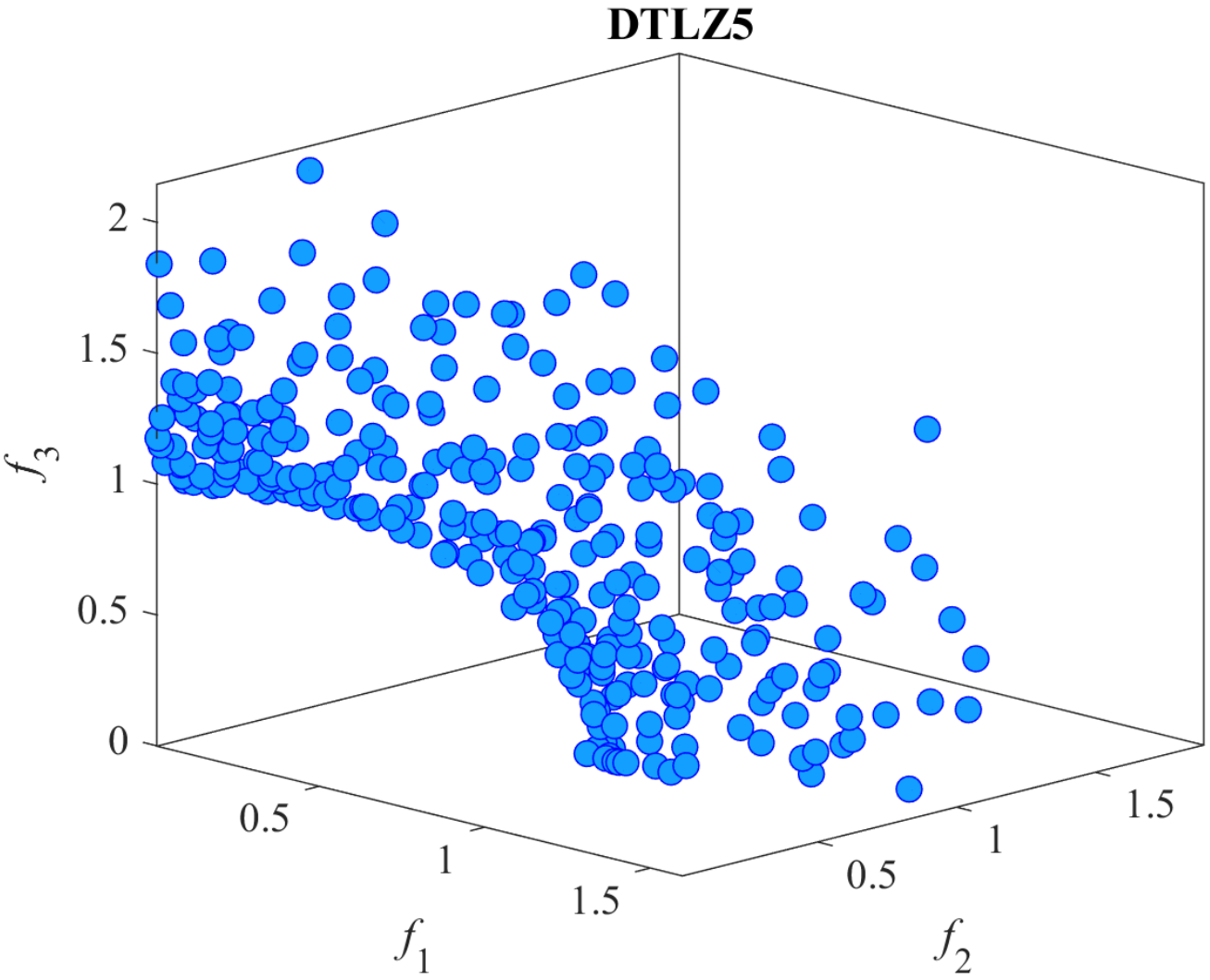}%
\hfil
}
}
\caption{The final solution set with the median IGD$^+$ values found by HK-RVEA and SBP-BO on DTLZ5 with $FE_{max}^{s}=300$ and $\mathbf{r}=(5,5,1)$ ((a) and (b)) and $\mathbf{r}=(9,1,1)$ ((c) and (d))}, respectively.
\label{DifferentR300}
\end{figure}

\begin{figure}[!htbp]
\centerline{
\subfloat[HK-RVEA]{\includegraphics[width=1.8in]{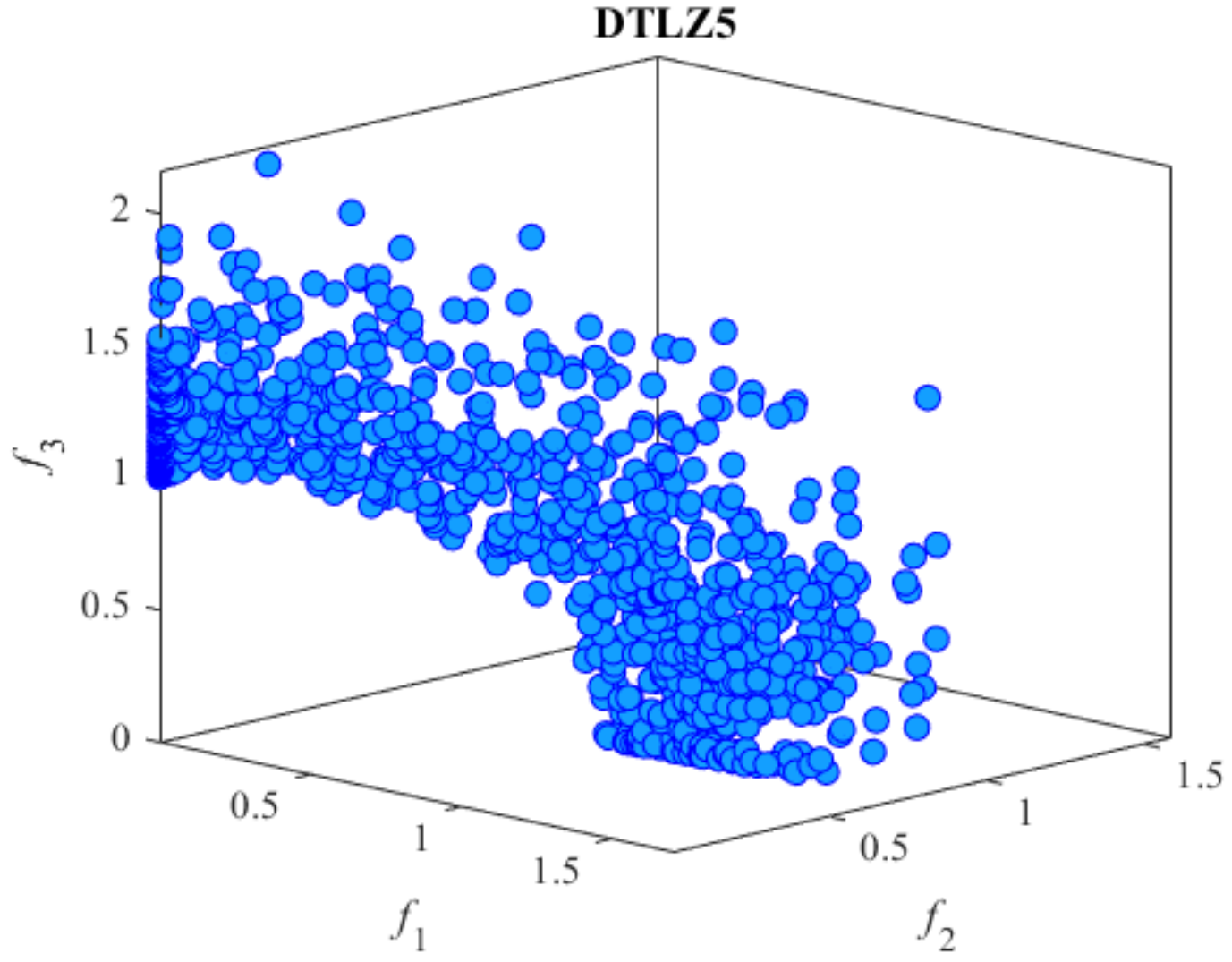}%
\hfil
}
\subfloat[SBP-BO]{\includegraphics[width=1.8in]{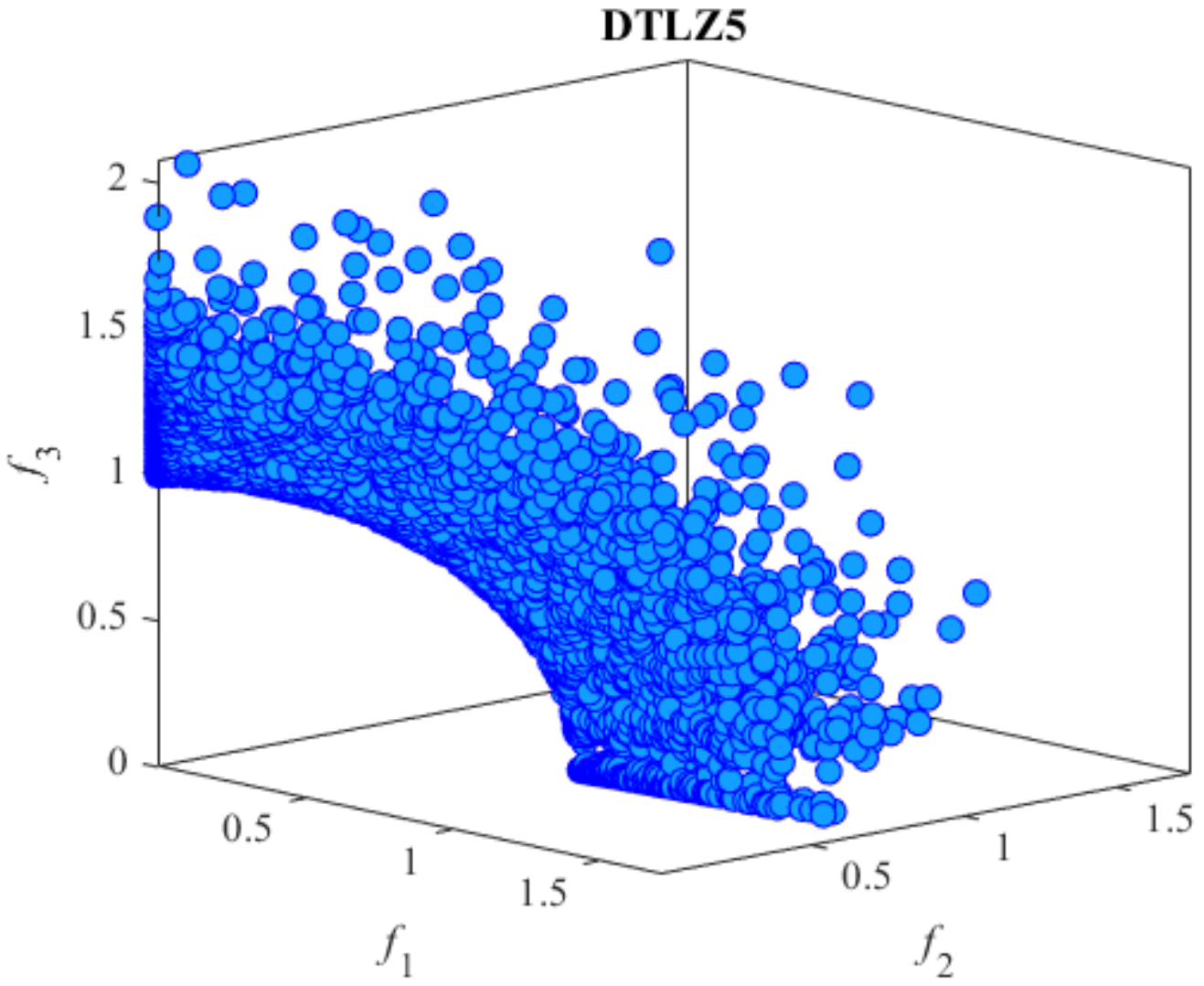}%
\hfil
}}
\centerline{
\subfloat[HK-RVEA]{\includegraphics[width=1.8in]{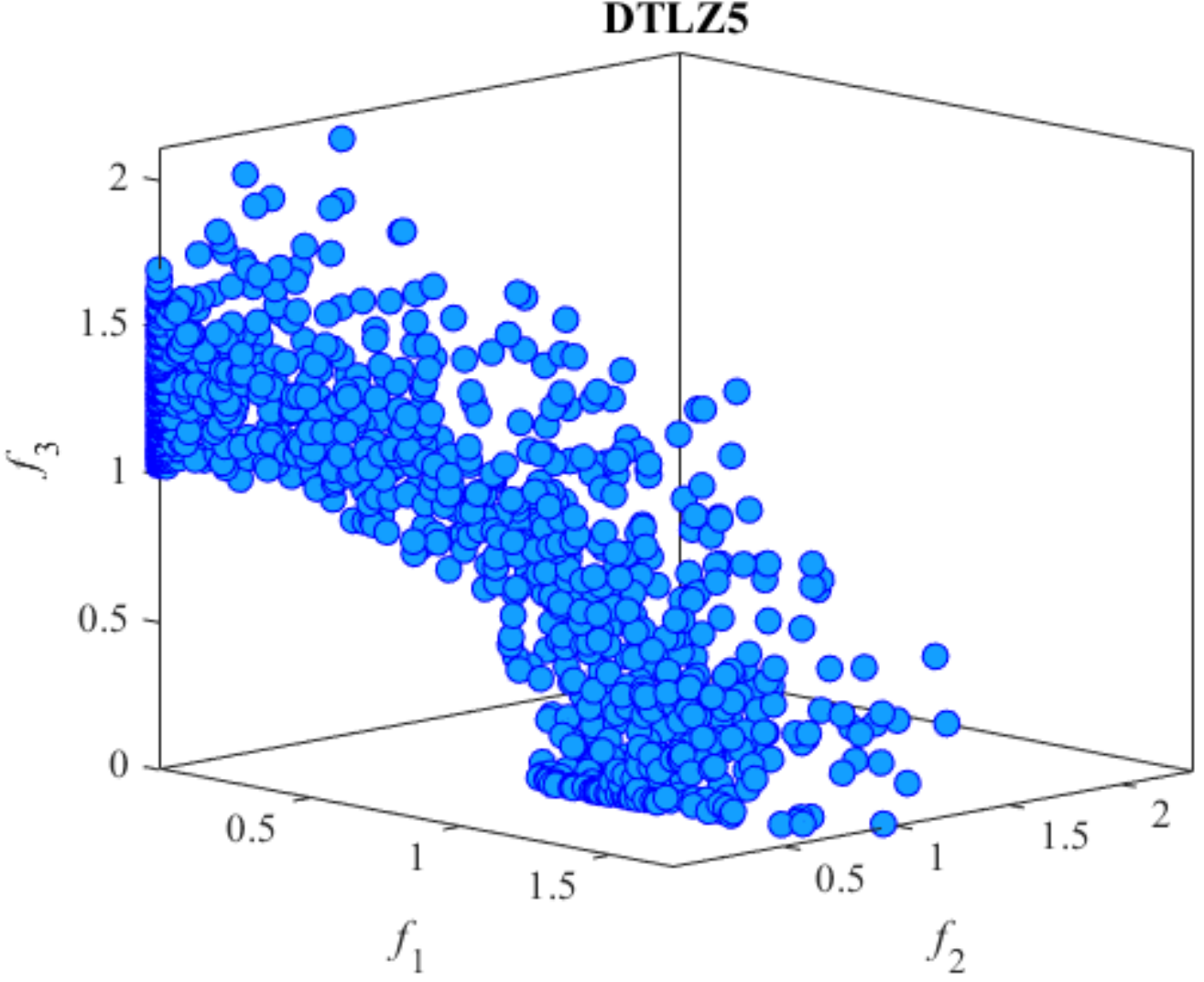}%
\hfil
}
\subfloat[SBP-BO]{\includegraphics[width=1.8in]{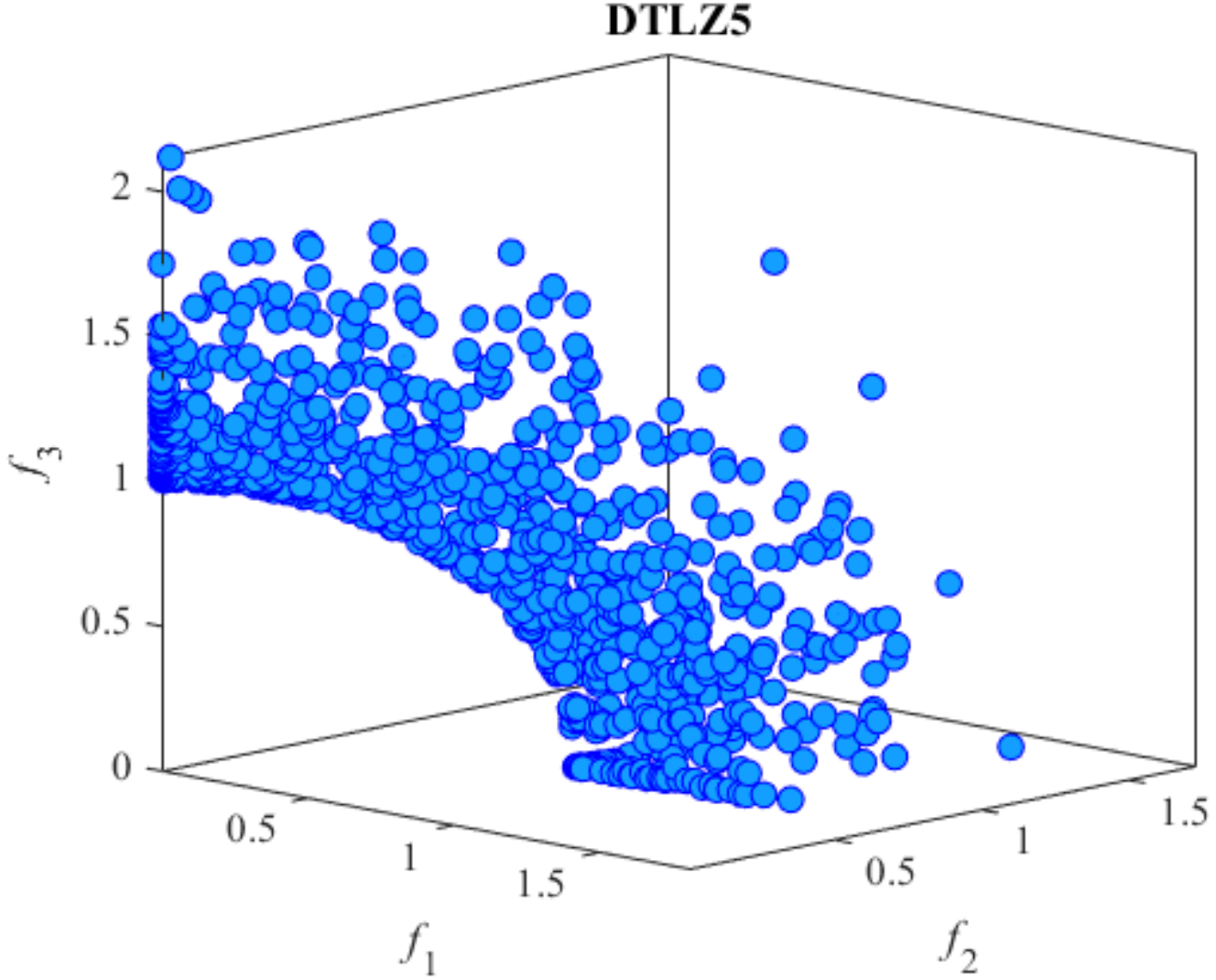}%
\hfil
}
}
\caption{The final solution set with the median IGD$^+$ values found by HK-RVEA and SBP-BO on DTLZ5 with $FE_{max}^{s}=1000$ and $\mathbf{r}=(5,5,1)$ ((a) and (b)) and $\mathbf{r}=(9,1,1)$ ((c) and (d)), respectively.}
\label{DifferentR1000}
\end{figure}

To further illustrate the influence of different $\mathbf{r}$ on the performance of the optimization algorithms considered in this work, the approximations of the true Pareto front on DTLZ5 with $\mathbf{r}=(5,5,1)$ and $\mathbf{r}=(9,1,1)$ obtained by HK-RVEA and SBP-BO are shown in Figs. \ref{DifferentR300}-\ref{DifferentR1000}, where $FE_{max}^{e}$ is set to 300 and 1000, respectively. From these results, our observations can be summarized as follows:
\begin{itemize}
\item Consistent with the aforementioned hypothesis, 
the search bias is more likely to occur on problems when $\mathbf{r}=(9,1,1)$ in the different function evaluation ratios considered between objectives, rendering a multi-objective optimization algorithm inefficient. An illustrative example is given in Fig. \ref{DifferentR300} when $FE_{max}^{e}=300$. SBP-BO can achieve better diversity on DTLZ5 with $\mathbf{r}=(5,5,1)$ compared with that on DTLZ5 with $\mathbf{r}=(9,1,1)$. Similar observations can be made from Fig. \ref{DifferentR1000}, where $FE_{max}^{e}=1000$. 
\item The proposed SBP-BO can find a set of solutions with good quality on all considered test instances, while HK-RVEA suffers from the heterogeneous objectives. As depicted in Fig. \ref{DifferentR300}, the solution set obtained by HK-RVEA only covers some subregions, especially those around the two end points of the true Pareto front of DTLZ5, which is a curve for that problem. We note that the proposed SBP-BO shows its advantage for handling HE-MOPs, which is achieved by an efficient use of the additional data on the cheap objectives with the help of the ensemble GP surrogate and the alleviation search bias by means of the proposed acquisition function. It is clear that SBP-BO finds a satisfying Pareto front approximation with respect to both diversity and convergence when $FE_{max}^{e}=1000$, confirming that the proposed SBP acquisition function can reduce the search bias introduced by heterogeneous objectives. 
\end{itemize}

\begin{table}[]
\center
\caption{Mean (Standard Deviation) IGD$^{+}$ values obtained by HK-RVEA, and SBP-BO on ten-objective problems with $r = (10,9,8,7,6,5,4,3,2,1)$ and different $r_\m{thres}$ and $FE_{max}^{e}=300$}.
\label{Tab.Differtrthre}
\renewcommand{\arraystretch}{1}
\resizebox{0.4\textwidth}{!}{
\begin{tabular}{lccc}
\toprule
Problem & \multicolumn{1}{c}{$r_\m{thres}$} & HKRVEA                & SBP-BO              \\\midrule[0.3pt]
        & 1                    & 3.45e-1   (2.26e-1) $\approx$ & \cellcolor[HTML]{C0C0C0}3.29e-1   (8.81e-2) \\
DTLZ1   & 3                   & 6.18e-1   (1.96e-1) $\approx$ & \cellcolor[HTML]{C0C0C0}6.07e-1 (3.44e-1)   \\
        & 5                    &\cellcolor[HTML]{C0C0C0}1.53e+0   (9.21e-1) $\approx$ & 1.55e+0   (5.18e-1) \\\midrule[0.3pt]
        & 1                    & 1.21e+0   (5.28e-2) --  & \cellcolor[HTML]{C0C0C0}1.04e+0   (3.23e-2) \\
DTLZ2   & 3                    & \cellcolor[HTML]{C0C0C0}1.49e+0   (1.89e-1) $\approx$ & 1.61e+0 (1.06e+0)   \\
        & 5                    & 2.77e+0   (1.66e+0) -- & \cellcolor[HTML]{C0C0C0}1.98e+0   (9.56e-1) \\\midrule[0.3pt]
        & 1                   & 2.37e-1   (7.60e-2) -- & \cellcolor[HTML]{C0C0C0}1.64e-1   (2.84e-2) \\
DTLZ3   & 3                    & 3.17e-1   (1.61e-1) $\approx$ & \cellcolor[HTML]{C0C0C0}3.15e-1 (1.12e-1)   \\
        & 5                   & 3.26e-1   (1.42e-1) -- & \cellcolor[HTML]{C0C0C0}3.08e-1   (1.39e-1) \\\midrule[0.3pt]
        & 1                    & 2.52e-1   (1.73e-2) $\approx$ & \cellcolor[HTML]{C0C0C0}2.45e-1 (1.59e-2)   \\
DTLZ4   & 3                    & 2.64e-1   (1.65e-2) -- & \cellcolor[HTML]{C0C0C0}2.48e-1 (1.98e-2)   \\
        & 5                    & \cellcolor[HTML]{C0C0C0}2.71e-1   (3.94e-2) $\approx$ & 2.80e-1   (3.25e-2)
                    \\ \midrule[0.3pt]
\multicolumn{1}{l}{+/--/ $\approx$} & \multicolumn{1}{l}{} & \multicolumn{1}{c}{0/5/7}                                           & \multicolumn{1}{l}{}                 \\ \bottomrule
\end{tabular}}
\end{table}

\begin{table}[]
\center
\caption{Mean (Standard Deviation) IGD$^{+}$ values obtained by HK-RVEA, and SBP-BO on ten-objective problems with $r_{thres}=3$, $\mathbf{r1} = (10,8,8,7,5,4,3,2,2,1)$, $\mathbf{r2} =(10,9,8,6,3,2,2,2,1,1)$ and $\mathbf{r3} = (9,7,3,3,3,2,2,2,1,1)$, and  $FE_{max}^{e}=300$}.
\label{Tab.Differtr}
\renewcommand{\arraystretch}{1}
\resizebox{0.4\textwidth}{!}{
\begin{tabular}{lccc}
\toprule
Problem & \multicolumn{1}{c}{$\mathbf{r}$} & HK-RVEA                & SBP-BO              \\\midrule[0.3pt]
        & $\mathbf{r1}$                    &3.56e-1  (1.33e-1) -- & \cellcolor[HTML]{C0C0C0}2.75e-1   (1.00e-1) \\
DTLZ1   & $\mathbf{r2}$                    & 2.66e-1   (1.01e-1) $\approx$& \cellcolor[HTML]{C0C0C0}2.51e-1 (1.16e-1)   \\
        & $\mathbf{r3}$                   & 3.29e-1   (1.65e-1) $\approx$ & \cellcolor[HTML]{C0C0C0}2.71e-1   (9.60e-2) \\\midrule[0.3pt]
        & $\mathbf{r1}$                    & 2.79e-1   (1.70e-2) $\approx$ & \cellcolor[HTML]{C0C0C0}2.67e-1   (2.15e-2) \\
DTLZ2   & $\mathbf{r2}$                    & \cellcolor[HTML]{C0C0C0}2.67e-1   (2.70e-2) $\approx$ & 2.77e-1 (2.36e-2)   \\
        & $\mathbf{r3}$                   &\cellcolor[HTML]{C0C0C0}2.80e-1   (4.18e-2) + & 3.11e-1 (2.70e-2)   \\\midrule[0.3pt]
        & $\mathbf{r1}$                   & 1.41e+0   (7.11e-1) $\approx$ & \cellcolor[HTML]{C0C0C0}1.35e+0 (6.09e-1)   \\
DTLZ3   & $\mathbf{r2}$                     & \cellcolor[HTML]{C0C0C0}7.30e-1   (4.76e-1) $\approx$ & 9.10e-1 (5.26e-1)   \\
        & $\mathbf{r3}$                   & 1.48e+0   (5.47e-1) $\approx$& \cellcolor[HTML]{C0C0C0}1.30e+0 (5.60e-1)   \\\midrule[0.3pt]
        & $\mathbf{r1}$                   & 3.44e+0   (1.66e+0) -- & \cellcolor[HTML]{C0C0C0}2.14e+0 (1.07e+0)   \\
DTLZ4   & $\mathbf{r2}$                     & 2.71e+0   (9.07e-1) -- & \cellcolor[HTML]{C0C0C0}1.58e+0 (1.76e-1)   \\
        & $\mathbf{r3}$                    & 4.01e+0   (9.13e-1) -- & \cellcolor[HTML]{C0C0C0}3.05e+0   (7.56e-1)
            \\ \midrule[0.3pt]
\multicolumn{1}{l}{+/--/ $\approx$} & \multicolumn{1}{l}{} & \multicolumn{1}{c}{1/4/7}                                           & \multicolumn{1}{l}{}                 \\ \bottomrule
\end{tabular}}
\end{table}

3) \emph{Results on bi-objective heterogeneous problems}: In this subsection, we compare SBP-BO with three transfer-learning (TL) based heterogeneity-handling methods, i.e., T-SAEA \cite{XiluWang2020TSAEA}, Tr-SAEA \cite{wang2021transfer} and TC-SAEA \cite{WangTC-SAEA2021}, on the same bi-objective heterogeneous problems with the same parameter setting reported in \cite{WangTC-SAEA2021}. Specifically, as presented in the Supplementary material, DTLZ1 to DTLZ7 and two modified counterparts (DTLZ1a and DTLZ3a) of DTLZ1 and DTLZ3, and UF1 to UF7 from the UF test suite \cite{zhang2008multiobjective}, are used as test instances. The statistical results in terms of IGD values \cite{zitzler2003performance} obtained by each algorithm on test instances with $r^c=5$ and $r^c=10$ are summarized in Tables \ref{Bi-objective1} and SV, respectively. We can observe that at least one of the TL-based methods significantly outperforms SBP-BO on 10 out of 16 problems for $r^c=5$, while SBP-BO is only significantly better than all TL approaches on two problems.
Similar observations can be made from Table SV. The performance difference is understandable since in the TL-based approaches information on the correlation of the two objectives is acquired, which allows for an estimation of the expensive objective from the search experience on the cheap objective. Such information is not used in SBP-BO and the selection of new samples is only guided by the heterogeneous evaluation times in SBP-BO.

However, the extension of the TL-based approaches to problems with more than two objectives is nontrivial. Multiple models for transferring knowledge between each pair of objectives will need to be trained, which increases the computational complexity substantially. More importantly, it is an open question how to utilize the information from these models in a consistent manner, as, for example, information on one objective will be provided by multiple models that might contain contradicting information.



\begin{table}[thp]
\center
\caption{Mean (Standard Deviation) IGD values obtained by T-SAEA, Tr-SAEA, TC-SAEA and SBP-BO for bi-objective problems with $FE_{max}^{e}=200$ and $r=(r^c,1)$ where $r^c=5$}.
\label{Bi-objective1}
\resizebox{0.45\textwidth}{!}{
\begin{tabular}{l|llll}
\hline
\multicolumn{1}{l|}{Problem} & T-SAEA         & Tr-SAEA       & TC-SAEA       & SBP-BO     \\ \hline
DTLZ1                        & 21.7 (11.9) $\approx$ & 20.7 (5.38) $\approx$ & 20.1 (8.16)$\approx$ & \cellcolor[HTML]{C0C0C0}19.2 (9.52) \\
DTLZ1a                       & \multicolumn{1}{l}{1.06 (1.00) --}   & \multicolumn{1}{l}{0.21 (0.07) --}  & \multicolumn{1}{l}{0.36 (0.04) --}  & 0.14 (0.05) \\
DTLZ2                        & 0.05 (0.03) $\approx$ & 0.03 (0.01) $\approx$ & \cellcolor[HTML]{C0C0C0}0.02 (0.00) $\approx$ & 0.05 (0.01) \\
DTLZ3                        & 203 (100) $\approx$    & 327 (82.1) $\approx$  & \cellcolor[HTML]{C0C0C0}132 (79.3) $\approx$  & 204 (92.2)  \\
DTLZ3a                       & 5.34 (37.5) +    & 3.39 (1.87) +   & \cellcolor[HTML]{C0C0C0}2.30 (0.66) +   & 13.8 (5.56) \\
DTLZ4                        & 0.60 (0.13) --    & 0.16 (0.07) +   & 0.44 (0.13) --  & 0.38 (0.12) \\
DTLZ5                        & 0.05 (0.02) $\approx$  & 0.03 (0.03) $\approx$ & \cellcolor[HTML]{C0C0C0}0.03 (0.00) $\approx$ & 0.05 (0.00) \\
DTLZ6                        & 2.56 (1.21) +    & \cellcolor[HTML]{C0C0C0}0.72 (0.09) +   & 2.62 (1.95) +   & 5.76 (0.47) \\
DTLZ7                        & 1.15 (0.91)  +    &\cellcolor[HTML]{C0C0C0}0.03 (0.01) +   & 0.05 (0.08) +   & 5.01 (0.46) \\
UF1                          & 0.19 (0.02) +    &\cellcolor[HTML]{C0C0C0}0.19 (0.01) +   & 0.19 (0.02) +   & 1.20 (0.14) \\
UF2                          & 0.14 (0.02) +    & \cellcolor[HTML]{C0C0C0}0.12 (0.01) +   & 0.13 (0.02) +   & 0.59 (0.03) \\
UF3                          & \cellcolor[HTML]{C0C0C0}0.19 (0.08) +    & 0.49 (0.01) $\approx$   & 0.42 (0.03) +   & 1.09 (0.05) \\
UF4                          & 0.23 (0.02) --   & 0.22 (0.00) --  & 0.19 (0.01) --  & \cellcolor[HTML]{C0C0C0}0.17 (0.00) \\
UF5                          & 2.49 (0.44) +    & 2.43 (0.28) +   & \cellcolor[HTML]{C0C0C0}2.42 (0.38) +   & 4.91 (0.36) \\
UF6                          & 1.01 (0.25) +    & 1.32 (0.39) +   & \cellcolor[HTML]{C0C0C0}0.81 (0.19) +   & 5.33 (0.69) \\
UF7                          & 0.37 (0.06) +    & \cellcolor[HTML]{C0C0C0}0.32 (0.11) +   & 0.33 (0.05) +   & 1.01 (0.09) \\ \hline
+/$\approx$/-- & \multicolumn{1}{c}{9/4/3} & \multicolumn{1}{c}{9/5/2} & \multicolumn{1}{c}{9/4/3} \\ \hline
\end{tabular}}
\end{table}


\begin{table*}[h]
\caption{Mean (Standard Deviation) IGD$^{+}$ values obtained by BO-AAF, SBP-BO-C, SBP-BO-R, BO-NoGP$^c$ and SBP-BO with $FE_{max}^{e}=300$ and $r^c=5$.}
\label{Tab.4}
\center
\resizebox{0.7\textwidth}{!}{
\setlength{\tabcolsep}{0.9mm}{
\begin{tabular}{llccccc}
\toprule
Problem        & m  & BO-AAF       & SBP-BO-C      & SBP-BO-R  & BO-NoGP$^c$ & SBP-BO                                \\  \midrule[0.3pt]
                             & 3  & 6.26e+1   (2.01e+1) $\approx$         & 8.18e+1   (1.79e+1) $\approx$   & 1.06e+2   (2.57e+1) -- & \cellcolor[HTML]{C0C0C0}4.14e+1 (8.26e+0) $\approx$
                                             & 5.35e+1   (2.10e+1) \\
                           & 5  & 3.24e+1   (1.51e+1) --                                                & 3.87e+1 (1.42e+1) --                           & 4.77e+1 (1.86e+1) --      &  2.59e+1 (5.24e+0) {$\approx$} 
                                      & \cellcolor[HTML]{C0C0C0}2.45e+1 (8.88e+0)   \\
\multirow{-3}{*}{DTLZ1}    & 10 &2.05e-1   (1.13e-1) --                                                & 2.05e-1 (4.17e-2) --                                                & 1.95e-1 (4.35e-2) --         &    1.63e-1 (5.84e-2) $\approx$ & \cellcolor[HTML]{C0C0C0}1.58e-1 (2.31e-2)   \\ \midrule[0.3pt]
                           & 3  & \cellcolor[HTML]{C0C0C0}3.95e-2   (2.71e-3) $\approx$ & 5.74e-2 (3.45e-3) --                                                & 7.87e-2 (1.22e-2) --  
                           &1.78e-1 (1.90e-2) --
                         & 4.15e-2 (3.27e-3)                                                  \\
                           & 5  & 1.06e-1   (5.42e-3) $\approx$                                                & \cellcolor[HTML]{C0C0C0}1.04e-1 (5.16e-3) $\approx$ & 1.54e-1 (1.14e-2) --                 &1.65e-1 (1.66e-2) $\approx$             & 1.06e-1 (7.07e-3)                                                  \\
\multirow{-3}{*}{DTLZ2}    & 10 & 2.07e-1   (1.22e-2) $\approx$                                               & 2.08e-1 (1.12e-2) $\approx$                                              & 2.08e-1 (1.02e-2) $\approx$    &3.38e-1 (6.38e-2) --
                                           & \cellcolor[HTML]{C0C0C0} 2.04e-1 (9.70e-3)   \\ \midrule[0.3pt]
                           & 3  & 1.78e+2   (5.19e+1) $\approx$                                              & 2.58e+2 (6.64e+1) --                                                & 2.87e+2 (5.33e+1) -- &\cellcolor[HTML]{C0C0C0}9.48e+1 (1.61e+1) +
                                               & 1.72e+2 (4.57e+1)   \\
                           & 5  &  8.91e+1   (3.30e+1) $\approx$ & 1.53e+2 (2.66e+1) --                                                & 1.47e+2 (5.51e+1) --  &\cellcolor[HTML]{C0C0C0}4.14e+1 (1.74e+1) +
                                              & 9.17e+1 (2.53e+1)                                                  \\
\multirow{-3}{*}{DTLZ3}    & 10 & \cellcolor[HTML]{C0C0C0} 5.96e-1   (1.32e-1) $\approx$ &4.96e-1 (2.97e-1) +
& 6.61e-1 (4.98e-2) --                                                & 6.88e-1 (1.53e-1) $\approx$                                               & 6.11e-1 (1.48e-1)                                                  \\ \midrule[0.3pt]
                           & 3  & 2.39e-1   (7.60e-2) $\approx$                                                & 2.32e-1 (5.58e-2) $\approx$                                              & 3.14e-1 (8.79e-2) --    &2.35e-1 (4.63e-2) $\approx$
                                       & \cellcolor[HTML]{C0C0C0}2.31e-1 (1.29e-1)   \\
                           & 5  & 2.85e-1   (6.61e-2) $\approx$                                              & 2.93e-1 (4.38e-2) $\approx$                                              & 3.86e-1 (1.02e-1) -- &\cellcolor[HTML]{C0C0C0}1.81e-1 (1.68e-2) +
                       & 2.89e-1 (9.04e-2)                                                  \\
\multirow{-3}{*}{DTLZ4}    & 10 & 2.61e-1   (2.15e-2) $\approx$                                             & 2.50e-1 (2.71e-2) $\approx$ & 2.50e-1 (2.62e-2) $\approx$                   &\cellcolor[HTML]{C0C0C0}2.19e-1 (1.67e-2) +
                             & 2.57e-1 (2.60e-2)                                                  \\ \midrule[0.3pt]
                           & 3  & 3.09e-2   (2.87e-3) $\approx$                                                & 3.83e-2 (5.50e-3) $\approx$                                                & 3.89e-2 (1.08e-2) $\approx$   &2.12e-1 (3.26e-2) --
                                            & \cellcolor[HTML]{C0C0C0} 2.98e-2 (3.24e-3)  \\
                           & 5  & 2.22e-2   (9.99e-3) --                                               & 1.91e-2 (2.52e-3) $\approx$                                               & \color[HTML]{000000} 1.89e-2 (2.54e-3) $\approx$ &1.03e-1 (3.79e-2) $\approx$
& 1.89e-2 (3.55e-3)                                                  \\
\multirow{-3}{*}{DTLZ5}    & 10 & 7.13e-3   (1.51e-3) $\approx$                                                & 7.09e-3 (8.51e-4) $\approx$                                           & \cellcolor[HTML]{C0C0C0}6.80e-3 (1.05e-3) $\approx$ &8.27e-3 (1.70e-3) $\approx$
 & 7.22e-3 (9.88e-4)                                                  \\ \midrule[0.3pt]
                           & 3  & 3.33e+0   (3.45e-1) --                                                & 3.48e+0 (5.70e-1) --                                                & 3.36e+0 (4.48e-1) --   &3.94e+0 (7.22e-1) --
                        & \cellcolor[HTML]{C0C0C0}3.05e+0 (4.83e-1)   \\
                           & 5  & 2.34e+0   (4.28e-1) --                                                & 2.12e+0 (4.22e-1) $\approx$                                                & 2.17e+0 (3.70e-1) $\approx$   &2.44e+0 (7.64e-1) --
                                       & \cellcolor[HTML]{C0C0C0}1.83e+0 (4.89e-1)  \\
\multirow{-3}{*}{DTLZ6}    & 10 & 4.32e-2   (4.05e-2) $\approx$                                              & 1.80e-2 (6.23e-3) $\approx$                                             & \cellcolor[HTML]{C0C0C0}1.77e-2 (7.33e-3) +       &3.06e-1 (4.84e-2) --
                 & 2.46e-2 (7.67e-3)                                                  \\ \midrule[0.3pt]
                           & 3  & 1.64e-1   (4.94e-2) $\approx$                                                & 1.68e-1 (5.70e-2) $\approx$        &6.82e-1 (1.29e-1) --
                                        & \cellcolor[HTML]{C0C0C0}1.22e-1 (3.92e-2) + & 1.65e-1 (4.89e-2)                                                  \\
                           & 5  & 9.29e-1   (3.73e-1) $\approx$                                                & \cellcolor[HTML]{C0C0C0}6.56e-1 (1.65e-1) $\approx$                         & 7.41e-1 (2.31e-1) $\approx$   &2.32e+0 (6.45e-1) --
                                             & 9.98e-1 (3.48e-1)                                                  \\
\multirow{-3}{*}{DTLZ7}    & 10 & 1.32e+0   (2.48e-1) --                                                & \cellcolor[HTML]{C0C0C0}1.05e+0 (1.33e-1) $\approx$ & 1.19e+0 (2.46e-1) $\approx$      &1.45e+0 (3.39e-1) --                            & 1.13e+0 (1.70e-1)                                                  \\ \midrule[0.3pt]
                           & 3  & 1.74e+0   (1.33e-1) $\approx$                      & 1.88e+0 (8.85e-2) --                              & 1.86e+0 (1.54e-1) $\approx$    &\cellcolor[HTML]{C0C0C0} 1.63e+0 (2.33e-2) +
                            & 1.75e+0 (1.34e-1)                                                  \\
                           & 5  & 2.18e+0   (8.17e-2) $\approx$                        & 2.32e+0 (5.55e-2) --                                                & 2.31e+0 (1.05e-1) -- &\cellcolor[HTML]{C0C0C0}2.13e+0 (3.42e-2) $\approx$
                                               & 2.20e+0 (6.82e-2)                                                  \\
\multirow{-3}{*}{WFG1}     & 10 & 2.85e+0   (1.05e-1) $\approx$                                             & 2.95e+0 (5.59e-2) $\approx$                                                & 2.93e+0 (1.29e-1) $\approx$    &\cellcolor[HTML]{C0C0C0}9.59e-1 (9.21e-1) + & 2.79e+0 (1.46e-1)   \\ \midrule[0.3pt]
                           & 3  & \cellcolor[HTML]{C0C0C0}{\color[HTML]{000000} 1.71e-1   (2.11e-2) $\approx$} & 2.96e-1 (3.58e-2) --                                               & 2.05e-1 (3.66e-2) $\approx$  &4.02e-1 (7.57e-2) --                           & 1.80e-1 (2.81e-2)                                                  \\
                           & 5  & 2.20e-1   (3.09e-2) $\approx$                                              & 3.25e-1 (3.14e-2) --                                                & 2.11e-1 (1.80e-2) $\approx$    &3.26e-1 (1.51e-2) -- & \cellcolor[HTML]{C0C0C0}2.10e-1 (2.89e-2)   \\
\multirow{-3}{*}{WFG2}     & 10 & 2.80e-1   (7.69e-2) $\approx$ & 4.84e-1 (1.98e-1) --                                               & 3.47e-1 (9.82e-2) $\approx$    & \cellcolor[HTML]{C0C0C0}2.01e-1 (2.32e-2) +    & 3.11e-1 (8.55e-2)                                                  \\ \midrule[0.3pt]
                           & 3  & \cellcolor[HTML]{C0C0C0}2.08e-1   (2.01e-2) $\approx$ & 2.98e-1 (1.79e-2) --                                               & 2.53e-1 (2.98e-2) -- 
                           &5.09e-1 (2.81e-2) --
                           & 2.10e-1 (3.02e-2)                                                  \\
                           & 5  & 3.76e-1   (5.16e-2) --                                                & 4.09e-1 (3.12e-2) --                                                & 3.19e-1 (4.20e-2) $\approx$  &6.26e-1 (7.80e-2) --                      &\cellcolor[HTML]{C0C0C0}3.47e-1 (8.35e-2)                                                  \\
\multirow{-3}{*}{WFG3}     & 10 & 5.64e-1   (1.03e-1) $\approx$                                               & 5.74e-1 (1.09e-1) $\approx$                                                & 5.47e-1 (8.30e-2) $\approx$       &7.88e-1 (1.02e-1) --                                        & \cellcolor[HTML]{C0C0C0} 5.35e-1 (7.64e-2)  \\\midrule[0.3pt]
                           & 3  & \cellcolor[HTML]{C0C0C0}3.15e-1   (2.17e-2) $\approx$ & 3.71e-1 (3.64e-2) --        & 3.58e-1 (1.98e-2) --   &3.65e-1 (6.95e-3) --  & 3.21e-1 (2.28e-2)                                                  \\
                           & 5  & 7.07e-1   (2.84e-2) $\approx$     & 7.72e-1 (4.34e-2) --    & 7.10e-1 (5.99e-2) $\approx$   &7.39e-1 (3.38e-2) --  & \cellcolor[HTML]{C0C0C0}7.06e-1 (3.89e-2)   \\
\multirow{-3}{*}{WFG4}     & 10 & 3.18e+0   (7.82e-1)  $\approx$                                               & 3.64e+0 (8.06e-1) $\approx$                                               & 4.34e+0 (9.00e-1) -- &3.13e+0 (1.19e+0) $\approx$                                               & \cellcolor[HTML]{C0C0C0}3.10e+0 (7.03e-1) \\ \midrule[0.3pt]
                           & 3  & \cellcolor[HTML]{C0C0C0}2.77e-1   (5.72e-2) $\approx$ & 3.10e-1 (2.86e-2) $\approx$                                                & 3.38e-1 (8.51e-2) $\approx$            &4.98e-1 (2.40e-2) --                                    & 2.99e-1 (8.32e-2)                                                  \\
                           & 5  & 6.75e-1   (6.23e-2) $\approx$     & 7.26e-1 (4.34e-2) --                  & 7.31e-1 (7.06e-2) --  &9.25e-1 (2.65e-2) -- & \cellcolor[HTML]{C0C0C0} 6.59e-1 (4.20e-2)   \\
\multirow{-3}{*}{WFG5}     & 10 & 1.95e+0   (4.75e-1) $\approx$                                                & 2.02e+0 (2.27e-1) $\approx$ & 2.28e+0 (1.03e+0) $\approx$      &2.14e+0 (5.89e-1) $\approx$
                                        & \cellcolor[HTML]{C0C0C0}1.78e+0 (3.69e-1)   \\ \midrule[0.3pt]
                           & 3  & \cellcolor[HTML]{C0C0C0}3.86e-1   (6.99e-2) $\approx$ & 4.74e-1 (5.82e-2) $\approx$                                                & 4.16e-1 (7.36e-2) $\approx$   &5.63e-1 (1.53e-2) --
                                           & 4.16e-1 (8.37e-2)     \\
                           & 5  & 7.80e-1   (9.71e-2) $\approx$     & 7.73e-1 (4.16e-2) $\approx$            & 7.08e-1 (8.41e-2) $\approx$  &8.71e-1 (4.61e-2) --
                          &\cellcolor[HTML]{C0C0C0} 7.61e-1 (8.25e-2)       \\
\multirow{-3}{*}{WFG6}     & 10 & 1.05e+0   (2.51e-2) $\approx$                                               & 1.11e+0 (7.31e-2) $\approx$                                                 & 1.12e+0 (3.17e-2) $\approx$  &1.17e+0 (6.15e-2)                 -- & \cellcolor[HTML]{C0C0C0}1.03e+0 (2.96e-2)   \\\midrule[0.3pt]
                           & 3  & \cellcolor[HTML]{C0C0C0}4.19e-1   (3.69e-2) $\approx$ & 4.81e-1 (4.36e-2) --                                                & 4.74e-1 (3.30e-2) --  
                           &4.63e-1 (4.12e-2) $\approx$
& 4.35e-1 (4.11e-2)                                                  \\
                           & 5  & 7.48e-1   (7.37e-2) $\approx$     & 9.00e-1 (9.24e-2) --                 & 7.96e-1 (4.26e-2) --   &1.14e+0 (1.43e-1) --
                     &\cellcolor[HTML]{C0C0C0} 7.28e-1 (4.22e-2)   \\
\multirow{-3}{*}{WFG7}     & 10 & 3.40e+0   (4.47e-1) $\approx$                                           & 3.99e+0 (7.68e-1) --                                                & 4.16e+0 (7.49e-1) --      &4.21e+0 (1.13e+0) --
                                          &\cellcolor[HTML]{C0C0C0} 3.33e+0 (4.32e-1)   \\ \midrule[0.3pt]
                           & 3  & 5.03e-1   (2.86e-2) --            & 5.62e-1 (2.24e-2) --                 & 5.36e-1 (3.61e-2) --   &7.22e-1 (5.01e-2) -- &\cellcolor[HTML]{C0C0C0} 4.86e-1 (2.89e-2) \\
                           & 5  & 1.29e+0   (2.66e-2) $\approx$     & 1.29e+0 (5.59e-2) $\approx$           & 1.29e+0 (4.99e-2) $\approx$ &1.71e+0 (9.66e-2) -- &\cellcolor[HTML]{C0C0C0}1.25e+0 (3.85e-2)  \\
\multirow{-3}{*}{WFG8}     & 10 & 2.08e+0   (9.08e-1) --                                                & 1.96e+0 (3.72e-1) $\approx$                                               & 2.36e+0 (7.12e-1) --      &3.05e+0 (1.79e+0) --
                & \cellcolor[HTML]{C0C0C0} 1.88e+0 (6.99e-1)   \\\midrule[0.3pt]
                           & 3  & 4.61e-1   (1.20e-1) $\approx$ & 6.98e-1 (9.33e-2) --                                                & 5.16e-1 (7.21e-2) $\approx$   
                           &\cellcolor[HTML]{C0C0C0}4.25e-1 (4.11e-2) +
& 5.50e-1 (1.36e-1)                                                  \\
                           & 5  & 1.27e+0   (2.24e-1) $\approx$                                               & 1.27e+0 (2.10e-1) $\approx$                                             & 1.10e+0 (2.42e-1) $\approx$ &\cellcolor[HTML]{C0C0C0}9.48e-1 (1.18e-1) $\approx$
 & 1.14e+0 (1.94e-1)                                                  \\
\multirow{-3}{*}{WFG9}     & 10 & 4.76e+0   (5.95e-1) $\approx$                                                & 4.93e+0 (1.07e+0) $\approx$                                                & 5.41e+0 (6.20e-1) --      &4.75e+0 (1.59e+0) $\approx$
                                          &\cellcolor[HTML]{C0C0C0}4.69e+0 (8.73e-1)     \\ \midrule[0.3pt]
+/--/$\approx$ &    & 0/9/39     & 0/23/25   & 2/21/25   & 9/14/25                                &                                           \\ \bottomrule
\end{tabular}}}
\end{table*}

4) \emph{Ablation studies}

Further experiments are performed here to provide a deeper understanding of the performance of SBP-BO by testing the effectiveness of each component. The IGD$^{+}$ values obtained by SBP-BO and its variants are presented in Table \ref{Tab.4} and Table SVI in the Supplementary material. The following observations can be made:

\begin{itemize}
\item According to Tables \ref{Tab.4} and SVI, we can see that the proposed algorithm yields the best IGD$^{+}$ values on 22 and 14 out of 48 test instances for $r^c=5$ and $r^c=10$ respectively, confirming the effectiveness of the ensemble model and the search bias penalized acquisition function. 
\item The effectiveness of the proposed way of utilizing the additional data in SBP-BO can be validated by comparing SBP-BO with SBP-BO-C and SBP-BO-R. From Table \ref{Tab.4}, SBP-BO significantly outperforms SBP-BO-C and SBP-BO-R on 23 and 21 test problems, respectively. This is consistent with the findings in \cite{XiluWang2020TSAEA}: how to utilize the additional data obtained from the search of cheap objectives plays a vital role in the optimization of HE-MOPs/HE-MaOPs. One weakness of the commonly used methods for training data selection based on clustering or randomly selection is that they cannot use all available data. This issue becomes more challenging for HE-MOPs where abundant training data are available for the fast objectives, making the algorithm inefficient for addressing problems with heterogeneous objectives. Similar observation can be made from Table SVI. Note that there is no limitation on the available FEs for the relatively cheap objectives and thousands of cheap FEs are consumed in SBP-NoGP$^c$. It is interesting to see that SBP-BO is able to significantly outperform SBP-NoGP$^c$ on 14 and 17 out of 48 test instances for $r^c=5$ and $r^c=10$, respectively. The comparison between SBP-BO and SBP-NoGP$^c$ indicates that the algorithm can benefit from the use of surrogates on the cheap objectives. 
A possible explanation is that surrogates may smooth out some local optima and thus accelerate the search, which was discussed intuitively in \cite{husken2005structure} and empirically verified in \cite{lim2009generalizing}.

\item Compared with BO-AAF, the proposed algorithm shows significantly better performance on 9 out of 48 test instances, and similar performance on the remaining test problems, according to the results in Table \ref{Tab.4}.
It is worthy of noting that for HE-MOPs/HE-MaOPs with $r^c=10$, the advantage of SBP-BO becomes a little less clear compared with BO-AAF, as can be observed from the results in Table SVI. SBP-BO is worse than BO-AFF on one test instance, but it only outperforms BO-AFF on 7 out of 48 instances. The results indicate that the algorithm can benefit from the use of the search bias penalty on some problems. However, since it is highly tricky to measure the search bias, it is challenging to apply an appropriate degree of penalty. This is might be the reason why SBP-BO and BO-AAF show similar performance on most test problems.

\end{itemize}

\section{Conclusion}
In this paper, we address heterogeneously expensive multi-/many-objective optimization problems, which have not received much attention in the evolutionary optimization community. We focus on exploiting the different amounts of data for the cheap and expensive objectives in constructing surrogates and reducing the search bias towards the cheap objectives within the Bayesian optimization framework. Specifically, an ensemble of Gaussian processes is constructed for each cheap objective to make use of both the solutions evaluated on all objectives and on the cheap objectives only. To reduce the bias towards the cheap objectives, we introduce a penalty term into the acquisition function, guiding the selection of new samples by taking the search bias into consideration. Different from most state-of-the-art algorithms that are limited to bi-objective optimization problems, the proposed algorithm is more generic in that it is applicable to problems with more than two objectives, where each objective can have a different evaluation time. Thus, the proposed work constitutes a valuable step forward towards solving real-world problems.  

Encouraged by the promising results of the present work, we are interested in further investigating the efficient use of additional data on the cheap objectives, e.g., by properly guiding the single objective search. Meanwhile, the experimental results of the current work suggest that the proposed algorithm is less effective on nonseparable, multi-modal and disconnected problems, implying that more powerful search operators are required. Finally, this work adopts a simplified way to measure the search bias resulting from heterogeneous objectives, which leaves much room for further improvement in alleviating the search bias.  

\section*{ACKNOWLEDGEMENT}
This work was supported in part by the Honda Research Institute Europe and in part by a Royal Society International Exchanges Program under No. IEC$\backslash$NSFC$\backslash$170279. YJ is supported in part by an Alexander von Humboldt Professorship endowed by the German Federal Ministry of Education and Research.  
\ifCLASSOPTIONcaptionsoff
  \newpage
\fi


{\footnotesize\bibliography{reference}

\begin{thebibliography}{10}

\bibitem{jin2011surrogate}
Y.~Jin, ``Surrogate-assisted evolutionary computation: Recent advances and
  future challenges,'' {\em Swarm and Evolutionary Computation}, vol.~1, no.~2,
  pp.~61--70, 2011.

\bibitem{allmendinger2015multiobjective}
R.~Allmendinger, J.~Handl, and J.~Knowles, ``Multiobjective optimization: When
  objectives exhibit non-uniform latencies,'' {\em European Journal of
  Operational Research}, vol.~243, no.~2, pp.~497--513, 2015.

\bibitem{jin2019data}
Y.~Jin, H.~Wang, T.~Chugh, D.~Guo, and K.~Miettinen, ``Data-driven evolutionary
  optimization: An overview and case studies,'' {\em IEEE Transactions on
  Evolutionary Computation}, vol.~23, no.~3, pp.~442--458, 2019.

\bibitem{kang2016slope}
F.~Kang, Q.~Xu, and J.~Li, ``Slope reliability analysis using surrogate models
  via new support vector machines with swarm intelligence,'' {\em Applied
  Mathematical Modelling}, vol.~40, no.~11-12, pp.~6105--6120, 2016.

\bibitem{guo2018heterogeneous}
D.~Guo, Y.~Jin, J.~Ding, and T.~Chai, ``Heterogeneous ensemble-based infill
  criterion for evolutionary multiobjective optimization of expensive
  problems,'' {\em IEEE transactions on cybernetics}, vol.~49, no.~3,
  pp.~1012--1025, 2018.

\bibitem{pan2018classification}
L.~Pan, C.~He, Y.~Tian, H.~Wang, X.~Zhang, and Y.~Jin, ``A classification-based
  surrogate-assisted evolutionary algorithm for expensive many-objective
  optimization,'' {\em IEEE Transactions on Evolutionary Computation}, vol.~23,
  no.~1, pp.~74--88, 2018.

\bibitem{chugh2018surrogate}
T.~Chugh, Y.~Jin, K.~Miettinen, J.~Hakanen, and K.~Sindhya, ``A
  surrogate-assisted reference vector guided evolutionary algorithm for
  computationally expensive many-objective optimization,'' {\em IEEE
  Transactions on Evolutionary Computation}, vol.~22, no.~1, pp.~129--142,
  2018.

\bibitem{wang2020adaptive}
X.~Wang, Y.~Jin, S.~Schmitt, and M.~Olhofer, ``An adaptive bayesian approach to
  surrogate-assisted evolutionary multi-objective optimization,'' {\em
  Information Sciences}, vol.~519, pp.~317--331, 2020.

\bibitem{shahriari2015taking}
B.~Shahriari, K.~Swersky, Z.~Wang, R.~P. Adams, and N.~De~Freitas, ``Taking the
  human out of the loop: A review of {B}ayesian optimization,'' {\em
  Proceedings of the IEEE}, vol.~104, no.~1, pp.~148--175, 2015.

\bibitem{jones1998efficient}
D.~R. Jones, M.~Schonlau, and W.~J. Welch, ``Efficient global optimization of
  expensive black-box functions,'' {\em Journal of Global Optimization},
  vol.~13, no.~4, pp.~455--492, 1998.

\bibitem{snoek2012practical}
J.~Snoek, H.~Larochelle, and R.~P. Adams, ``Practical bayesian optimization of
  machine learning algorithms,'' {\em Advances in neural information processing
  systems}, vol.~25, 2012.

\bibitem{allmendinger2021heterogeneous}
R.~Allmendinger and J.~Knowles, ``Heterogeneous objectives: state-of-the-art
  and future research,'' {\em arXiv preprint arXiv:2103.15546}, 2021.

\bibitem{allmendinger2013hang}
R.~Allmendinger and J.~Knowles, ``{Hang On a Minute}: Investigations on the
  effects of delayed objective functions in multiobjective optimization,'' in
  {\em International Conference on Evolutionary Multi-Criterion Optimization},
  pp.~6--20, Springer, 2013.

\bibitem{chugh2018hkrvea}
T.~Chugh, R.~Allmendinger, V.~Ojalehto, and K.~Miettinen, ``Surrogate-assisted
  evolutionary biobjective optimization for objectives with non-uniform
  latencies,'' in {\em Proceedings of the Genetic and Evolutionary Computation
  Conference}, pp.~609--616, 2018.

\bibitem{XiluWang2020TSAEA}
X.~Wang, Y.~Jin, S.~Schmitt, and M.~Olhofer, ``Transfer learning for gaussian
  process assisted evolutionary bi-objective optimization for objectives with
  different evaluation times,'' in {\em Proceedings of the 2020 Genetic and
  Evolutionary Computation Conference}, pp.~587--594, 2020.

\bibitem{wang2021transfer}
X.~Wang, Y.~Jin, S.~Schmitt, M.~Olhofer, and R.~Allmendinger, ``Transfer
  learning based surrogate assisted evolutionary bi-objective optimization for
  objectives with different evaluation times,'' {\em Knowledge-Based Systems},
  p.~107190, 2021.

\bibitem{WangTC-SAEA2021}
X.~Wang, Y.~Jin, S.~Schmitt, and M.~Olhofer, ``Transfer learning based
  co-surrogate assisted evolutionary bi-objective optimization for objectives
  with non-uniform evaluation times,'' {\em Evolutionary computation},
  pp.~1--27, 2021.

\bibitem{thomann2019representation}
J.~Thomann and G.~Eichfelder, ``Representation of the {P}areto front for
  heterogeneous multi-objective optimization,'' {\em J. Appl. Numer. Optim},
  vol.~1, no.~3, pp.~293--323, 2019.

\bibitem{thomann2019trust}
J.~Thomann and G.~Eichfelder, ``A trust-region algorithm for heterogeneous
  multiobjective optimization,'' {\em SIAM Journal on Optimization}, vol.~29,
  no.~2, pp.~1017--1047, 2019.

\bibitem{gelbart2014bayesian}
M.~A. Gelbart, J.~Snoek, and R.~P. Adams, ``Bayesian optimization with unknown
  constraints,'' in {\em Proceedings of the Thirtieth Conference on Uncertainty
  in Artificial Intelligence}, pp.~250--259, 2014.

\bibitem{hernandez2016predictive}
D.~Hern{\'a}ndez-Lobato, J.~Hernandez-Lobato, A.~Shah, and R.~Adams,
  ``Predictive entropy search for multi-objective {B}ayesian optimization,'' in
  {\em International Conference on Machine Learning}, pp.~1492--1501, PMLR,
  2016.

\bibitem{suzuki2020multi}
S.~Suzuki, S.~Takeno, T.~Tamura, K.~Shitara, and M.~Karasuyama,
  ``Multi-objective {B}ayesian optimization using {P}areto-frontier entropy,''
  in {\em International Conference on Machine Learning}, pp.~9279--9288, PMLR,
  2020.

\bibitem{belakaria2019max}
S.~Belakaria and A.~Deshwal, ``Max-value entropy search for multi-objective
  {B}ayesian optimization,'' in {\em International Conference on Neural
  Information Processing Systems (NeurIPS)}, 2019.

\bibitem{allmendinger2017surrogate}
R.~Allmendinger, M.~T. Emmerich, J.~Hakanen, Y.~Jin, and E.~Rigoni,
  ``Surrogate-assisted multicriteria optimization: Complexities, prospective
  solutions, and business case,'' {\em Journal of Multi-Criteria Decision
  Analysis}, vol.~24, no.~1-2, pp.~5--24, 2017.

\bibitem{qin2019bayesian}
S.~Qin, C.~Sun, Y.~Jin, and G.~Zhang, ``Bayesian approaches to
  surrogate-assisted evolutionary multi-objective optimization: a comparative
  study,'' in {\em 2019 IEEE Symposium Series on Computational Intelligence
  (SSCI)}, pp.~2074--2080, IEEE, 2019.

\bibitem{knowles2006parego}
J.~Knowles, ``{ParEGO}: A hybrid algorithm with on-line landscape approximation
  for expensive multiobjective optimization problems,'' {\em IEEE Transactions
  on Evolutionary Computation}, vol.~10, no.~1, pp.~50--66, 2006.

\bibitem{zhang2007moea}
Q.~Zhang and H.~Li, ``{MOEA/D}: A multiobjective evolutionary algorithm based
  on decomposition,'' {\em IEEE Transactions on Evolutionary Computation},
  vol.~11, no.~6, pp.~712--731, 2007.

\bibitem{cheng2016reference}
R.~Cheng, Y.~Jin, M.~Olhofer, and B.~Sendhoff, ``A reference vector guided
  evolutionary algorithm for many-objective optimization,'' {\em IEEE
  Transactions on Evolutionary Computation}, vol.~20, no.~5, pp.~773--791,
  2016.

\bibitem{zhang2009expensive}
Q.~Zhang, W.~Liu, E.~Tsang, and B.~Virginas, ``Expensive multiobjective
  optimization by {MOEA/D} with gaussian process model,'' {\em IEEE
  Transactions on Evolutionary Computation}, vol.~14, no.~3, pp.~456--474,
  2009.

\bibitem{zitzler2003performance}
E.~Zitzler, L.~Thiele, M.~Laumanns, C.~M. Fonseca, and V.~G. Da~Fonseca,
  ``Performance assessment of multiobjective optimizers: An analysis and
  review,'' {\em IEEE Transactions on Evolutionary Computation}, vol.~7, no.~2,
  pp.~117--132, 2003.

\bibitem{ponweiser2008multiobjective}
W.~Ponweiser, T.~Wagner, D.~Biermann, and M.~Vincze, ``Multiobjective
  optimization on a limited budget of evaluations using model-assisted
  $\mathcal{S}$-metric selection,'' in {\em International Conference on
  Parallel Problem Solving from Nature}, pp.~784--794, Springer, 2008.

\bibitem{beume2007sms}
N.~Beume, B.~Naujoks, and M.~Emmerich, ``{SMS-EMOA}: Multiobjective selection
  based on dominated hypervolume,'' {\em European Journal of Operational
  Research}, vol.~181, no.~3, pp.~1653--1669, 2007.

\bibitem{emmerich2006single}
M.~T. Emmerich, K.~C. Giannakoglou, and B.~Naujoks, ``Single-and multiobjective
  evolutionary optimization assisted by {G}aussian random field metamodels,''
  {\em IEEE Transactions on Evolutionary Computation}, vol.~10, no.~4,
  pp.~421--439, 2006.

\bibitem{couckuyt2014fast}
I.~Couckuyt, D.~Deschrijver, and T.~Dhaene, ``Fast calculation of
  multiobjective probability of improvement and expected improvement criteria
  for {P}areto optimization,'' {\em Journal of Global Optimization}, vol.~60,
  no.~3, pp.~575--594, 2014.

\bibitem{yang2019multi}
K.~Yang, M.~Emmerich, A.~Deutz, and T.~B{\"a}ck, ``Multi-objective {B}ayesian
  global optimization using expected hypervolume improvement gradient,'' {\em
  Swarm and evolutionary computation}, vol.~44, pp.~945--956, 2019.

\bibitem{hernandez2014predictive}
J.~M. Hern{\'a}ndez-Lobato, M.~W. Hoffman, and Z.~Ghahramani, ``Predictive
  entropy search for efficient global optimization of black-box functions,''
  {\em arXiv preprint arXiv:1406.2541}, 2014.

\bibitem{cox1992statistical}
D.~D. Cox and S.~John, ``A statistical method for global optimization,'' in
  {\em [Proceedings] 1992 IEEE International Conference on Systems, Man, and
  Cybernetics}, pp.~1241--1246, IEEE, 1992.

\bibitem{cao2014generalized}
Y.~Cao and D.~J. Fleet, ``Generalized product of experts for automatic and
  principled fusion of {G}aussian process predictions,'' {\em arXiv preprint
  arXiv:1410.7827}, 2014.

\bibitem{abdolshah2019cost}
M.~Abdolshah, A.~Shilton, S.~Rana, S.~Gupta, and S.~Venkatesh, ``Cost-aware
  multi-objective bayesian optimisation,'' {\em arXiv preprint
  arXiv:1909.03600}, 2019.

\bibitem{deb2002scalable}
K.~Deb, L.~Thiele, M.~Laumanns, and E.~Zitzler, ``Scalable multi-objective
  optimization test problems,'' in {\em Proceedings of the 2002 Congress on
  Evolutionary Computation. CEC'02 (Cat. No. 02TH8600)}, vol.~1, pp.~825--830,
  IEEE, 2002.

\bibitem{huband2006review}
S.~Huband, P.~Hingston, L.~Barone, and L.~While, ``A review of multiobjective
  test problems and a scalable test problem toolkit,'' {\em IEEE Transactions
  on Evolutionary Computation}, vol.~10, no.~5, pp.~477--506, 2006.

\bibitem{ishibuchi2015modified}
H.~Ishibuchi, H.~Masuda, Y.~Tanigaki, and Y.~Nojima, ``Modified distance
  calculation in generational distance and inverted generational distance,'' in
  {\em International Conference on Evolutionary Multi-criterion Optimization},
  pp.~110--125, Springer, 2015.

\bibitem{lophaven2002dace}
S.~N. Lophaven, H.~B. Nielsen, J.~S{\o}ndergaard, {\em et~al.}, {\em {DACE}: a
  {M}atlab kriging toolbox}, vol.~2.
\newblock Citeseer, 2002.

\bibitem{zhang2008multiobjective}
Q.~Zhang, A.~Zhou, S.~Zhao, P.~N. Suganthan, W.~Liu, S.~Tiwari, {\em et~al.},
  ``Multiobjective optimization test instances for the {CEC} 2009 special
  session and competition,'' tech. rep., University of Essex, Colchester, UK
  and Nanyang technological University, 2008.

\bibitem{husken2005structure}
M.~H{\"u}sken, Y.~Jin, and B.~Sendhoff, ``Structure optimization of neural
  networks for evolutionary design optimization,'' {\em Soft Computing},
  vol.~9, no.~1, pp.~21--28, 2005.

\bibitem{lim2009generalizing}
D.~Lim, Y.~Jin, Y.-S. Ong, and B.~Sendhoff, ``Generalizing surrogate-assisted
  evolutionary computation,'' {\em IEEE Transactions on Evolutionary
  Computation}, vol.~14, no.~3, pp.~329--355, 2009.

\end{thebibliography}
\bibliographystyle{ieeetr}}

\end{document}